\documentclass[twoside]{article}
\usepackage[T1]{fontenc}
\usepackage{lmodern}

\usepackage[accepted]{aistats2026}
\usepackage[round]{natbib}

\usepackage[utf8]{inputenc} 
\usepackage[T1]{fontenc}    
\usepackage{url}            
\usepackage{booktabs}       
\usepackage{nicefrac}       
\usepackage{microtype}      
\usepackage{xcolor}         

\usepackage{fancyhdr}       
\usepackage{graphicx}       
\graphicspath{{media/}}     
\usepackage{adjustbox}

\usepackage{wrapfig,lipsum}

\usepackage{amssymb,latexsym,amsfonts,amsmath,amsthm}

\usepackage{thmtools}
\usepackage{xspace}

\usepackage{booktabs}
\usepackage{mathtools}
\usepackage{tikz} 
\usepackage{dsfont}

\usepackage{mathtools}

\usepackage{dsfont}

\usepackage{graphicx} 
\usepackage{booktabs} 

\usepackage{color,soul}
\usepackage{bm}
\usepackage{graphicx}
\usepackage[inline]{enumitem}
\setlist[itemize]{topsep=2pt, partopsep=0pt, parsep=0pt, itemsep=2pt}

\usepackage{pythonhighlight}
\usepackage{nccmath}
\usepackage{algorithm}
\usepackage{algpseudocode}

\usepackage{booktabs}
\usepackage{hyperref}
\usepackage{float,caption,hypcap}

\usepackage{booktabs}
\usepackage{pifont}

\usepackage{scalerel}
\usepackage{comment}

\usepackage[capitalize]{cleveref}
\usepackage[textsize=tiny]{todonotes}

\newtheorem{theorem}{Theorem}[section]
\newtheorem{assumption}[theorem]{Assumption}

\newtheorem{Lemma}[theorem]{Lemma}

\newtheorem{proposition}[theorem]{Proposition}

\renewcommand{\Cref}[1]{\cref{#1}}

\newcommand{\nbone}{\ding{182}\xspace}
\newcommand{\nbtwo}{\ding{183}\xspace}
\newcommand{\nbthree}{\ding{184}\xspace}

\usepackage{cases}
\definecolor{tabblue}{RGB}{31, 119, 180}
\definecolor{tabred}{RGB}{214, 39, 40}
\definecolor{tabgreen}{RGB}{44,160,44}

\usepackage{caption}
\crefname{figure}{Figure}{Figures}
\Crefname{figure}{Figure}{Figures}
\newcommand{\newimh}{Latent-IMH}

\newcommand{\dx}{\ensuremath{d_x}}
\newcommand{\dy}{\ensuremath{d_y}}
\newcommand{\du}{\ensuremath{d_u}}
\newcommand{\mR}{\ensuremath{\mathbb{R}}}

\newcommand{\Exact}{{\texttt{Exact posterior}}\xspace}
\newcommand{\Approx}{{\texttt{Approx-IMH posterior}}\xspace}
\newcommand{\Hybrid}{{\texttt{{\newimh} posterior}}\xspace}

\newcommand{\ApproxMH}{{\texttt{Approx-IMH}}\xspace}
\newcommand{\HybridMH}{{\texttt{\newimh}}\xspace}
\newcommand{\KL}{{\mathrm{KL}\xspace}}

\newcommand{\KLapprox}{\mathbb{D}_a}
\newcommand{\KLnew}{\mathbb{D}_l}

\newcommand{\mixapprox}{\tau_{mix}^a}
\newcommand{\mixnew}{\tau_{mix}^l}

\renewcommand{\b}[1]{{\bf #1}}

\newcommand{\A}{\b{A}}

\newcommand{\U}{\b{U}}
\newcommand{\V}{\b{V}}
\newcommand{\Vpar}{\b{V}_{\parallel}}
\newcommand{\Vperp}{\b{V}_{\perp}}

\newcommand{\Uapprox}{\widetilde{\b{U}}}

\newcommand{\Vapproxpar}{\widetilde{\b{V}}_{\parallel}}
\newcommand{\Vapproxperp}{\widetilde{\b{V}}_{\perp}}
\newcommand{\Sapprox}{\widetilde{\b{S}}}
\newcommand{\daapprox}{\mathbf{\Delta}_a}
\newcommand{\danew}{\mathbf{\Delta}_l}

\newcommand{\piapprox}{\ensuremath{\pi_a}}
\newcommand{\pinew}{\ensuremath{\pi_l}}

\newcommand{\xapprox}{x_a}
\newcommand{\xnew}{x_l}
\newcommand{\wapprox}{w_a}
\newcommand{\wnew}{w_l}
\newcommand{\Capprox}{C_a}
\newcommand{\Cnew}{C_l}
\newcommand{\Rapprox}{R_a}
\newcommand{\Rnew}{R_l}

\newcommand{\Cmatapprox}{\widetilde{\b C}}

\newcommand{\Cmatnew}{\widehat{\b C}}

\newcommand{\bpinew}{\widehat{\bm \pi}}

\newcommand{\Oo}{\b O}
\newcommand{\F}{\b F}
\newcommand{\K}{\b K}
\newcommand{\Lo}{\b L}
\newcommand{\B}{\b B}
\newcommand{\Id}{\b I}

\newcommand{\Fapprox}{\widetilde{\F}}
\newcommand{\Lapprox}{\widetilde{\Lo}}
\newcommand{\Aapprox}{\widetilde{\A}}
\newcommand{\Anew}{\A_l}
\newcommand{\aapprox}{a_a}
\newcommand{\anew}{a_l}

\DeclareMathOperator{\E}{\mathbb{E}}

\DeclareMathOperator{\Tr}{Tr}

\DeclareMathOperator*{\argmax}{arg\,max}

\definecolor{mygreen}{RGB}{0,153,0}
\definecolor{light-gray}{gray}{0.93}
\definecolor{mid-gray}{gray}{0.88}

\newcommand{\bSigma}{\bm{\Sigma}}

\makeatletter
\newcommand*{\rom}[1]{\expandafter\@slowromancap\romannumeral #1@}
\makeatother

\newcommand{\apu}[1]{\ensuremath{\widetilde{p}(#1)}}

\begin{document}

%

%

\runningtitle{\newimh: Bayesian Inference with Approximate Operators}
\runningauthor{Chen and Biros}
\twocolumn[

\aistatstitle{
\newimh: Efficient Bayesian Inference for Inverse Problems with Approximate Operators
}

\aistatsauthor{ Youguang Chen \And George Biros}

\aistatsaddress{Oden Institute for Computational Engineering and Sciences\\ The University of Texas at Austin } 

]

\begin{abstract}
  We study sampling from posterior distributions in Bayesian linear inverse problems where $\A$, the parameters         to observables operator, is computationally expensive. In many applications $\A$  can be factored in a manner that facilitates the construction of a cost-effective approximation $\Aapprox$.
  In this framework, we introduce \newimh{}, a sampling method based on the Metropolis-Hastings independence (IMH) sampler. \newimh{} first generates intermediate latent variables using the approximate $\Aapprox$, and then refines them using the exact $\A$. Its primary benefit is that it shifts the computational cost to an offline phase. We theoretically analyze the performance of \newimh{} using KL divergence and mixing time bounds. Using numerical experiments on several model problems, we show that, under reasonable assumptions, it outperforms state-of-the-art methods such as the No-U-Turn sampler (NUTS) in computational efficiency. 
  In some cases \newimh{} can be orders of magnitude faster than existing schemes.
\end{abstract}

\section{INTRODUCTION}\label{sec:introduction}

\begin{figure*}[t]
\small
\centering
    \def\width{2.3}
    \def\vpos{-3}
\begin{tikzpicture}
\node[] at (0,1.55) {$x_{\text{true}}$};
\node[] at (6.4,1.4) {$\textcolor{tabblue}{\overbrace{\hspace{9cm}}^{}}$};
\node[] at (6.4,1.6) {\color{tabblue}\small \ApproxMH};
\draw[line width=1pt,dashed] (1.3, -4.85) rectangle (11.5,1.85);
\node[] at (0,-4.55) {$x_{\text{mean}}$};
\node[] at (6.4,-4.4) {$\textcolor{tabred}{\underbrace{\hspace{9cm}}_{}}$};
\node[] at (6.4,-4.55) {\color{tabred}\small \HybridMH};
\node[] (A) at (2.55,-1.45) {\scriptsize 200 steps};
\node[] (B) at (5.1,-1.45) {\scriptsize 500 steps};
\node[] (C) at (7.65,-1.45) {\scriptsize 2,000 steps};
\node[] (D) at (10.2,-1.45) {\scriptsize 5,000 steps};
\draw[->] (A) -- (B);
\draw[->] (B) -- (C);
\draw[->] (C) -- (D);
%
\node[inner sep=0pt] (a0) at (13.8,1.6) {\includegraphics[width=3.5 cm]{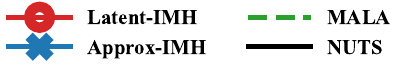}};
\node[inner sep=0pt] (a1) at (0,0) {\includegraphics[width=\width cm]{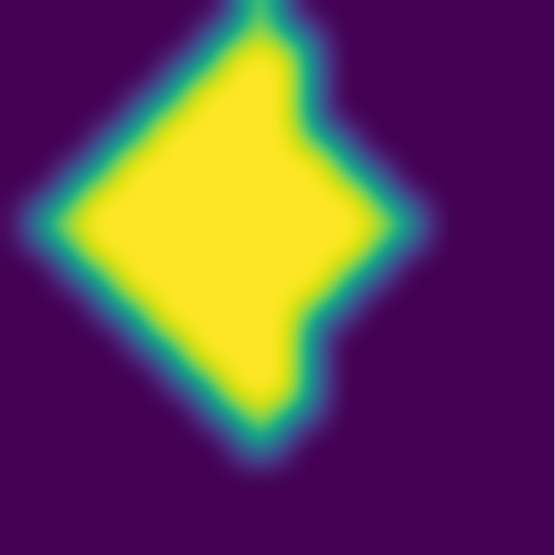}};
\node[inner sep=0pt] (a2) at (2.55,0) {\includegraphics[width=\width cm]{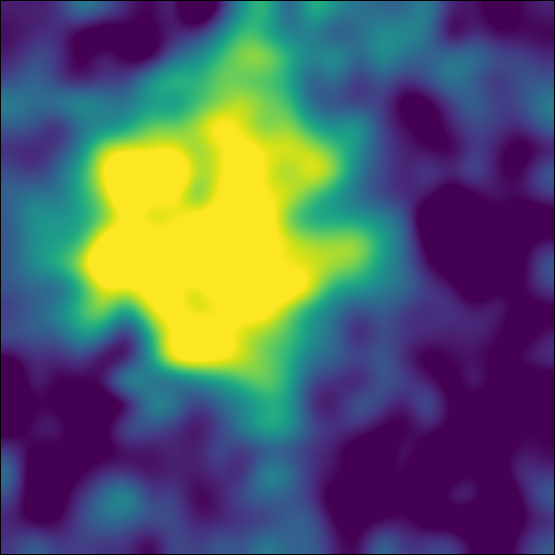}};
\node[inner sep=0pt] (a3) at (5.1,0) {\includegraphics[width=\width cm]{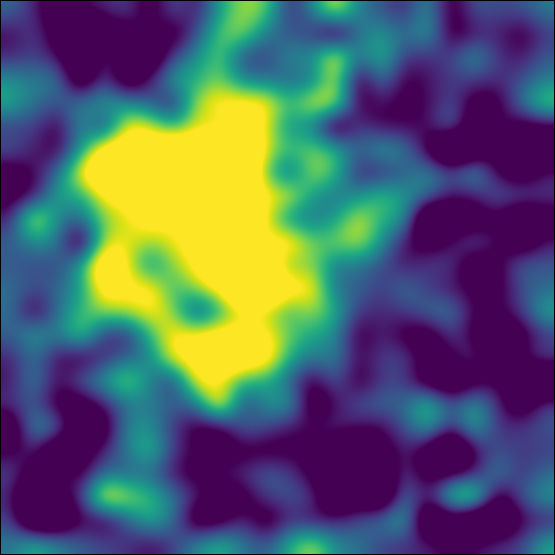}};
\node[inner sep=0pt] (a4) at (7.65,0) {\includegraphics[width=\width cm]{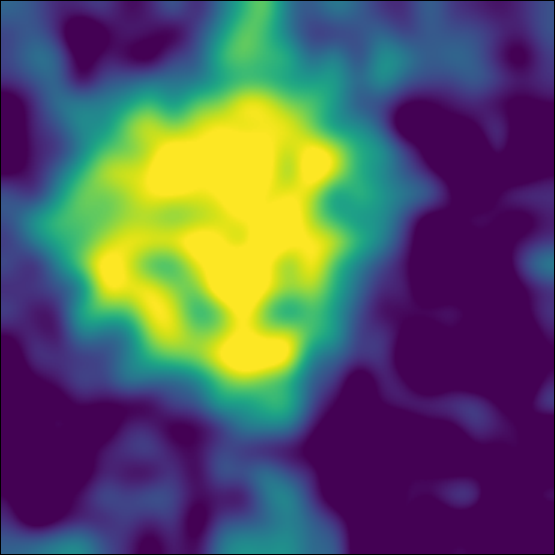}};
\node[inner sep=0pt] (a5) at (10.2,0) {\includegraphics[width=\width cm]{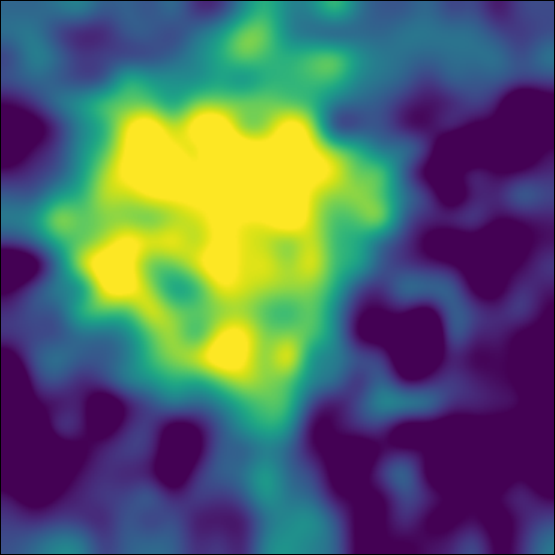}};
\node[inner sep=0pt] (a6) at (13.65,-0.15) {\includegraphics[height=2.8 cm]{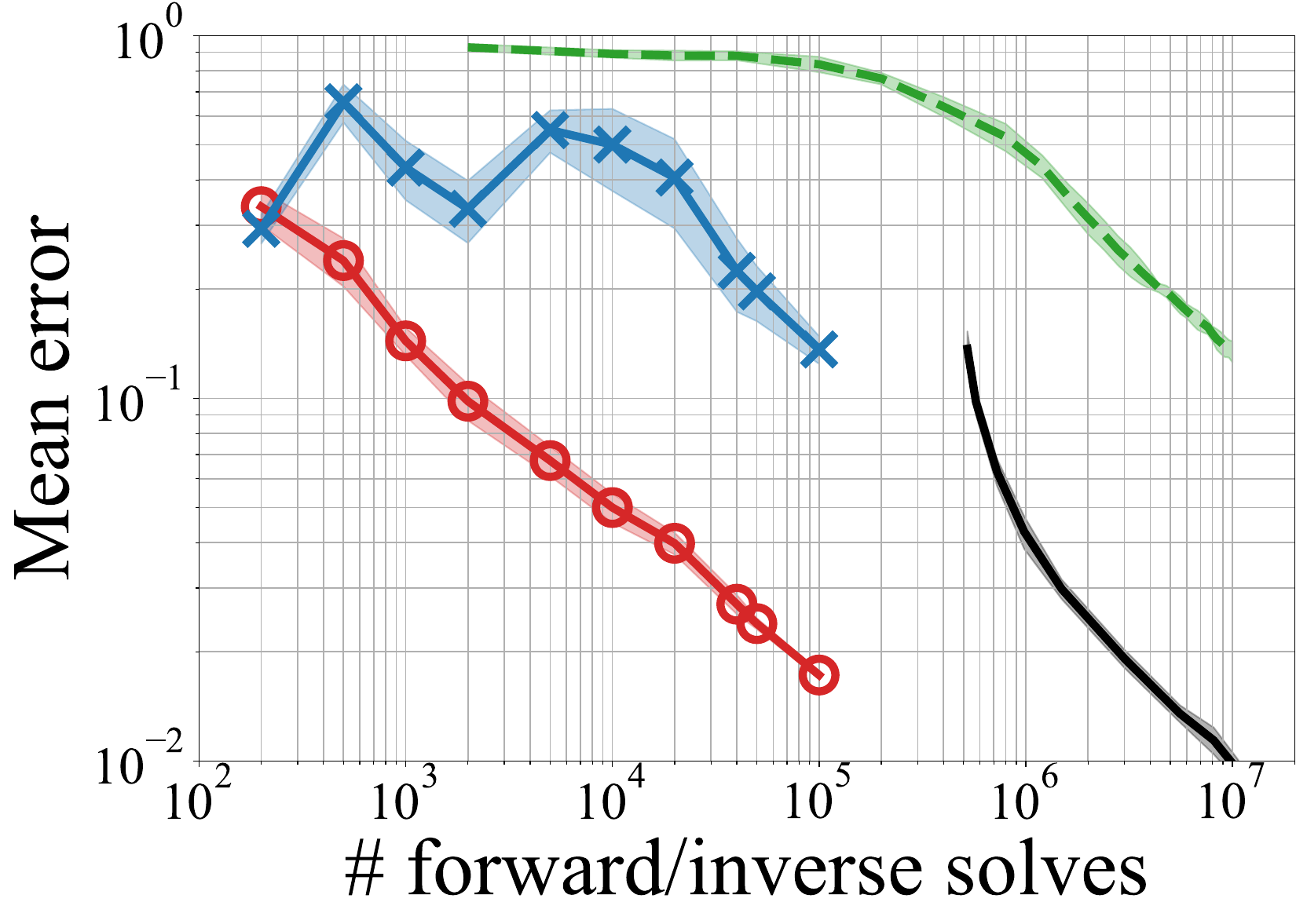}};
\node[inner sep=0pt] (b1) at (0,\vpos) {\includegraphics[width=\width cm]{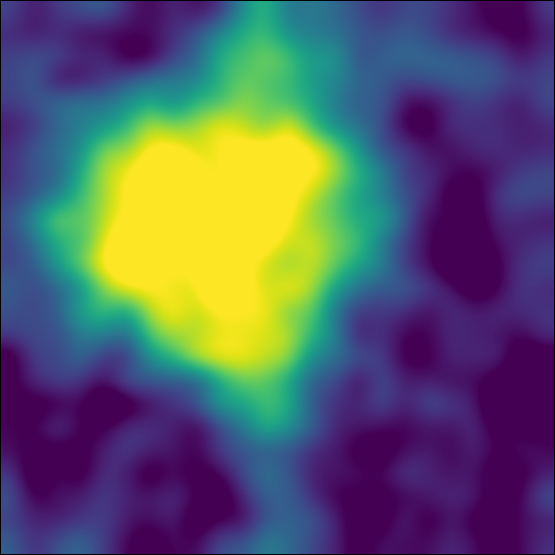}};
\node[inner sep=0pt] (b2) at (2.55,\vpos) {\includegraphics[width=\width cm]{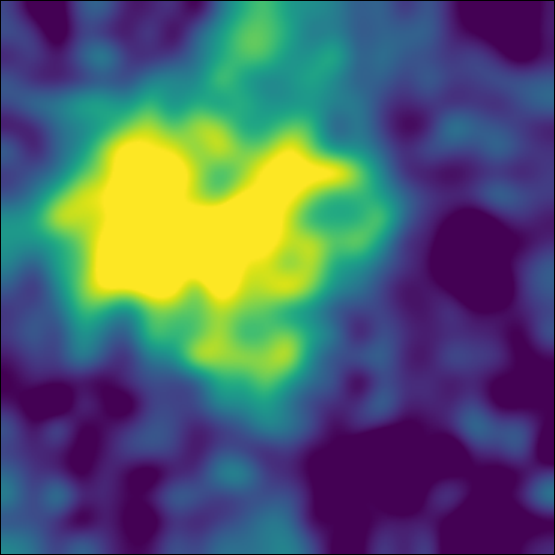}};
\node[inner sep=0pt] (b3) at (5.1,\vpos) {\includegraphics[width=\width cm]{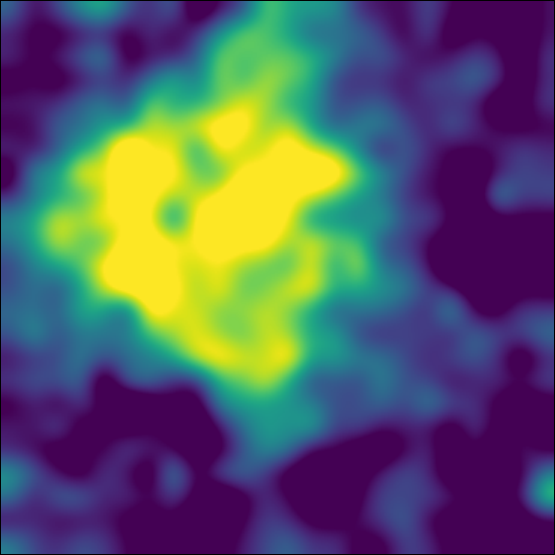}};
\node[inner sep=0pt] (b4) at (7.65,\vpos) {\includegraphics[width=\width cm]{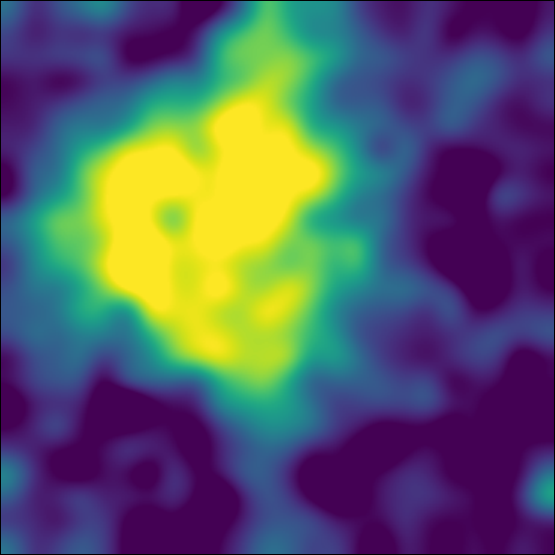}};
\node[inner sep=0pt] (b5) at (10.2,\vpos) {\includegraphics[width=\width cm]{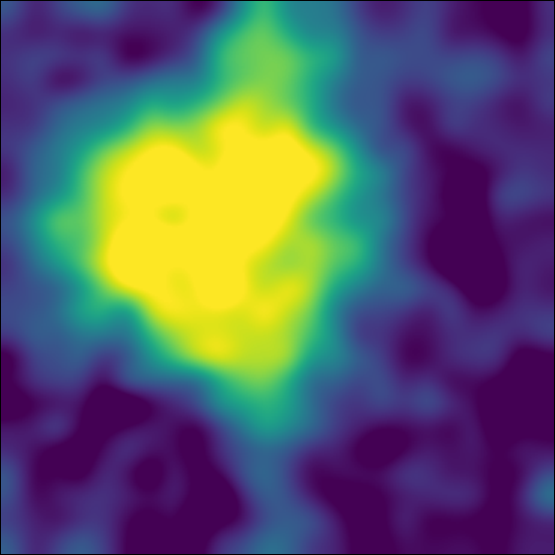}};
\node[inner sep=0pt] (b6) at (13.65,-3.15) {\includegraphics[height=2.8 cm]{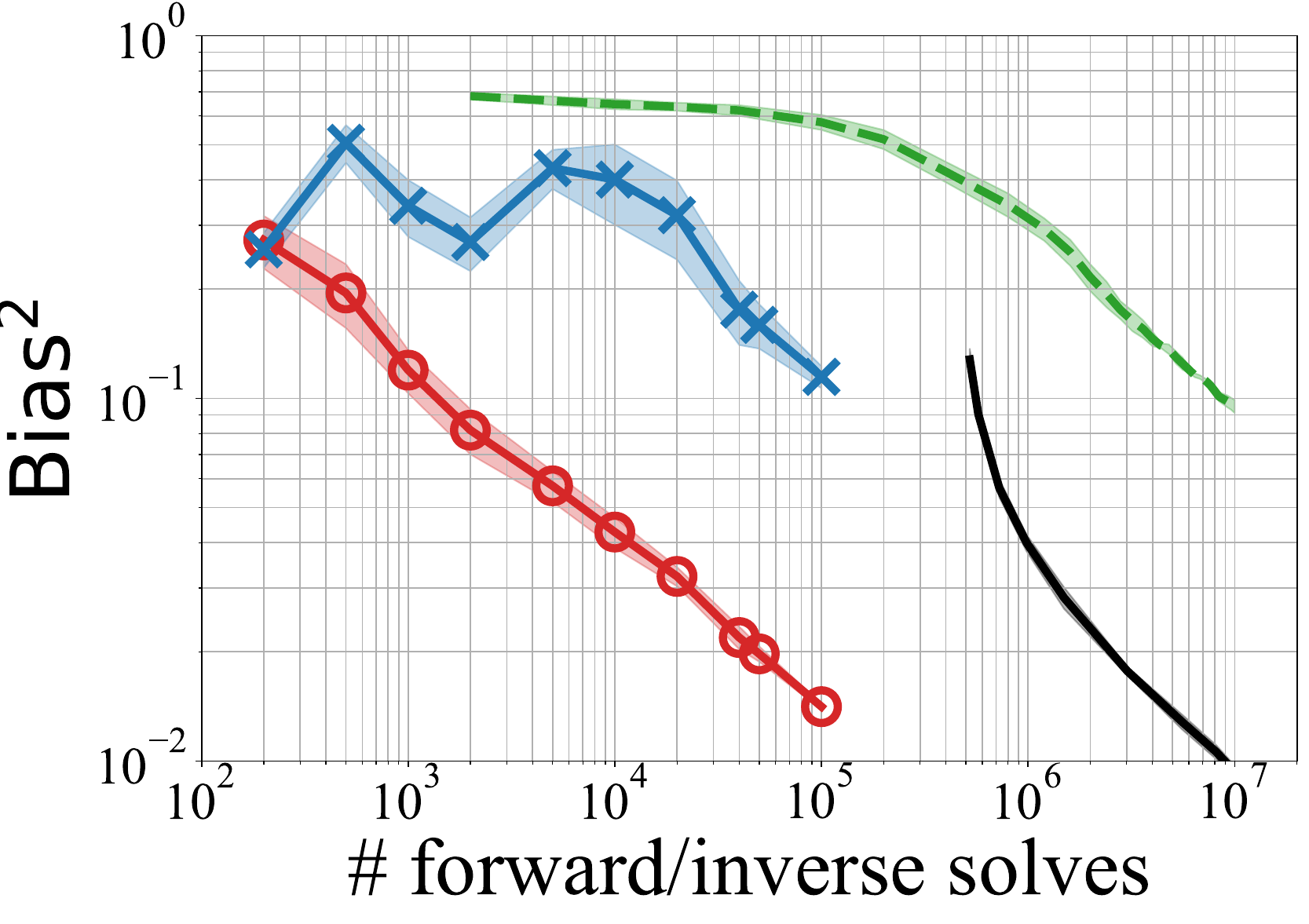}};
\end{tikzpicture}
    \caption{\textbf{Reconstructed acoustic source fields.} We reconstruct the sound source field $x$ from partial observations under a non-Gaussian prior, where posterior inference is only possible via sampling. From left to right: $x_\text{true}$ is the ground truth, and $x_\text{mean}$ is an accurate posterior mean computed from thousands of samples. The \textbf{Approx-IMH} and \textbf{Latent-IMH} rows show posterior averages after 200, 500, 2000, and 5000 MCMC steps. The last column shows the relative mean error and squared bias of the second moment between the true posterior and  sampled estimates as a function of the number of forward $\F$ and inverse $\F^{-1}$ solves. 
    (MALA: Metropolis Adjusted Langevin sampler; NUTS: No-U-Turn Sampler.)
    Latent-IMH achieves high accuracy with substantial computational savings: for 10\% relative mean error, it requires $\sim 10^3$ solves, Approx-IMH $\sim 10^5$, and MALA/NUTS millions of solves.}
    \label{fig:helmholtz}
\end{figure*}
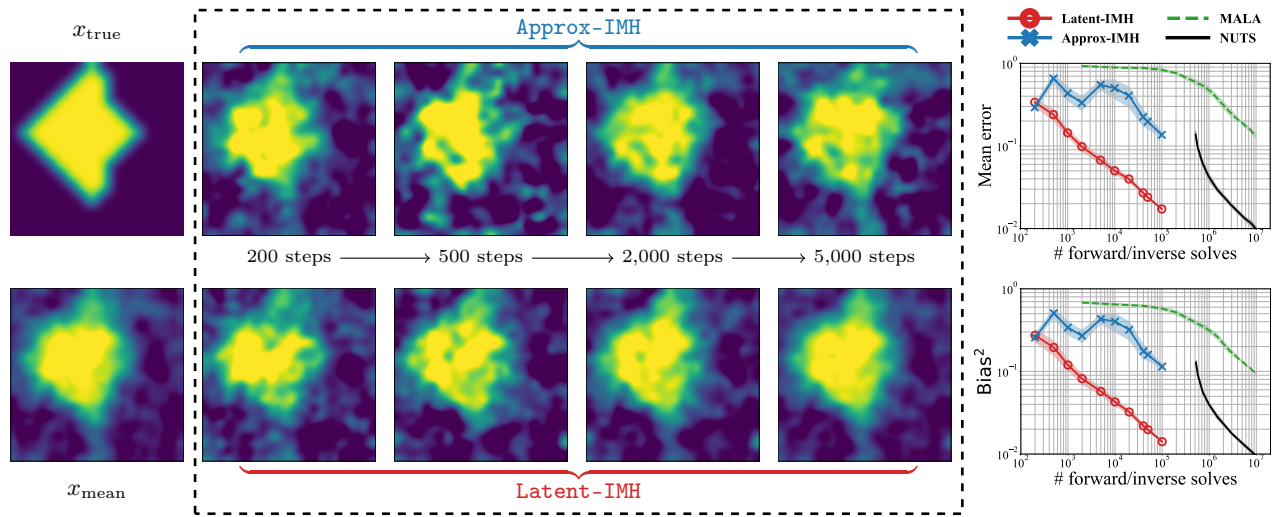

In linear inverse problems, we assume that $y=\A x$, where $x \in \mR^{\dx}$ are the unknown parameters with prior $p(x)$, and $y \in \mR^{\dy}$ are the observed variables corrupted by noise with a known distribution. 
In the Bayesian setting, we wish to sample from a posterior probability distribution $p(x|y)$. The main challenge is that, depending on the prior $p(x)$, the only practical way to sample from the posterior is by a numerical sampling scheme.  Here we present such a scheme for  problems with the following structure:
\begin{subequations}\label{eq:prob}
\begin{flalign}
      u&= \Lo^{-1}\B x,\ \quad u \in \mR^{\du},\ x \in \mR^{\dx},  \label{eq:forward}\\
      y&= \Oo u + e, \ \quad y,e\in\mR^{\dy}, \label{eq:observation}
\end{flalign}
\end{subequations}
with $e$ being noise with known distribution $q(e)$; and thus, $\pi(x|y)\propto q(y-\Oo \Lo^{-1} \B x)\ p(x)$.
Here we introduce the \emph{latent} variable $u$, which in applications is a real but unobserved variable. 
The linear \emph{observation}  operator $\Oo \in \mR^{\dy \times \du}$ is assumed to be full rank. 
$\Lo \in \mR^{\du \times \du}$ is the \emph{latent} operator, a linear, invertible operator related to the physics or dynamics of the underlying problem.
For example, it can be a discretized partial differential operator,  an integral operator, or a graph Laplacian. 
$\B \in \mR^{\du \times \dx}$, the \emph{lifting operator}, is another linear operator that lifts the parameterization of $x$ to the input space of $\Lo$.
We also define the \emph{forward operator} $\F = \Lo^{-1}\B$. 
We can easily see that $\A = \Oo \F$, and equivalently $p(x|y)\propto q(y- \A x)p(x)$.

Bayesian inverse problems that can be formulated using \cref{eq:prob} include
electrical impedance tomography~\citep{kaipio-e00},
acoustic and electromagnetic scattering~\citep{colton-kress98},
exploration geophysics~\citep{bunks-fatimetou-chavent-95,sambridge2002monte},
photoacoustic tomography~\citep{saratoon-arridge-e13},
optical tomography~\citep{arridge-schotland99},  image processing~\citep{chan1998total}, and many others~\citep{book-kaipio2006,tarantola-05,vogel02,ghattas2021learning}.
Inverse problems also appear in graph inference tasks, in which we use partially known vertex features and seek to reconstruct unknown node values or edge weights~\citep{graphs-ortega-moura18,graphs-dong-bronstein20,graphs-mateos2019}.

As a specific example, consider \cref{eq:prob} in the context of time-harmonic acoustics.  In the forward problem, given $x$, we can compute the sound (pressure) $u$ at every spatial point by solving $\Lo u  = \B x$, where $\Lo=k^2 I+\Delta$ is the Helmholtz operator ($k$ is the wavenumber) and $\B x = \sum_i s_i x_i$ with $s_i$ being known spatial profiles of sound sources and $x_i$ their unknown amplitudes. In the inverse problem, we wish to estimate $x_i$ from noisy measurements  $y=\Oo u +e$ of the sound field.

Our proposed sampler is based on the existence of an approximate operator $\Lapprox$ such that $\Lapprox^{-1}\Lo \approx \Id$.
Examples of $\Lapprox$ include incomplete factorizations~\citep{axelsson-94},
 approximate graph Laplacians~\citep{cohen2014solving},
truncated preconditioned Krylov solvers~\citep{van2003iterative},
multigrid solvers and coarse grid representations~\citep{boomeramg00}, approximate hierarchical matrices~\citep{rouet-li-e16},  and many others.
In these examples $\Lapprox^{-1}$  can be $10\times$ or faster than $\Lo^{-1}$.
Using $\Lapprox$, we define $\Fapprox =  \Lapprox^{-1} \B$ and $\Aapprox=\Oo \Fapprox$.

Now, let us return to our main goal, sampling from $\pi(x \mid y)$.
Related work includes Markov chain Monte Carlo (MCMC) methods such as Langevin methods~\citep{girolami2011riemann}, NUTS~\citep{hoffman2014no}, and the Metropolis-adjusted Langevin algorithm~\citep{mala-roberts1996}, as well as generative models~\citep{inverse-song-ermon21}, normalizing flows~\citep{papamakarios2021normalizing}, low-rank approximations~\citep{spantini2015optimal}, and many others~\citep{biegler2010large}. These methods aim to generate samples of $x$ by targeting the posterior density $q(y - \A x)\, p(x)$.

In \newimh{} we generate samples from $u$ and then, using the properties of the forward problem, we convert them into samples of $x$.
This approach is motivated by the following observations. 
\begin{enumerate*}[label=(\arabic*)]
\item We have an approximate operator $\Lapprox$ at our disposal.
\item The observation operator $\Oo$ is computationally inexpensive. 
\item \Cref{eq:forward}, can be used to define an implicit distribution $p(u)$ given $p(x)$.
Then, using precomputation (that is, before ``seeing'' $y$), we can approximate $p(u)$ by $\widetilde{p}(u)$.
\end{enumerate*}

The use of approximate forward models has also been extensively studied in multilevel Monte Carlo and multilevel MCMC. For example, \citep{multimcmc} propose a multilevel MCMC method based on independent Metropolis--Hastings, where chains at different levels are coupled to construct low-variance estimators for the fine posterior. These approaches are complementary to ours: they focus on variance reduction via coupling, whereas we design improved proposal distributions for a single-level Metropolis--Hastings sampler. In particular, we use the approximate operator to construct an independence proposal targeting the exact posterior.

Another related class of methods is delayed-acceptance MCMC (e.g., two-stage or multi-fidelity MCMC~\citep{2stage}), where proposals from an approximate model are screened before evaluating the exact model. While effective, these methods are inherently local, as proposals depend on the current state. In contrast, our approach uses independent proposals, enabling precomputation and parallel evaluation, and improving mixing in high-dimensional settings.

{\bf In this context our contributions are as follows:}
\nbone
With \newimh{} we introduce a proposal distribution in which  we  first sample $u\sim q(y-\Oo u) \widetilde{p}(u)$, and then we propose  $x$ such that $\Lo u = \B x$ (\cref{sec:background}).
\nbtwo
We theoretically analyze \newimh\ and compare it with a reference approximate IMH that uses $\Fapprox$ 
to sample directly from $p(x)$ (\cref{sec:KL} and \cref{sec:mcmc}); 
in particular, we introduce a new theoretical result for the mixing time (\cref{thm:mix-general} and \cref{thm:mix-normal}).
\nbthree
We conduct numerical experiments to demonstrate the effectiveness of the method (\cref{sec:mcmc-experiment}).\\
%

\section{PROBLEM SETTING AND METHODS}\label{sec:background}

\subsection{Problem setting}


Our objective is to sample $x$ from the \Exact $\pi(x|y)$, given $y$:
\begin{equation}\label{eq:exact-post}
  \qquad  \pi(x|y)  \propto q(y-\A x) p(x) = q(y-\Oo \F x) p(x).  
\end{equation}
MCMC methods require repeated evaluations of the forward operator $\F$, which makes them costly.
To address this, our aim is to leverage the approximate operator $\Fapprox $ to facilitate sampling from $\pi(x|y)$.

As mentioned, we would like to sample an approximate distribution $\piapprox(u)$ and then use it to define a unique $x$.
To define such a unique mapping, we first introduce {\bf two assumptions}.
\nbone $\dx \leq \du$; and \nbtwo $\dy \leq \dx$.
The first assumption is satisfied for nearly all applications that have the \cref{eq:prob}'s structure. 
The second assumption is also common, but not always true. 
Under these assumptions, $\F$ is a thin, tall, full-rank matrix.
That means that $p(u)$ is degenerate since it is supported in a lower dimensional subspace of $\mR^{\du}$.
To address this, we introduce a {\bf reparameterization trick} based on the SVD decomposition of $\Oo$ (which in general should be cheap to compute as it does not involve $\F$ of $\Fapprox$. Since $\dy<\dx$, we introduce a pseudo-observation operator to accomplish this. 

Let $\Oo = \U [\b S\vdots \b 0] \V^\top$ be the full SVD of $\Oo$, so that $\V$ is unitary. Let $\V_y$ be the first $\dy$ columns of $\V$, which define the rowspace of $\Oo$. Let $\V = [\V_y, \V_\perp]$. Let $\V_+$ be $\dx - \dy$ columns from $\V_\perp$. 
Define $\V_x = [\V_y, \V_+]$, $\V_x  \in\mathbb{R}^{d_u \times d_x}$. 
Then we define a new observation operator, $\b Z \in\mathbb{R}^{d_y \times d_x}$ by $\b Z= \Oo \V_x$. With these definitions, we can rewrite \cref{eq:prob} as,
\begin{equation}\label{eq:transform}
    \begin{cases}
        u = \V_x^\top \b L^{-1} \B x,\\
        y = \b Z u+ e.
    \end{cases}\qquad
    \begin{cases}
        \F:= \V_x^\top \Lo^{-1} \B,\\
        \Fapprox := \V_x^\top \Lapprox^{-1} \B.
    \end{cases}
\end{equation}
Notice that now the redefined operator $\F$ is square, and {\bf assuming w.l.g.} that $\B$ is full rank, then both $\F$ and $\Fapprox$ are full rank and thus invertible. The question is how to choose $\V_+$. We do to maximize 
conditioning of $\V_x^T \Lo^{-1} \B$. We give an algorithm to do so in the appendix (\cref{sec:transform}).
%
%

\paragraph{Remark:} Given this transformation,  the analysis is done for $\F$  invertible square and square matrix with $\b O$ being  the modified observation operator.
With this understanding, we set $d=\du=\dx$ and denote the observation operator as $\b O\in\mathbb{R}^{d_y \times d}$, so that the observation model is simplified to  $y = \b O \F x + e$.  


\subsubsection{\Approx and \Hybrid}
We consider two approximate posterior distributions based on $\Fapprox$.
These distributions will be used as importance sampling proposals in the IMH algorithm.
Since we expect $\Aapprox$ to be computationally cheaper than $\A$,
a natural approach is to substitute $\A$ in \cref{eq:exact-post} with $\Aapprox$.
This yields the following posterior, which we refer to as \Approx:
\begin{align}\label{eq:approx-post}
    \piapprox(x|y)\propto q(y-\Aapprox x) p(x)
\end{align}
We use this distribution as the baseline as we consider it represents of methods that use $\Aapprox$ to accelerate sampling from $\pi(x|y)$~\cite{martin2012stochastic}.

Now, let us introduce \newimh{}:
\begin{align}\label{eq:hybrid-sample}
    \begin{cases}
        u\sim \piapprox (u|y) \propto q(y-\b O u) \widetilde{p}(u) \\
        x \gets \F^{-1} u ,
    \end{cases}
\end{align}

%
where $\piapprox(u|y)$ is the posterior of $u$ given $y$ when $u$ has the prior distribution $\widetilde{p}(u)$.
It is easy to see that samples from this process follow a new posterior distribution, which we refer to as \Hybrid:
\begin{align}\label{eq:hybrid-post}
     \pinew(x|y)\propto q(y-\A x) \widetilde{p}(\F x). 
\end{align}

Let us now briefly discuss the choice of $\apu{u}$.
One natural choice is to formally define $\apu{u}  = p(\Fapprox^{-1} u) |\det \Fapprox^{-1}|$ when $\Fapprox$ is invertible.  
Alternatively, one can sample $x \sim p(x)$, set $u = \Fapprox x$, and then construct $\apu{u}$ using machine learning techniques such as normalizing flows or variational autoencoders.


\subsection{Independence Metropolis-Hastings}

\begin{algorithm}
\caption{Independence Metropolis-Hastings}
\label{algo:imh}
 \hspace*{\algorithmicindent} \textbf{Input:} observations $y$, number of  MH steps $k$,  proposal distribution $g(x|y)$\\
\hspace*{\algorithmicindent} \textbf{Output:} samples $\{x_i\}_{1}^{k+1}$ from $p(x|y)$
\begin{algorithmic}[1]
\State $x_1 \sim g(x|y)$
\For{$t=1$ to $k$}
\State Propose  $x^\prime \sim  g(x|y)$
\State Compute acceptance ratio $a(x^\prime,x_t) = \min\left\{1, \frac{p(x^\prime|y)}{p(x_t|y)}\, \frac{g(x_t|y)}{g(x^\prime|y)}\right\}$
\State Accept $x^\prime$ with probability $a(x^\prime,x_t)$; otherwise accept $x_t$
\EndFor
\end{algorithmic}
\end{algorithm}

In \Cref{algo:imh}, we recall the IMH algorithm with $g(x|y)$  being the proposal distribution.
Given this template algorithm, we define two variants depending on $g(x|y)$: \nbone\ \ApproxMH, where $g=\piapprox$; or  \nbtwo\ \HybridMH, where  $g = \pinew$.
For these variants, the acceptance ratio becomes
\begin{itemize}[leftmargin=*]
    \item \ \ApproxMH: 
    \begin{align}\label{eq:ar-approx}
       \aapprox(x^\prime, x_t) = \min \left\{1, \frac{q(y-\A x^\prime)/q(y-\Aapprox x^\prime)}{q(y-\A x_t)/q(y-\Aapprox x_t)}\right\}.
    \end{align}
    \item  \HybridMH: 
      \begin{subequations}       \label{eq:ar-hybrid}
        \begin{align}
      \anew(x^\prime, x_t) 
      &= \min \left\{1, \frac{p(x^\prime)/\apu{u^\prime}}{p(x_t)/\apu{u_t}}\right\} \text{  or equivalently,} \label{eq:ar-hybrid-1}   \\
      \anew(x^\prime, x_t)  
      &= \min \left\{1, \frac{p(x^\prime)/p(\Fapprox^{-1} u^\prime)}{p(x_t)/p(\Fapprox^{-1} u_t)}\right\}. \label{eq:ar-hybrid-2} 
      \end{align}
      \end{subequations}
\end{itemize}

Let us make a few remarks.
Note that although both \cref{eq:ar-approx} and \cref{eq:ar-hybrid} are based on $\Fapprox$, their acceptance ratios differ.
Both use the exact forward operator. \Cref{eq:ar-approx} uses $\F$ in the evaluation of $\A$. \Cref{eq:ar-hybrid} uses $\F^{-1}$ to compute $x^\prime = \F^{-1} u$ in the evaluation of $p(x^\prime)$.  Regarding the acceptance ratio of \HybridMH in \cref{eq:ar-hybrid}, note that \eqref{eq:ar-hybrid-1} is derived directly from the \Hybrid expression in \eqref{eq:hybrid-post} and suitable when we have a black box $\apu{u}$. Alternatively, \eqref{eq:ar-hybrid-2} assumes that $\Fapprox$ is a linear invertible operator and the $\det \Fapprox^{-1}$ term cancels out, which makes it more efficient when $\Fapprox^{-1} u$ is inexpensive to compute. Second, notice that, remarkably, the likelihood term involving the noise distribution $q()$ does not appear in \eqref{eq:ar-hybrid}.  It turns out that this property makes it quite accurate when the noise is small. 

\section{APPROX-IMH POSTERIOR VS. LATENT-IMH POSTERIOR}\label{sec:KL}
In this section, we compare \Approx and \Hybrid with \Exact in terms of the expected KL-divergence, where the expectation is taken over the observation $y$.
We assume that both the prior and the noise distributions are Gaussian, as stated in \cref{assume:p-and-q}.
We  derive exact expressions for the KL-divergence in a simplified setting (\cref{prop:kl-simple}) and establish upper bounds for more general cases (\cref{thm:kl-general}).
First we define $\KLapprox$ and $\KLnew$ by
\begin{align}
    &\KLapprox:= 2 \E_y\left[\KL\left(\piapprox(x|y)|| \pi(x|y)\right)\right],\nonumber\\ 
    &\KLnew:= 2 \E_y\left[\KL\left(\pinew(x|y)|| \pi(x|y)
    \right)\right].
\end{align}

\begin{assumption}\label{assume:p-and-q}
The prior $p(x)=\mathcal{N}(\b 0, \b I)$; the noise $q(e) = \mathcal{N}(\b 0, \sigma^2 \b I)$.
\end{assumption}
Given the Gaussian assumptions in \cref{assume:p-and-q}, the posteriors for \texttt{Exact}, \ApproxMH and \HybridMH are all multivariate Gaussian distributions, as summarized in \cref{tab:posterior-gaussian}.  Then $\KLapprox$ and $\KLnew$ can be expressed in a closed form as follows:
\begin{proposition}\label{prop:kl-closedform}
    Assume that \cref{assume:p-and-q} holds. Define $\daapprox := \Aapprox^\dagger - \A^\dagger$ and $\danew := \Anew^\dagger -\A^\dagger$, where $\A^\dagger$,$ \Aapprox^\dagger$, and $\Anew^\dagger$ are defined in \cref{tab:posterior-gaussian}. Let $\|\cdot\|_F$ be the Frobenius norm, then
\begin{align}\label{eq:kl-approx}
  \KLapprox\! &=\! \log\frac{|\bSigma|}{|\bSigma_a|} + \Tr(\bSigma^{-1} \bSigma_a) - d  + \frac{1}{\sigma^2}\|\A \daapprox \A \|_F^2 \nonumber\\
  &\quad+\sigma^2 \|\daapprox \|_F^2+ \| \daapprox \A\|_F^2 +  \|\A \daapprox\|_F^2.
\end{align}
The corresponding  expression for $\KLnew$ is similar, with the substitutions $\bSigma_a\rightarrow \bSigma_l$ and $\daapprox\rightarrow \danew$.
\end{proposition}

\begin{table*}[t]
    \label{tab:posterior-gaussian}
    \caption{Distributions of \texttt{Exact}, \texttt{Approx} and \texttt{Latent} posteriors for normal prior, and normal noise with variance $\sigma^2$  (\cref{assume:p-and-q}).}
\footnotesize
\centering
    \begin{tabular}{ccccc}
\toprule
IMH & Distribution & Mean& Variance & Pseudo-inverse\\\midrule
\texttt{Exact} &$ \pi(x|y)=\mathcal{N}(\mu, \bSigma)$  &$\mu =\A^\dagger y $ &$\bSigma=\b I - \A^\dagger \A$ & $\A^\dagger=\A^\top \left( \A\A^\top + \sigma^2 \b I\right)^{-1}$ \\
\texttt{Approx} &$ \piapprox(x|y)=\mathcal{N}(\mu_a, \bSigma_a)$  &$\mu_a =\Aapprox^\dagger y $ &$\bSigma_a=\b I - \Aapprox^\dagger \Aapprox$ & $\Aapprox^\dagger=\Aapprox^\top \left( \Aapprox\Aapprox^\top + \sigma^2 \b I\right)^{-1}$ \\
\texttt{Latent} &$ \pinew(x|y)=\mathcal{N}(\mu_l, \bSigma_l)$  &$\mu_l =\A_l^\dagger y $ &$\bSigma_l=\F^{-1} \Fapprox\bSigma_a  \Fapprox^\top \F^{-\top} $ & $\Anew^\dagger= \F^{-1} \Fapprox\Aapprox^\dagger$ \\
\bottomrule
    \end{tabular}
\end{table*}

\paragraph{Diagonal $\F$, $\Fapprox$, and $\Oo$.}
To give a more intuitive understanding of the differences between $\pi_a$ and $\pi_l$, 
consider the case where
  (1) $\F$ and $\Fapprox$ are diagonal, with $\F_{ii} = s_{i}$ and $\Fapprox_{ii}= \alpha_i s_i$, $i \in [d]$.
  (2) $\Oo = [\Id \ \ \b 0]$. 
Then, the respective expressions for $\KLapprox$ and $\KLnew$ are given in \cref{prop:kl-simple}. 
\begin{proposition}\label{prop:kl-simple}
  Under \cref{assume:p-and-q}, diagonal $\F$, $\Fapprox$, $\Oo$,
  and defining $\rho_i = \nicefrac{\alpha_i^2 s_i^2 + \sigma^2}{s_i^2 + \sigma^2}$, $\zeta_i = \nicefrac{1}{(\alpha_i^2 s_i^2 + \sigma^2)^2} $,
   we obtain the following expressions for expected KL-divergence:
\begin{align}
    \KLapprox&= -d_y + \sum_{i\in[d_y]}\log \rho_i + \frac{1}{\rho_i}  \nonumber\\
   &\quad +\sum_{i\in[d_y]}\zeta_i (\alpha_i - 1)^2(\alpha_i s_i^2 - \sigma^2)^2\frac{s_i^2}{\sigma^2}, \label{eq:kl-simple-approx}\\
    \KLnew &= -d + \sum_{i=d_y+1}^d \alpha_i^2 -\log \alpha_i^2\nonumber\\
  & \quad +\sum_{i\in[d_y]}\log \frac{\rho_i}{\alpha_i^2} + \frac{\alpha_i^2}{\rho_i}+  \zeta_i (\alpha_i^2 - 1)^2 s_i^2 \sigma^2. \label{eq:kl-simple-new}
\end{align}

\end{proposition}
It is easy to observe that both $\KLapprox$ and $\KLnew$ increase with increasing spectrum perturbation $|\alpha_i|$ and the dimension $d$; $\KLapprox$ also increases as the number of observations $d_y$ increases. If we denote the scale of the perturbation by $\delta =|\alpha_i -1|$, the last term of $\KLnew$ in \eqref{eq:kl-simple-new} scales as $\mathcal{O}(\delta\sigma^2/s_i^2)$, which is significantly smaller than the corresponding term in $\KLapprox$ (scaling as $\mathcal{O}(\delta^2 s_i^2/\sigma^2)$), especially when the signal strength $s_i$ is large.

\paragraph{General case.} In the more general setting where both $\F$ and $\Fapprox$ are symmetric, we assume that $\Fapprox$ closely approximates $\F$ in both its eigenvalues and eigenvectors. Under this assumption, we derive upper bounds for $\KLapprox$ and $\KLnew$ in \cref{thm:kl-general} in the appendix. These upper bounds reveal similar conclusion as observed in the diagonal case.



\begin{figure}[t]
\small
    \centering
\begin{tikzpicture}
\node[inner sep=0pt] (a1) at (0,0) {\includegraphics[width=3.5cm]{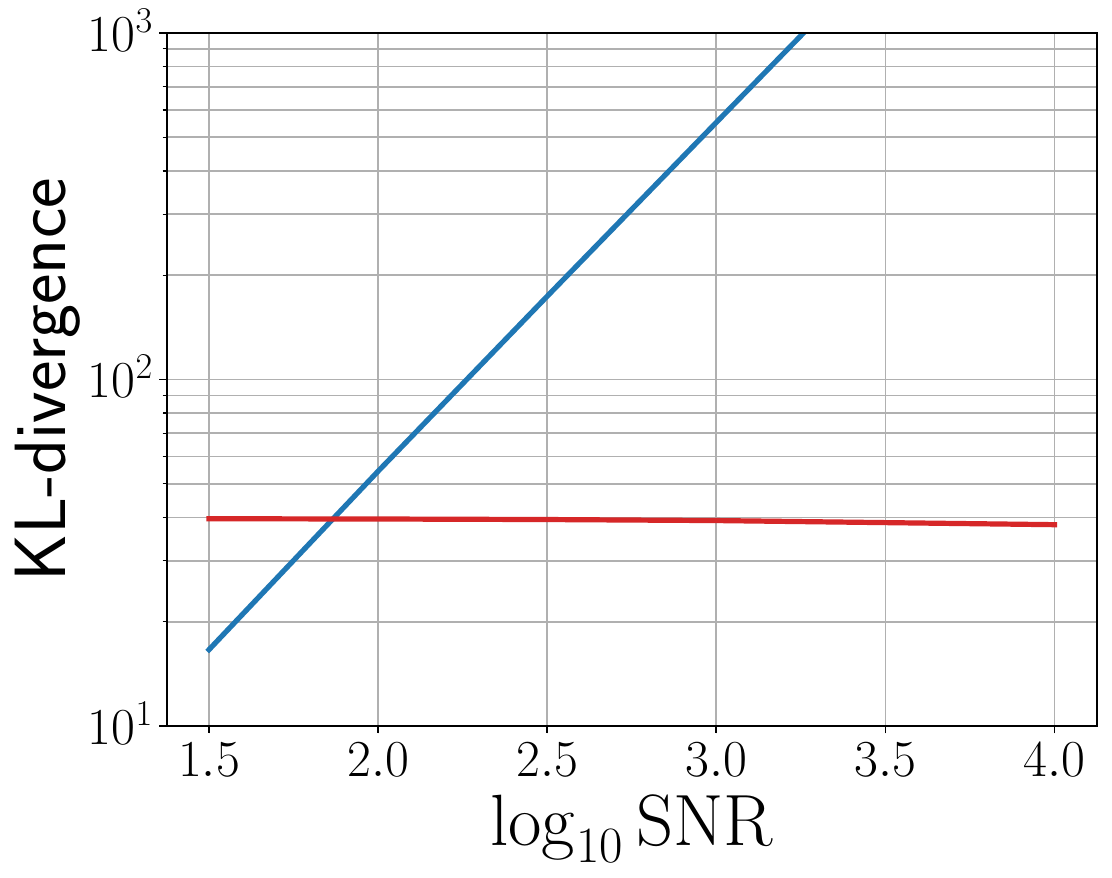}};
\node[inner sep=0pt] (b1) at (3.7,-0.13) {\includegraphics[width=3.5cm]{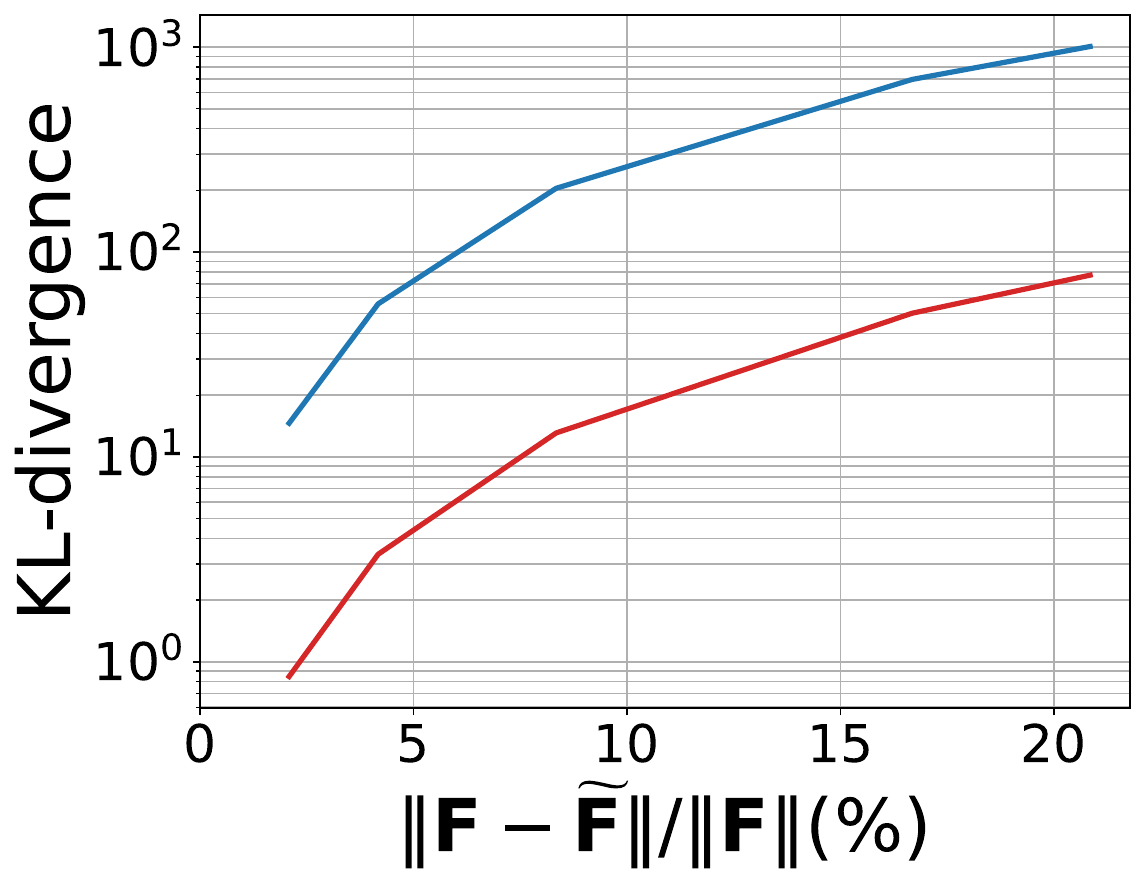}};
\node[inner sep=0pt] (c1) at (0,-3.3) {\includegraphics[width=3.5cm]{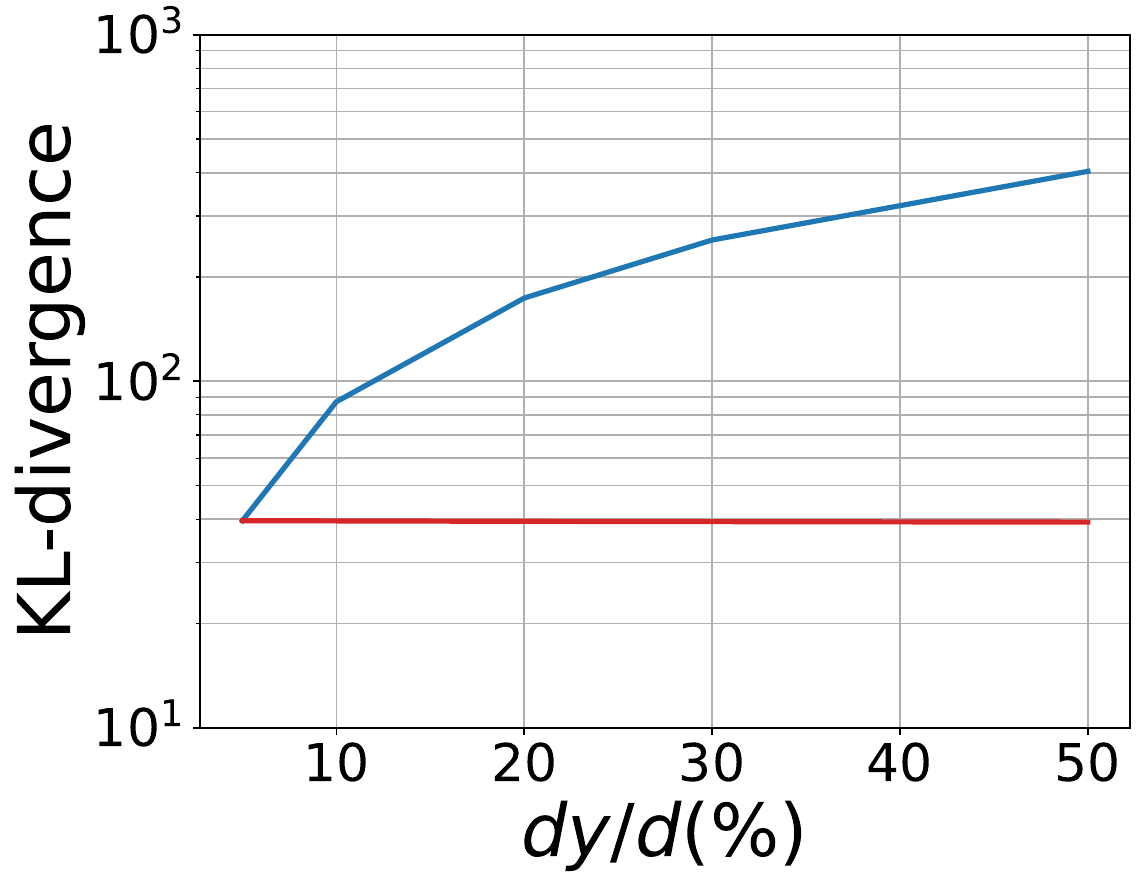}};
\node[inner sep=0pt] (d1) at (3.7,-3.3) {\includegraphics[width=3.5cm]{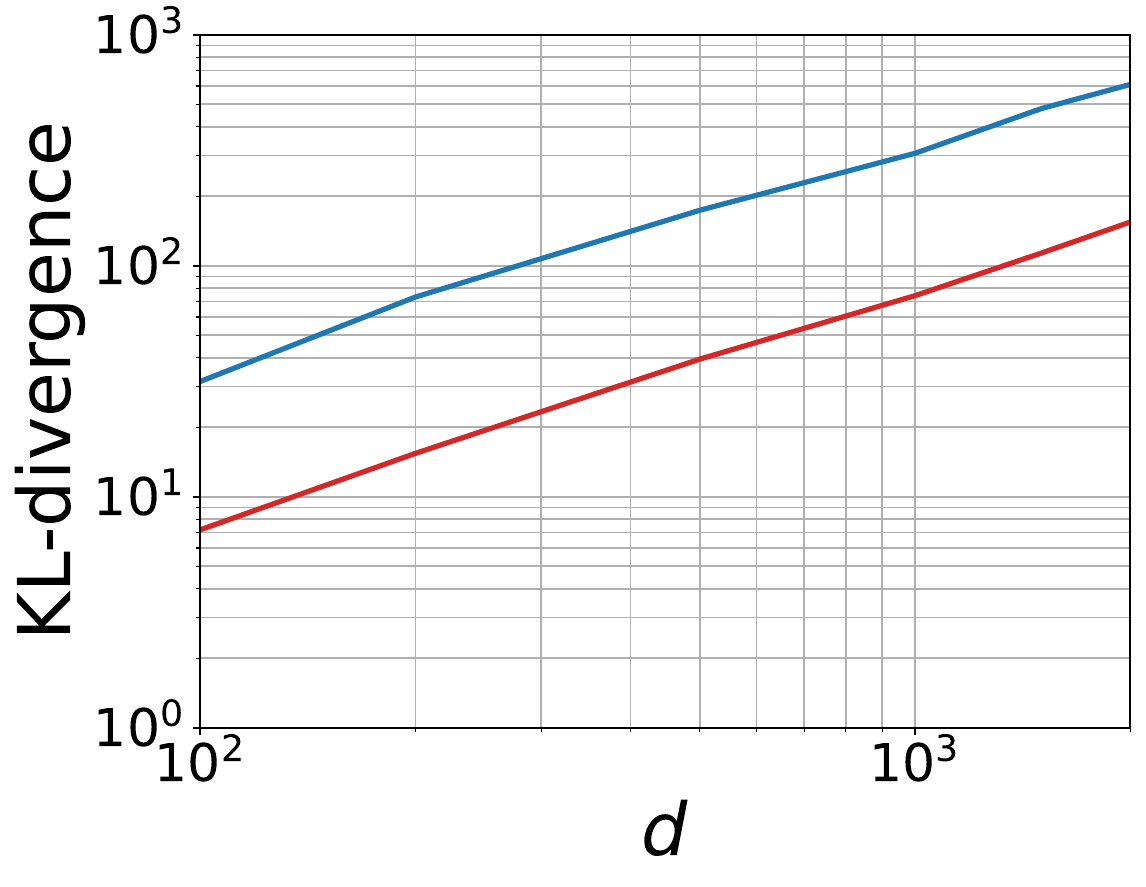}};
\node[draw,rounded corners,fill=black!10] at (0.1, 1.6) {\scriptsize Noise level};
\node[draw,rounded corners,fill=black!10] at (3.8, 1.6) {\scriptsize Spectral error};
\node[draw,rounded corners,fill=black!10] at (0.2, -1.7) {\scriptsize Observation ratio};
\node[draw,rounded corners,fill=black!10] at (3.8, -1.7) {\scriptsize Dimension};
\draw[tabblue, thick] (-.5,2.7) -- (.3,2.7);
\node[] at (2.5, 2.7) {\Approx ($\KLapprox$)} ;
\draw[tabred, thick] (-.5,2.2) -- (.3,2.2);
\node[] at (2.5, 2.2) {\Hybrid ($\KLnew$)} ;
\end{tikzpicture}
    \caption{Sensitivity test results for the expected KL-divergence of \Approx and \Hybrid relative to \Exact.}
    \label{fig:kl-experiments}
\end{figure}


\paragraph{Sensitivity to operator scalings.}
To further demonstrate the difference between $\KLapprox$  and $\KLnew$, we perform numerical experiments with synthetic data.
We adopt the distributions specified in \cref{assume:p-and-q} for prior and noise. Define $\F = \V \b S\V^\top$ and $\Fapprox = \V \widetilde{\b S}\V^\top$, where $\V$ is unitary derived from eigendecomposition of some random matrix, $\b S$ as a diagonal matrix with entries $\{1 /i^2 \}_{i\in[d]}$ and  $\widetilde{\b S}$ be the diagonal matrix with diagonals of $\{\alpha_i /i^2 \}_{i\in[d]}$ ($\alpha_i$ is drawn uniformly from a range around 1). We report both the expected KL divergence and the value of this divergence over the KL divergence between the prior and \Exact. 
We compare $\KLapprox$ and $\KLnew$ by examining the impact of four factors: noise level, spectral error (between $\Fapprox$ and $\F$), observation ratio ($d_y/d$), and problem dimension ($d$). To quantify the noise level, we use the signal-to-noise ratio (SNR), defined as $\mathrm{SNR} := \E\|y\|^2/\E\|e\|^2$.

The complete experimental design is summarized in \cref{tab:kl-experiments} (see \cref{sec:add-experiments}), with the corresponding absolute KL-divergence results presented in \cref{fig:kl-experiments} and relative values shown in \cref{fig:kl-experiments-relative} (see \cref{sec:add-experiments}). Our empirical findings align well with the theoretical analysis: both $\KLapprox$ and $\KLnew$ increase with larger spectral errors and higher problem dimensions. However, $\KLapprox$ exhibits high sensitivity to noise levels and observation ratios, whereas $\KLnew$ demonstrates greater robustness to these factors.

\section{APPROX-IMH VS. LATENT-IMH}\label{sec:mcmc}

As discussed in the previous section, the KL divergence for \Approx and \Hybrid can vary depending on factors like noise level, observation ratio, spectral error, and problem dimensionality. These differences are also likely to have an impact on the performance of \ApproxMH and \HybridMH. To explore this, we first establish general mixing time bounds for both methods in \cref{thm:mix-general}. We then apply these bounds to a simplified diagonal case in \cref{thm:mix-normal} to directly compare their mixing rates.


Our analysis focuses on the log-concave setting, as specified in~\cref{assume:log-concave}. If the noise $q(e)$ is Gaussian, a strongly log-concave prior $p(x)$ is sufficient to satisfy this assumption.

\begin{assumption}\label{assume:log-concave}
    The exact posterior $\pi(x|y)$ is strongly log-concave, i.e., $-\log \pi(x|y)$ is $m$-strongly convex on $\mathbb{R}^d$.
\end{assumption}
We denote the modes of the exact and approximate posteriors as follows:
\begin{align}
       & x^* := \arg\max \pi(x|y),\quad x_a^* :=\arg\max \piapprox(x|y), \notag \\
        & x_l^* :=\arg\max \pinew(x|y) .\nonumber
\end{align}
We define the log-weight functions between the exact posterior with the approximate proposals by
\begin{align}
        \wapprox(x):=\log\frac{\pi(x|y)}{\piapprox(x|y)},\qquad \wnew(x) :=\log\frac{\pi(x|y)}{\pinew(x|y)}.\nonumber
\end{align}
We assume the log-weight functions are locally Lipschitz, as formalized in \cref{assume:lipschitz}. 
\begin{assumption}\label{assume:lipschitz}
  Log-weight functions are locally Lipschitz. For any $R>0$, there exist constants $C_a(R) \geq 0$ such that for all $(x,x') \in \mathsf{Ball}(x^*,R)$,  $\|\wapprox(x) - \wapprox(x')\|\leq C_a(R)\|x-x'\|$.
    For the function $\wnew(x)$, we assume that this condition holds with constant $C_l(R)$.
   
\end{assumption}

The mixing time $\mixapprox(\epsilon)$ for \ApproxMH is defined as the minimum number of steps needed for the Markov chain to reach a total variation (TV) distance below $\epsilon$ from the target distribution $\pi(x|y)$. That is,
\begin{align}
    \mixapprox (\epsilon) := \inf \{n\in\mathbb{N}: \|g P_a^n -\pi \|_{\text{TV}}\leq \epsilon\},
\end{align}
where $P_a$ is the IMH transition kernel with $\piapprox(x|y)$ as proposal,
and $\| f(x) \|_{\text{TV}}:=\nicefrac{1}{2}\int_x |f|$.
The mixing time for \HybridMH, denoted by $\mixnew(g, \epsilon)$,  is defined similarly using $P_l$.

In \cref{thm:mix-general}, we establish mixing time bounds for both \ApproxMH and \HybridMH under the log-concavity and local Lipschitz assumptions.
Following \cite{log-concave1,log-concave2}, we prove these results in \cref{sec:mcmc-proof}.

\begin{theorem}\label{thm:mix-general}
 Let $\epsilon\in(0,1)$. 
 Assume that $\pi_a(x)$ and $\pi_l(x)$ are both $\beta-$warm start with respect to $\pi(x|y)$  (i.e., for any Borel set $\mathcal{E}$, $\pi_a(\mathcal{E}) \leq \beta \pi(\mathcal{E})$).
 Assume that \cref{assume:log-concave} and \cref{assume:lipschitz} hold. For \cref{assume:lipschitz}, suppose that the Lipschitz conditions hold with $C_a({R_a}) \leq\log 2 \sqrt{m}/32 $ and $C_l({R_l}) \leq\log 2 \sqrt{m}/32 $, where 
 \begin{align}
      R_a &= \max\Biggl(\sqrt{\frac{d}{m}}r\left(\frac{\epsilon}{17\beta}\right), \notag\\
      &\quad \sqrt{\frac{d}{m}}r\left(\frac{\epsilon}{272\beta}\right)+ \|x^* - \xapprox^*\|\Biggr)
 \end{align}
 \begin{align}
     R_l &= \max\Biggl(\sqrt{\frac{d}{m}}r\left(\frac{\epsilon}{17\beta}\right), \notag \\
    &\quad \sqrt{\frac{d}{m}}r\left(\frac{\epsilon}{272\beta}\right)\frac{1}{\sigma_{\min}(\Fapprox^{-1} \F)}+ \|x^* - \xnew^*\|\Biggr),
 \end{align}
 
    where  $r(\cdot)$ is a constant  defined in \cref{sec:mcmc-proof}, $\sigma_{\min}(\cdot)$ is the smallest singular value of the matrix.
    Then the mixing time for \ApproxMH has
    \begin{align}\label{eq:mix-approx}
       \mixapprox(\epsilon) \leq 128 \log \left(\frac{2\beta}{\epsilon}\right)\max\left(1,\frac{128^2 C_a^2(R_a)}{(\log 2)^2 m}\right).
    \end{align}
The mixing time for \HybridMH, $\mixnew(\epsilon)$,  satisfies a similar bound with $C_a(R_a)$  replaced by $C_l(R_l)$.
\end{theorem}


\paragraph{Mixing time in the diagonal case.} We now use \cref{thm:mix-general} to compare \ApproxMH and \HybridMH in a simple case, where we assume prior and noise satisfying \cref{assume:p-and-q} and $\F$, $\Fapprox$, $\Oo$ are all diagonal as in \cref{prop:kl-simple}. 
\begin{theorem}\label{thm:mix-normal}
  The definitions and assumptions in \cref{thm:mix-general} and \cref{assume:p-and-q} are true.
  Then for large $d$ the mixing times for \ApproxMH and \HybridMH scale as
    \begin{align}
     & \mixapprox(\epsilon) \sim  \frac{d}{m^2} \max_{i\in[d_y]} \left(1-\alpha_i^2\right)^2 \frac{s_i^4}{\sigma^4},  \notag\\
      &\mixnew(\epsilon) \sim \frac{d}{m^2}  \max_{i\in[d]}\left(1-\frac{1}{\alpha_i^2}\right)^2\!\!.
     \end{align}
\end{theorem}

By \cref{thm:mix-normal}, when the perturbation $\alpha_i$ is fixed, the mixing time for \ApproxMH scales as $\mathcal{O}(d^2 d_y^2 \|\A\|^4/\sigma^4)$, while the mixing time for \HybridMH scales as $\mathcal{O}(d^2)$. This implies that \ApproxMH can exhibit significantly longer mixing times when the noise level is low (i.e., $\|\A\|/\sigma$ is large) or when the proportion of observations is high  (i.e., $\dy/d$ is large).


\begin{figure}[h]
\small
    \centering
\begin{tikzpicture}
\node[inner sep=0pt] (a1) at (0,0) {\includegraphics[width=3.5cm]{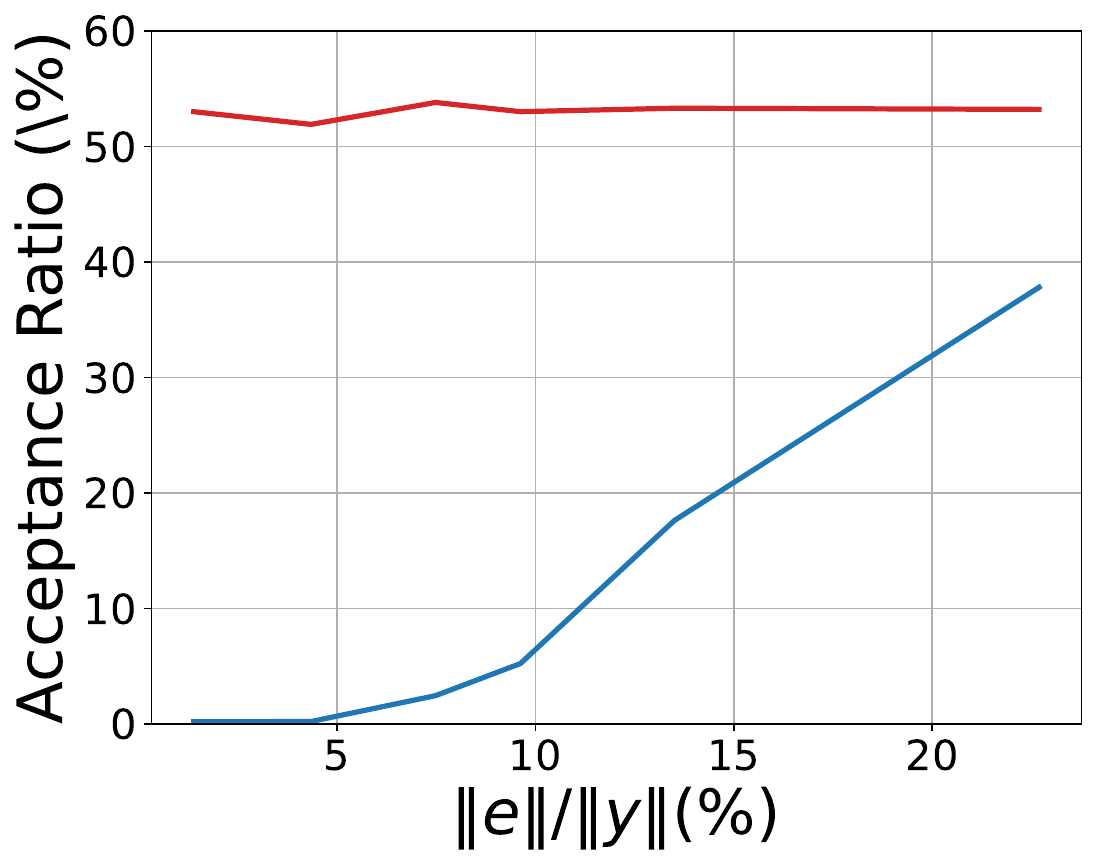}};
\node[inner sep=0pt] (b2) at (3.7,0) {\includegraphics[width=3.5cm]{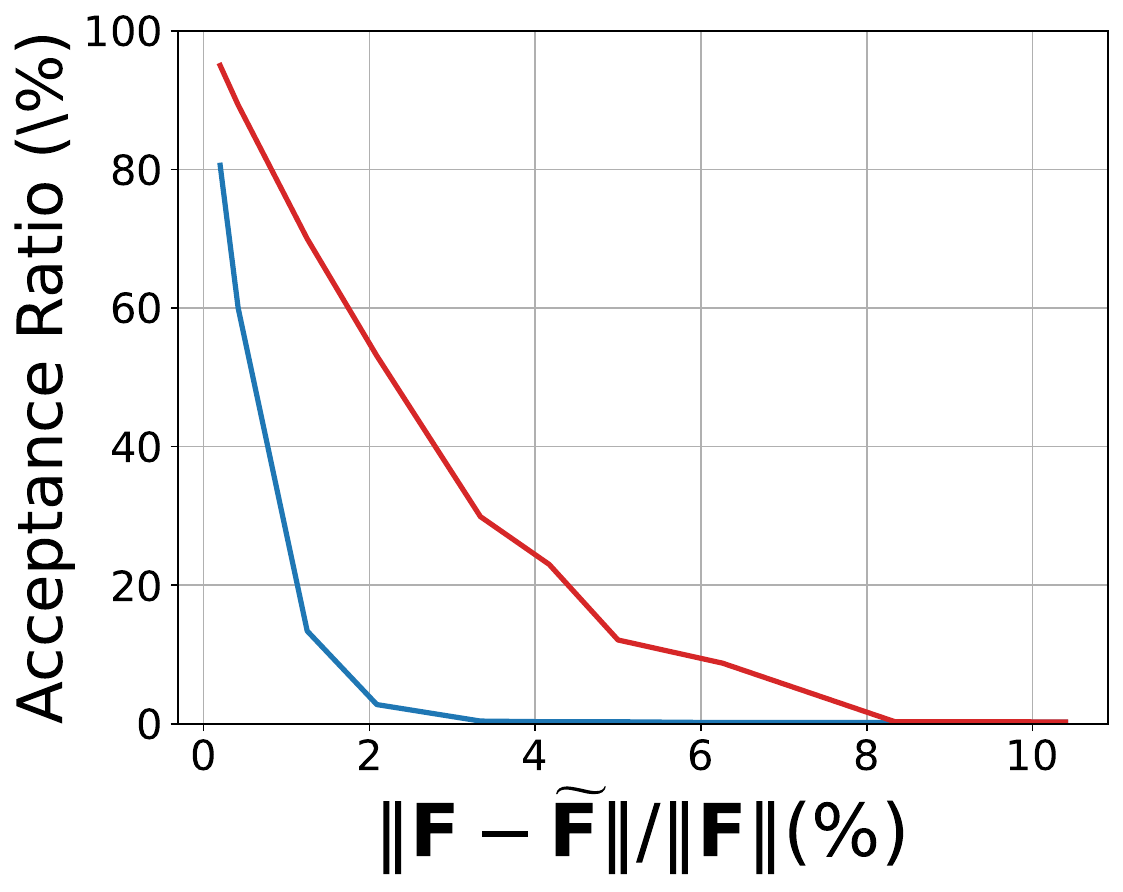}};
\node[inner sep=0pt] (c1) at (0,-3.5) {\includegraphics[width=3.5cm]{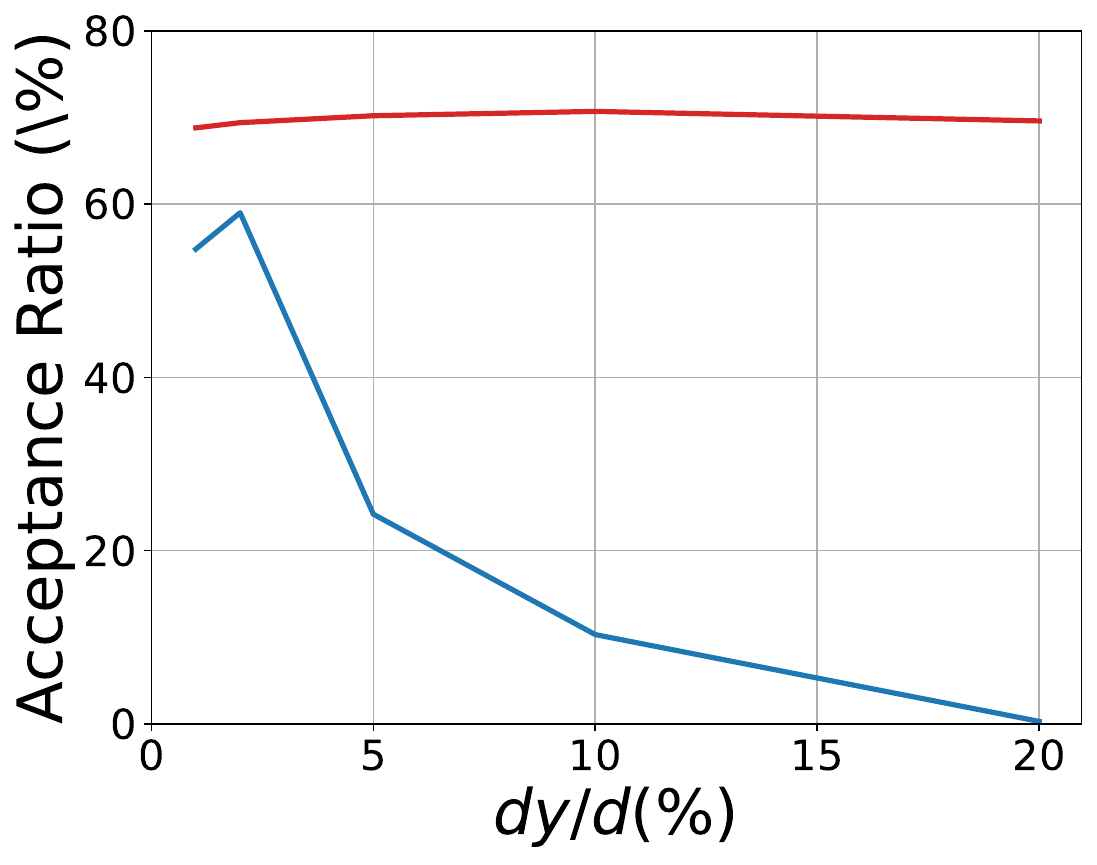}};
\node[inner sep=0pt] (d1) at (3.7,-3.5) {\includegraphics[width=3.6cm]{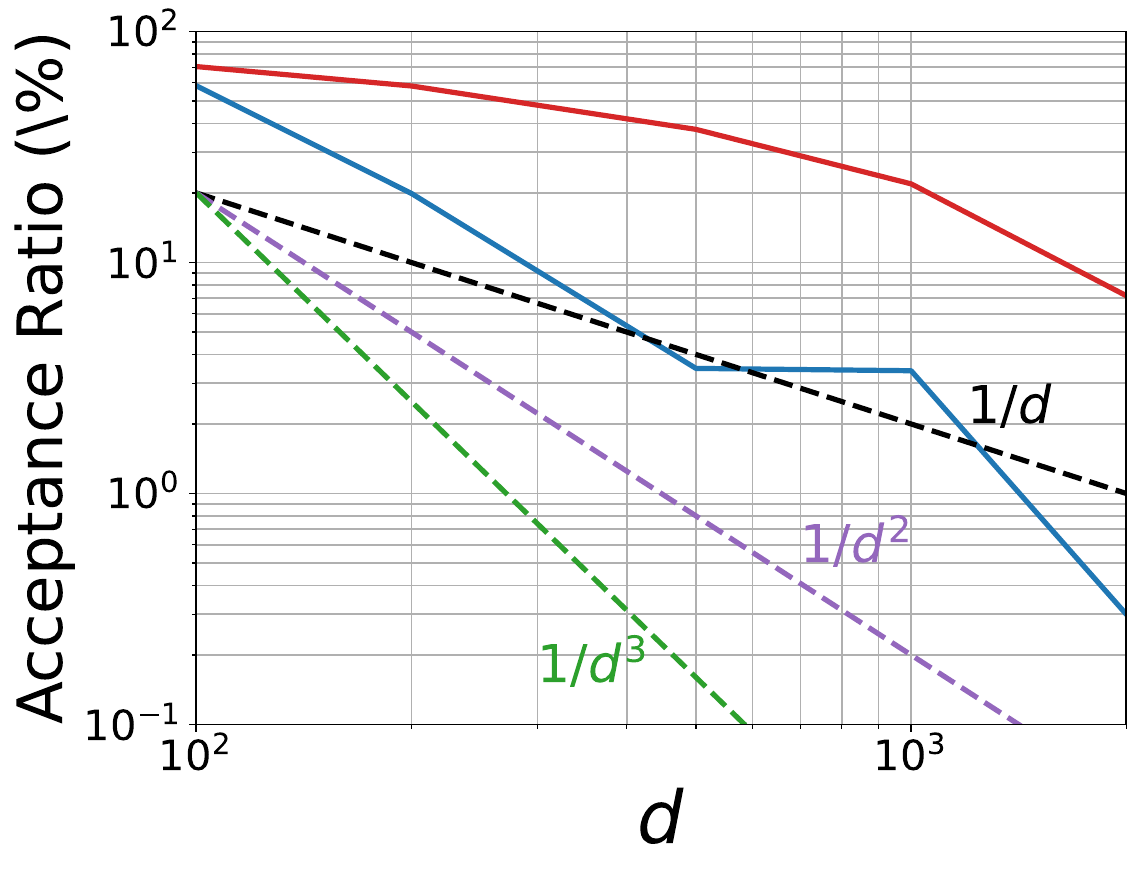}};
\node[draw,rounded corners,fill=black!10] at (.1, 1.7) {\scriptsize Noise level};
\node[draw,rounded corners,fill=black!10] at (3.8, 1.7) {\scriptsize Spectral error};
\node[draw,rounded corners,fill=black!10] at (0.1, -1.7) {\scriptsize Observation ratio};
\node[draw,rounded corners,fill=black!10] at (3.8, -1.7) {\scriptsize Dimension $d$};
\draw[tabblue, thick] (-1,2.2) -- (-.2,2.2);
\node[] at (.7, 2.2) {\ApproxMH} ;
\draw[tabred, thick] (2.5,2.2) -- (3.3,2.2);
\node[] at (4.2, 2.2) {\HybridMH} ;
\end{tikzpicture}
    \caption{Acceptance ratio of \ApproxMH and \HybridMH for different problem scenarios.}
    \label{fig:mcmc-sensitivity}
\end{figure}
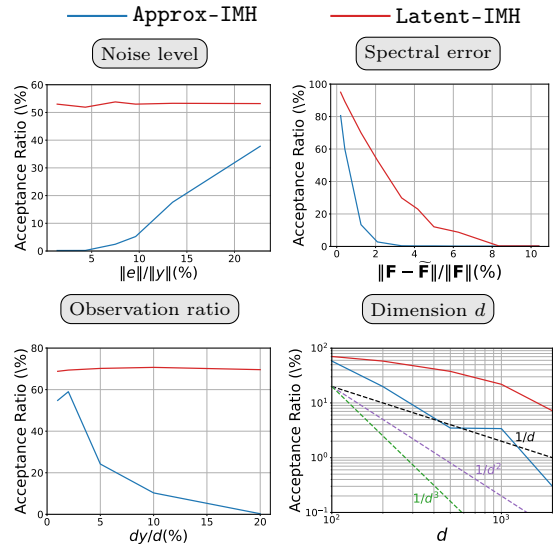
\section{NUMERICAL EXPERIMENTS} \label{sec:mcmc-experiment}

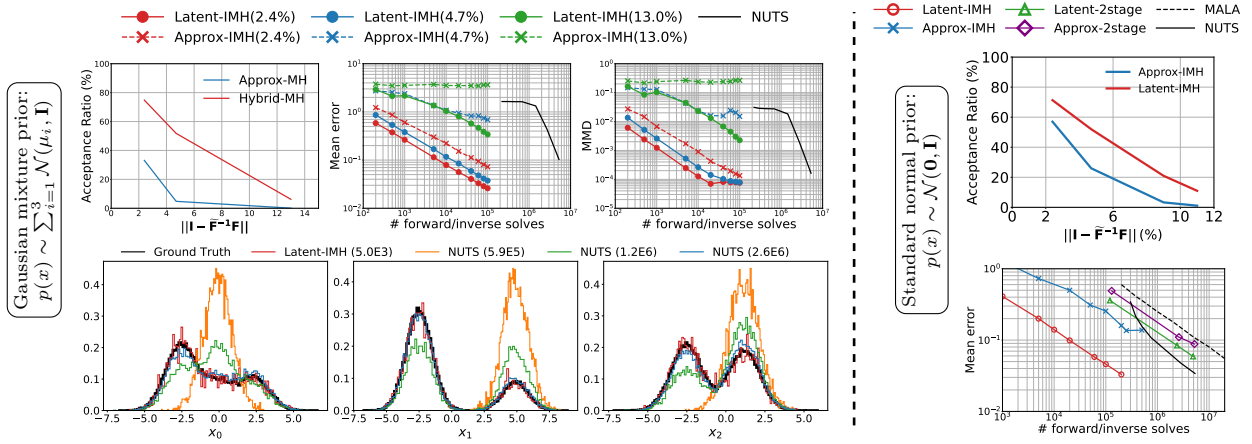
\begin{figure*}[t]
\scriptsize
    \centering
\begin{tikzpicture}
\node[inner sep=0pt] (a0) at (0,-3.7) {\includegraphics[width=9cm]{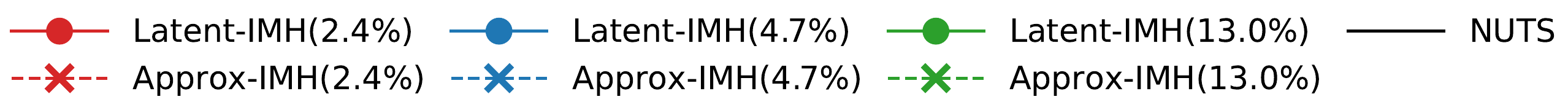}};
\node[inner sep=0pt] (a1) at (0,-5.3) {\includegraphics[width=10cm]{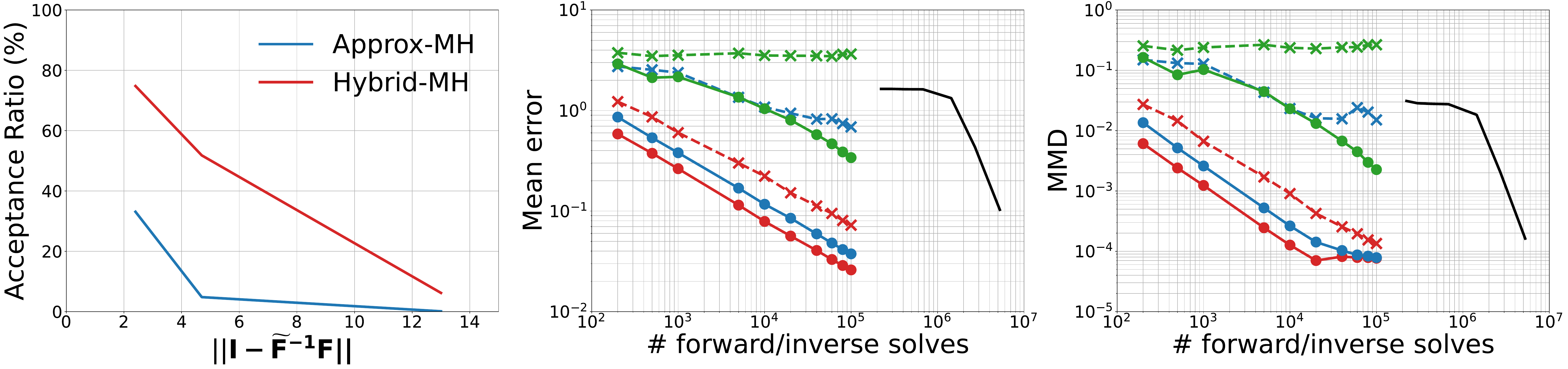}};
\node[inner sep=0pt] (a2) at (0,-6.7) 
{\includegraphics[width=9cm]{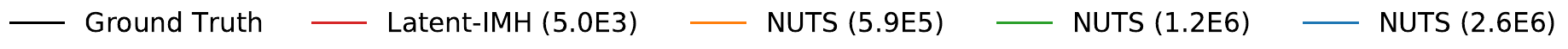}};
\node[inner sep=0pt] (a3) at (0,-8) {\includegraphics[width=10cm]{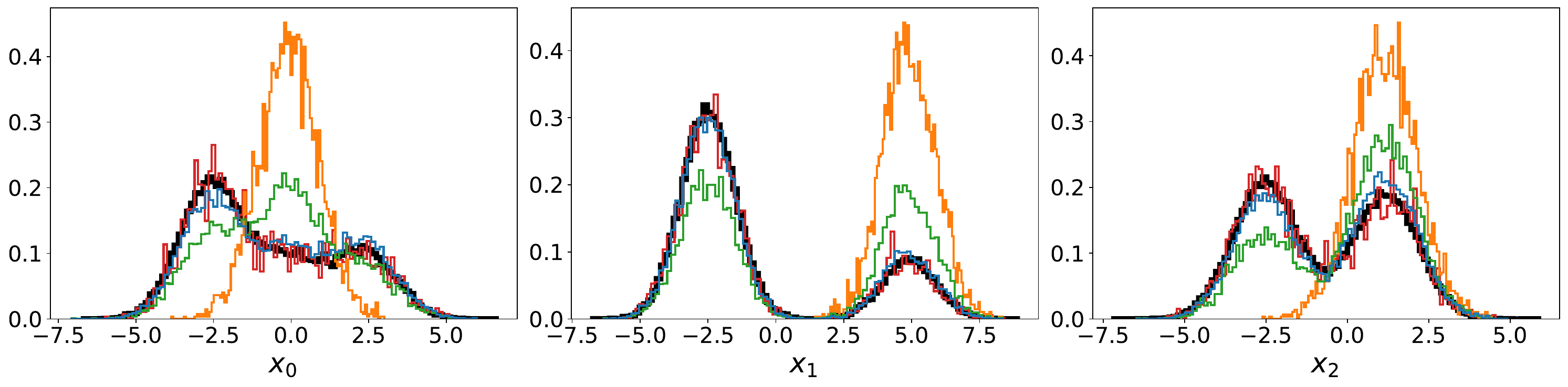}};
\node[inner sep=0pt] (c0) at (8.5,-5.3) {\includegraphics[width=3.5cm]{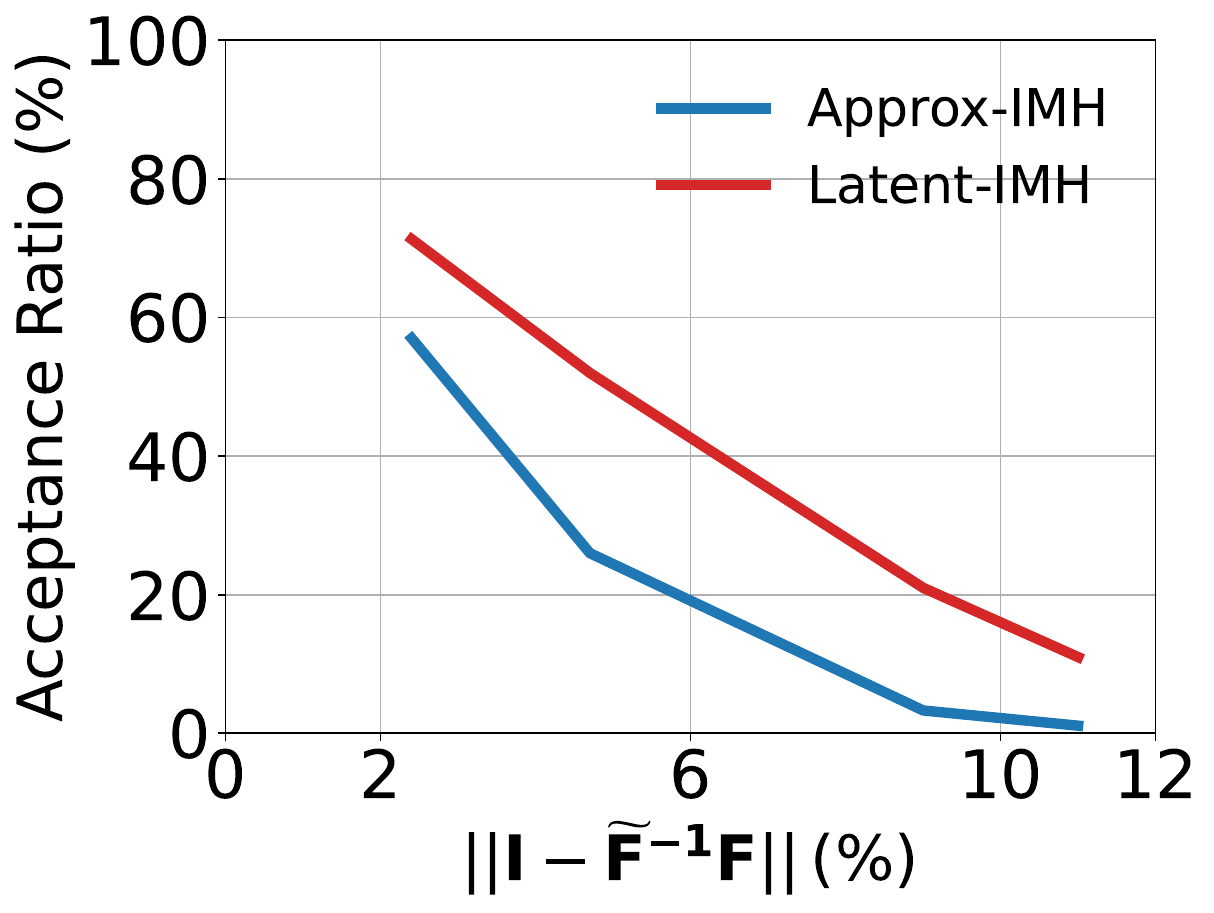}};
\node[inner sep=0pt] (c1) at (8.5,-8.) {\includegraphics[width=3.5cm]{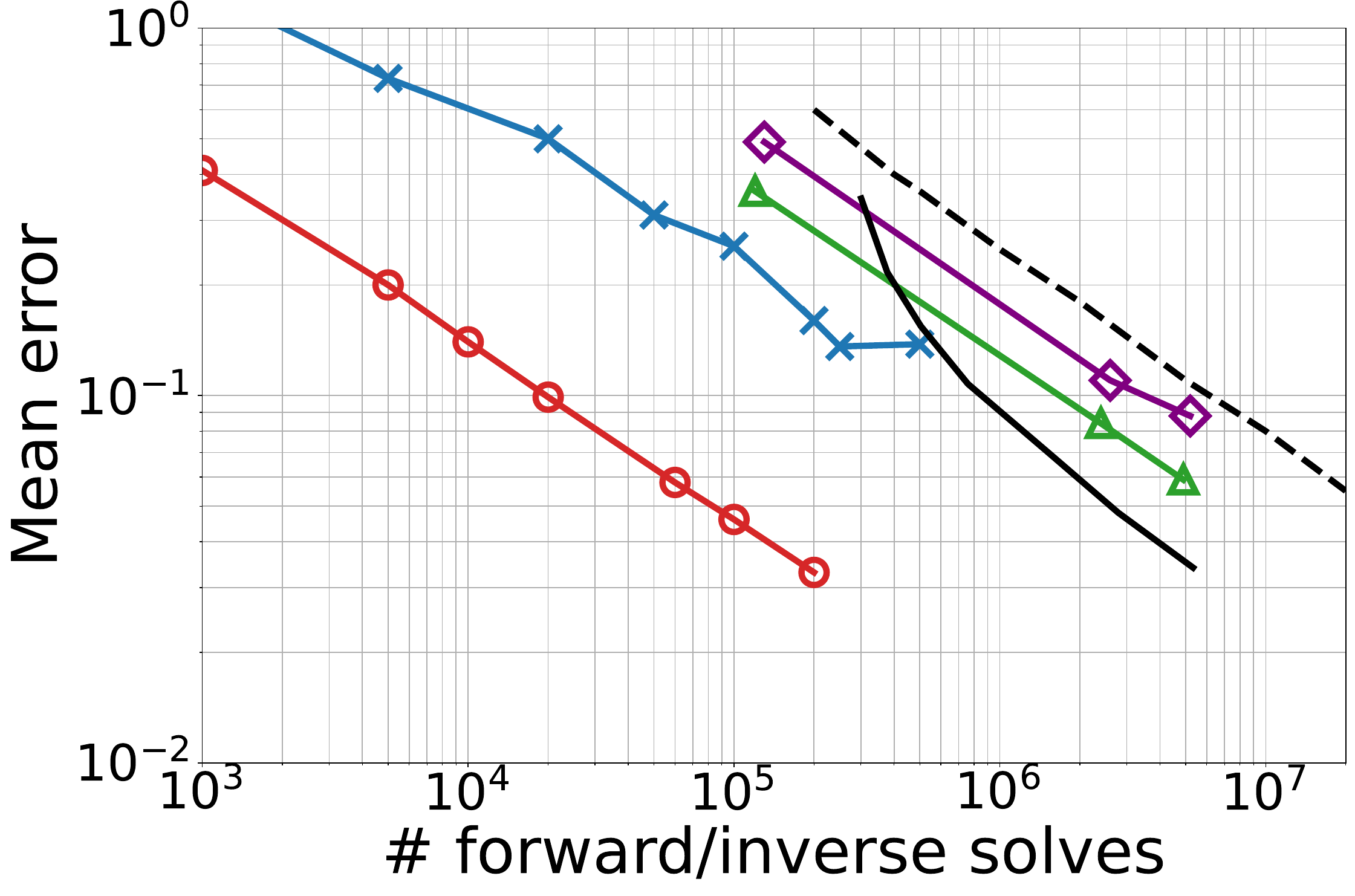}};
\node[inner sep=0pt] (c2) at (8,-3.6) {\includegraphics[width=5cm]{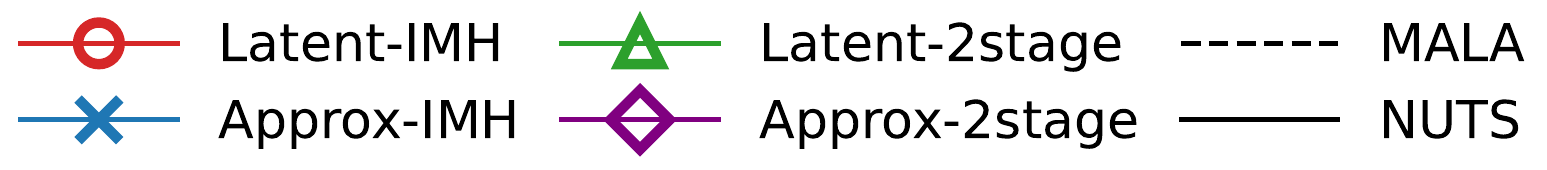}};
\draw[dashed,line width=1pt] (5.3, -9) -- (5.3, -3.5);
\node[draw,rotate=90,rounded corners,fill=black!0,align=center] at (-5.6,-6) {Gaussian mixture prior:\\ $p(x)\sim \sum_{i=1}^3 \mathcal{N}(\mu_i, \b I)$};
\node[draw,rotate=90,rounded corners,fill=black!0,align=center] at (6.2,-6) {Standard normal prior:\\ $p(x)\sim \mathcal{N}(\b 0, \b I)$};
\end{tikzpicture}
\caption{Sample efficiency comparison for Gaussian priors: results averaged over 5 independent runs. \textbf{Left (Gaussian mixture prior):} Numbers in parentheses in the top legend indicate spectral error $\|\b I - \Fapprox^{-1} \F\|_2$; numbers above histogram plots denote the total number of forward and inverse solves of $\F$ required by each sampler. \textbf{Right (standard normal prior):} The approximate operator in the mean error plot has spectral error of 4.7\%.}
    \label{fig:mcmc-gaussian}
\end{figure*}


Our numerical experiments aim to investigate the following questions: How do \HybridMH and \ApproxMH compare in terms of acceptance rate and sampling efficiency? How do they perform relative to local MCMC methods such as NUTS~\citep{hoffman2014no}, MALA~\citep{mala}, and the two-stage multi-fidelity MCMC algorithm~\citep{2stage}? Finally, how sensitive are these methods to variations in problem structure and parameter settings?

\paragraph{Sensitivity tests (acceptance ratio).} 
We first compare acceptance ratios of \ApproxMH and \HybridMH using the same setting as in \cref{sec:KL} (see~\cref{tab:kl-experiments}). Results in \cref{fig:mcmc-sensitivity} show that as spectral error or dimensionality increases, acceptance ratios decline for both methods, but much more sharply for \ApproxMH. Moreover, \ApproxMH is highly sensitive to noise level and observation ratio, while \HybridMH remains robust.

\paragraph{Experimental setup for efficiency tests.} 
We evaluate sample quality using three metrics: \nbone relative sample mean error, \nbtwo squared bias of the second moment \citep{Hoffman2019}, \nbthree maximum mean discrepancy (MMD) (see \cref{sec:add-experiments}). Since forward and inverse solves of the exact operator $\F$ are computationally expensive, we measure cost by the number of such solves: each \HybridMH step requires an inverse solve of $\F^{-1}$, while each \ApproxMH step requires a forward solve of $\F$ to compute acceptance ratios.

Baselines include NUTS and MALA. For NUTS, we use 500-1000 warm-up steps with target acceptance 45\%; for MALA, step size is tuned to yield $\sim$50\% acceptance. We assume Gaussian noise with relative level $\|e\|/\|y\| \in [5\%,15\%]$, and report averages over 5 independent runs for each test (see details in \cref{sec:add-experiments}).

\paragraph{Gaussian and Gaussian mixture priors}
We consider $d=500$, $d_y=50$, with synthetic $\F$ and $\Fapprox$ as in \cref{sec:KL}, and test three priors:  
\nbone a standard normal $\mathcal{N}(\b 0,\b I)$,  
\nbtwo an ill-conditioned Gaussian $\mathcal{N}(\b 0,\bSigma)$ with condition number $\sim$1000,  
\nbthree a three-component Gaussian mixture $\sum_{i=1}^3 \mathcal{N}(\mu_i,\b I)$.  

Results for the standard normal and Gaussian mixture are in \cref{fig:mcmc-gaussian}, while the ill-conditioned case appears in \cref{fig:mcmc-ill-condition} (\cref{sec:add-experiments}). For the test of standard normal prior, we include NUTS and MALA as baselines, as well as the two-stage delayed acceptance MCMC method~\citep{2stage}, which also leverages approximate operator to improve sampling efficiency. In this method, proposals are first generated via local updates using the approximate model, and acceptance is decided using the exact model in the second stage. We use MALA for the first stage, and acceptance is then determined using the exact model. We employ MALA in the first stage, denoting the variant using \Approx as Approx-2stage and the variant using \Hybrid as Latent-2stage. We include more comparison with delayed-acceptance MCMC in \cref{sec:add-results}.

From \cref{fig:mcmc-gaussian}, \HybridMH consistently outperforms other methods in sampling efficiency, with the advantage being particularly pronounced for the Gaussian mixture prior. In the histogram plots (bottom row of \cref{fig:mcmc-gaussian}), local MCMC methods such as NUTS exhibit poor mixing in multimodal settings due to potential barriers between modes.

\begin{figure}[t]
    \centering
\begin{tikzpicture}
\node[inner sep=0pt] (a) at (0,0) {\includegraphics[width=3.5cm]{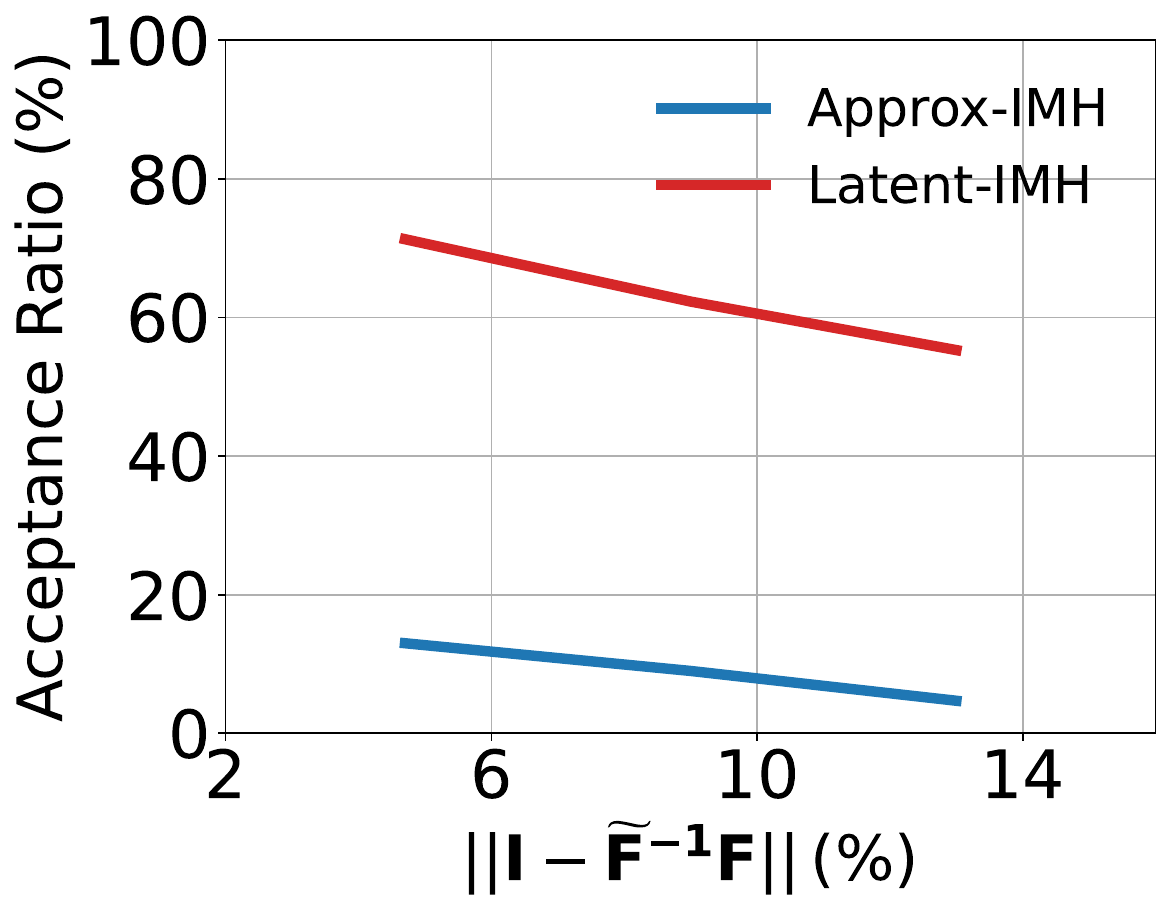}};
\node[inner sep=0pt] (b) at (3.7,0) {\includegraphics[width=3cm]{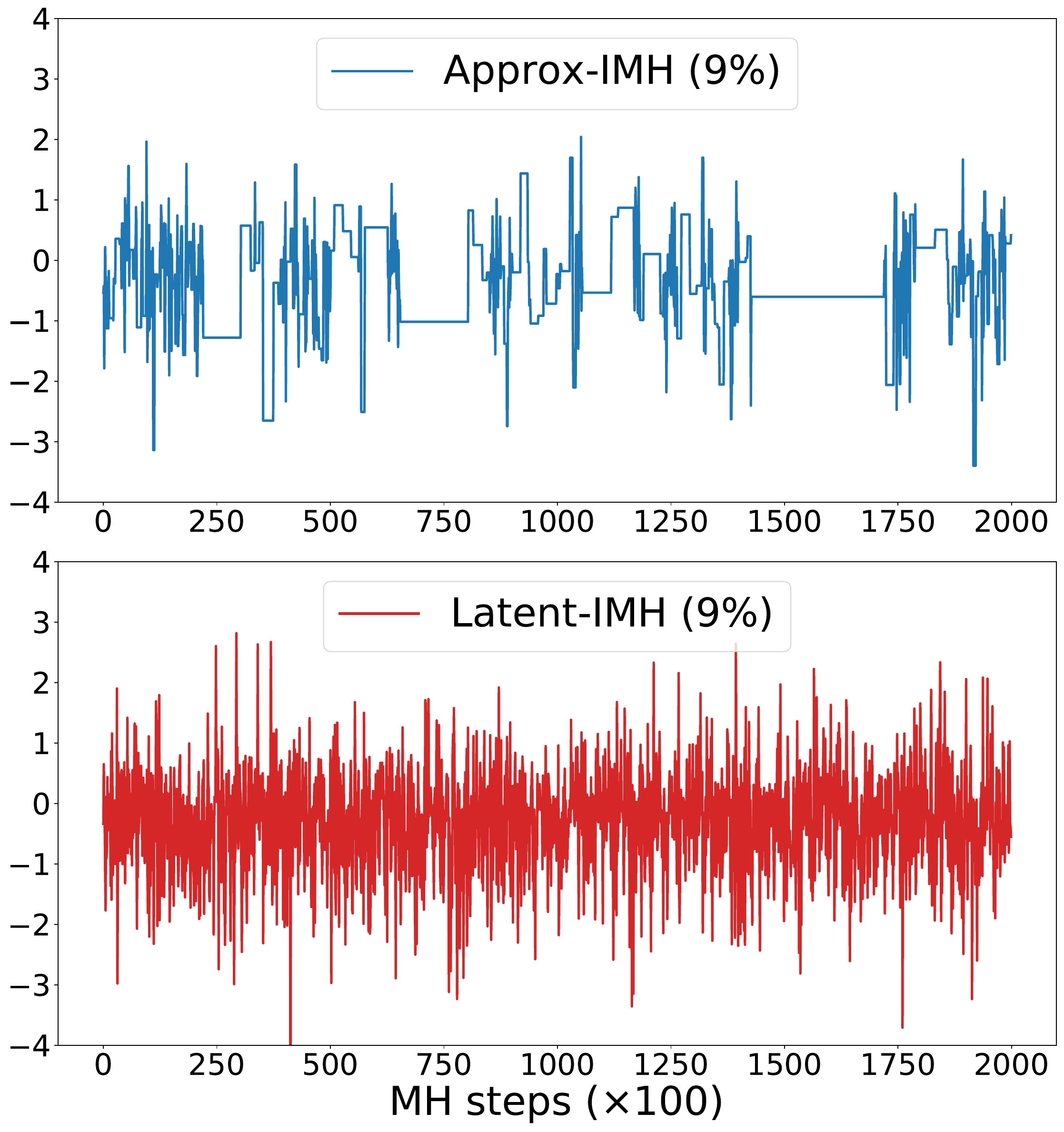}};
\node[inner sep=0pt] (d) at (2,-2) {\includegraphics[width=7.5cm]{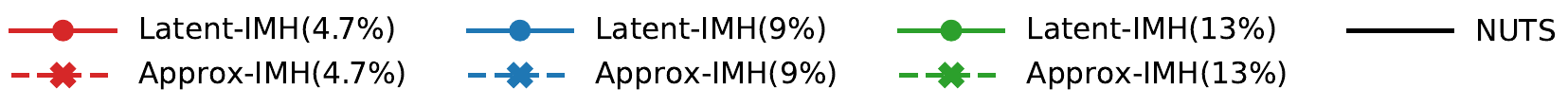}};
\node[inner sep=0pt] (c) at (1.8,-3.5) {\includegraphics[width=7cm]{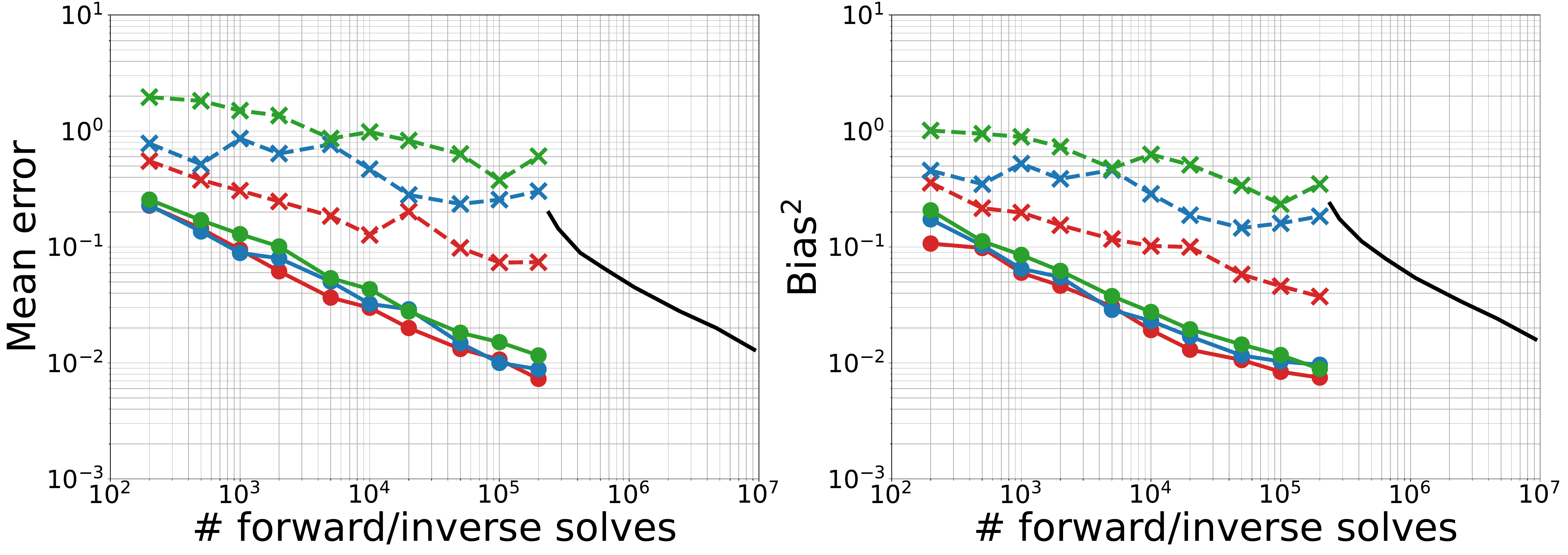}};
\end{tikzpicture}
   \caption{Results for Laplace prior test with normalizing flow. All results averaged over 5 independent runs. Numbers in parentheses indicate spectral error $\|\b I - \Fapprox^{-1} \F\|_2$.}
    \label{fig:nf}
\end{figure}

\paragraph{Laplace prior with normalizing flow.}
We consider a setting with $d=50$ and $d_y=10$, where $x$ follows an independent Laplace prior
$p(x) = 2^{-d}\exp(-|x|1)$. The operators $\F$ and $\Fapprox$ are constructed as in the Gaussian prior experiments.
In \HybridMH, we train a normalizing flow to approximate $\widetilde{p}(u)$ using samples ${x_i, \Fapprox x_i}{i=1}^N$, and then apply NUTS to sample $u$ (\cref{eq:hybrid-sample}), with both log probabilities and gradients provided by the flow. Hence, the approximation arises from both $\Fapprox$ and the normalizing flow, while the acceptance ratio is computed via \cref{eq:ar-hybrid-1}.
As shown in \cref{fig:nf}, \HybridMH consistently outperforms both \ApproxMH and NUTS. Although \HybridMH and NUTS achieve similar convergence rates, both substantially outperform \ApproxMH, which frequently rejects proposals, leading to stagnation in its trace plots compared with the more dynamic behavior of \HybridMH.

\begin{figure*}[t]
    \centering
\begin{tikzpicture}
\node[inner sep=0pt] (a1) at (0,0) {\includegraphics[width=8cm]{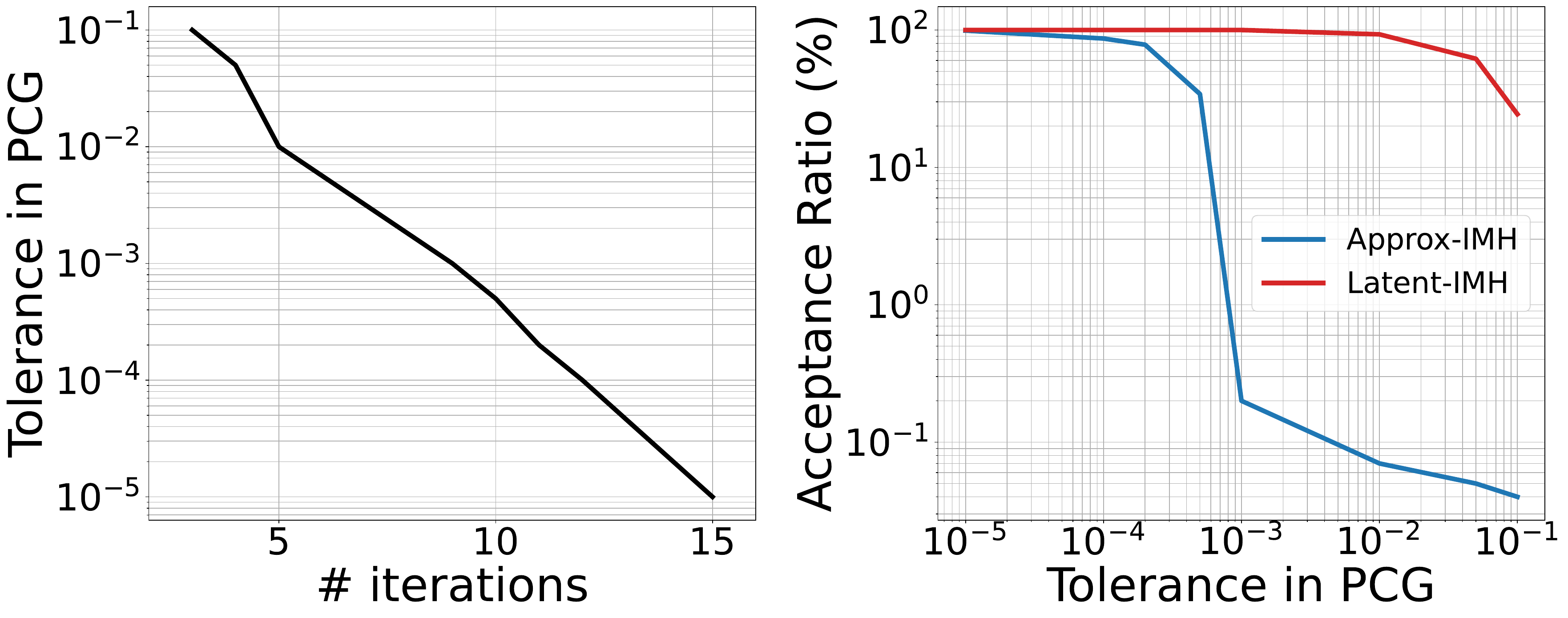}};
\node[inner sep=0pt] (b) at (8.5,2) {\includegraphics[width=8cm]{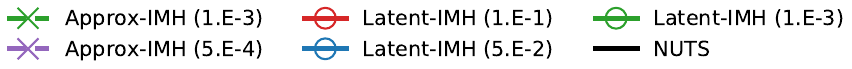}};
\node[inner sep=0pt] (a2) at (8.1,0) {\includegraphics[width=8cm]{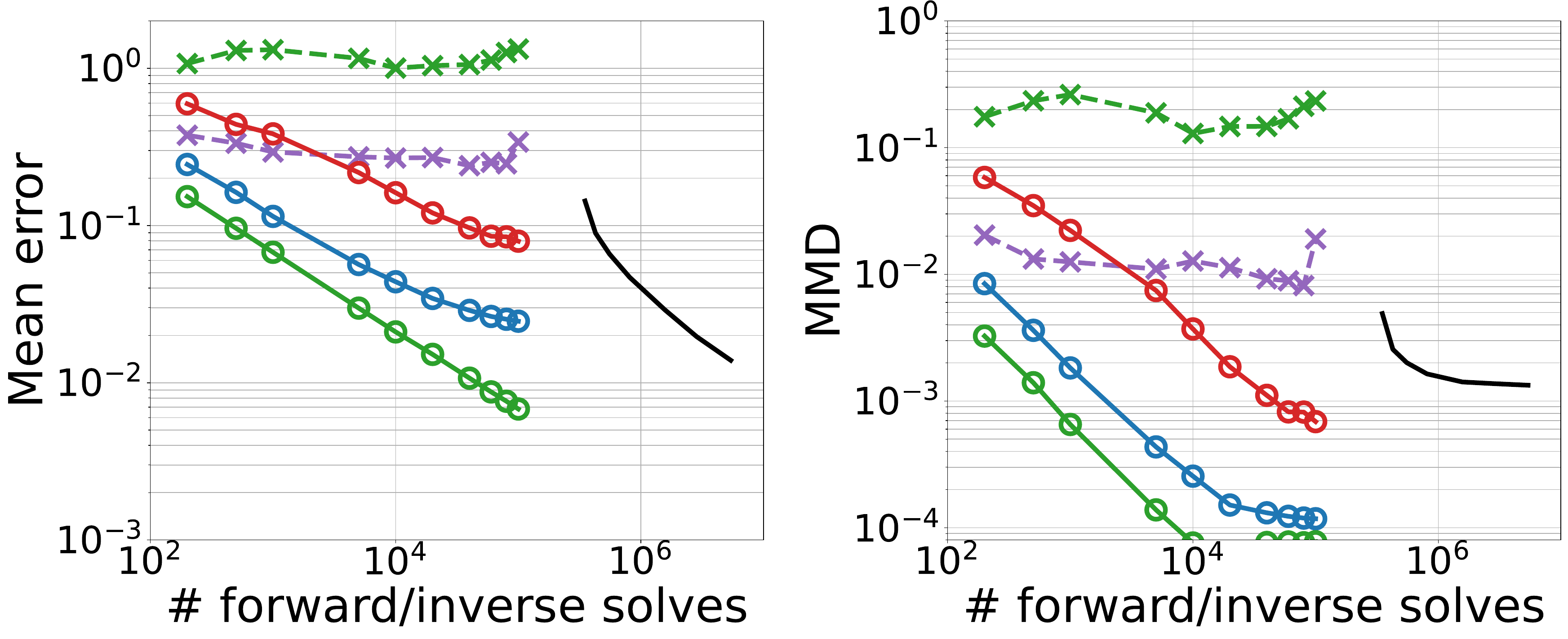}};
\end{tikzpicture}
    \caption{Results for Graph Laplacian test with PCG approximation. Left plot shows PCG iterations vs. tolerance parameter. All results averaged over 5 independent runs. Numbers in parentheses of the legend indicate PCG tolerance (smaller values yield more iterations and higher accuracy).}
    \label{fig:graph-cg}
\end{figure*}

\paragraph{Graph Laplacian test with PCG approximation.} 
We consider problem \cref{eq:prob} with $d_u=8,000$, $d_x =800$ {\color{gray}($=0.1 \du$)}, and $d_y = 160$ {\color{gray}($=0.2 \dx$)}, and standard normal for the prior of $x$. The Laplacian $\Lo$ constructed rom a six-nearest neighbor graph on a $20^3$ lattice in the  unit cube; $\B$ and $\Oo$ are random Gaussian matrices. Since $\F$ is not square,  we construct $\V_x$ and $\b Z$ as described in \cref{sec:background} to transform \eqref{eq:prob} into \eqref{eq:transform}. To approximate $\b L^{-1} x$, we compute a randomized Cholesky factorization $\b L \approx \b G \b G^\top$~\citep{Chen21}, then solve using preconditioned conjugate gradient (PCG) with 
$\b G \b G^\top$ as preconditioner. PCG tolerance controls approximation accuracy: smaller tolerance requires more iterations and higher cost (left plot, \cref{fig:graph-cg}). Results in \cref{fig:graph-cg} show \HybridMH consistently achieves higher acceptance rates than \ApproxMH for the PCG tolerances (second plot). The rightmost plots show that \HybridMH  outperforms both \ApproxMH and NUTS in terms of the relative mean error and the MMD error. In particular, \HybridMH at a tolerance of 0.1 (solid red) achieves even better performance than \ApproxMH at a much stricter tolerance of $5\times 10^{-4}$ (purple dashed), despite the latter having an even higher acceptance rate of 34.2\%.

\paragraph{Scattering problem.} 
We consider Bayesian inference for a 2D scattering problem introduced in \cref{sec:introduction}. We study problem \cref{eq:prob}, where $\b L$ is the Helmholtz operator and $\B$ is the lifting operator. Operators are constructed via a finite difference scheme on a fine grid ($n_u \times n_u$ nodes). For the approximation operator, we first construct coarse-grid operators $\widetilde{\b L}$ and $\widetilde{\B}$ (on $\widetilde{n}_u \times \widetilde{n}_u$ nodes) and then map them to the fine grid using a prolongation operator $\b P$. The resulting approximation to $\b L^{-1}\B$ is $\b P \widetilde{\b L}^{-1} \widetilde{\B}$. We consider multiple events with multiple observations and place a total variation prior on $x$, i.e., $p(x) \propto \exp(-\lambda \mathrm{TV}(x))$ (details in \cref{sec:add-experiments}). In our experiments, we consider 4 events with $n_x=24$, $n_u=128$, and $\widetilde{n}_u=60$, resulting in dimensions $d_u = 165{,}536$, $\widetilde{d}_u = 14{,}400$ {(\color{gray}$\approx 0.2 d_u$)}, $d_x = 576$, and $d_y = 128$ {(\color{gray}$\approx 0.2 d_x$)}. Results in \cref{fig:helmholtz} show that \HybridMH efficiently samples the posterior using the cheaper approximate operator.

\section{CONCLUSIONS}\label{sec:conclusion}
The advantage of the proposed methodology is that it is simple to implement and can be effective in a wide range of scenarios, including multimodal distributions, where multimodality is related to the prior $p(x)$.
The reparameterization trick in \cref{sec:background} required an SVD of the observation operator that, in general, is inexpensive to apply.
In summary, we shift computational costs to constructing an approximation $\apu{u}$.
While this step can be costly, it can be performed offline and reused across multiple problem instances.
Our results show that the method is not universally preferable-its benefit depends on the accuracy and cost of $\Fapprox$, the problem dimension, and the application context, where alternatives such as  NUTS, or other standard methods may be more suitable.

\textbf{Limitations.}  
\newimh{} is most relevant for large problems where memory or computational constraints prevent forming a dense $\F$. Constructing $\F$ requires $\dx$ applications of ${\bf L}^{-1}$ and $\dx\du$ storage, and applying the dense $\F$ to a vector has similar cost. Note that in many applications $\Oo, \Lo, \B$ are sparse, allowing matrix-free application of $\F$ at $O(\du)$ cost. These tradeoffs determine the practicality of \newimh{} and are problem-dependent. A clear  limitation of \newimh{} is the assumption of linearity in $\A$; generalization to nonlinear $\Lo$ or $\B$ is possible if $\Lo^{-1}$ yields unique solutions, but handling nonlinear $\Oo$ is more challenging since the reparameterization trick is no longer applicable. 

Another limitation is the requirement that $\dy\leq \dx$. 
For linear operators, our method can be generalized, but it is unclear whether it will be computationally beneficial.
The approximation $\apu{u}$  can have several components. It comprises the error due to $\Fapprox$ and possibly a machine learning algorithm to reconstruct it for the sample. 
From a theoretical perspective, we would like to connect an error between $p(u)$ and $\apu{u}$ to the overall performance of the method. 
Overcoming all these limitations is ongoing work in our group. 
\subsubsection*{Acknowledgements}

This material is based upon work supported by NSF award OAC 22042261; and  Cooperative Agreement 2421782 and the Simons Foundation award MPS-AI-00010515 (NSF-Simons AI Institute for Cosmic Origins---CosmicAI, \url{https://www.cosmicai.org/}); by the U.S. Department of Energy, Office of Science, Office of Advanced Scientific Computing Research, Applied Mathematics program, Mathematical Multifaceted Integrated Capability Centers (MMICCS) program, under award number DE-SC0023171; by the U.S. Department of Energy, National Nuclear Security Administration Award Number DE-NA0003969; and by the U.S. National Institute on Aging under award number  R21AG074276-01. Any opinions, findings, and conclusions or recommendations expressed herein are those of the authors and do not necessarily reflect the views of the DOE, NIH, and NSF. Computing time on the Texas Advanced Computing Centers Stampede system was provided by an allocation from TACC and the NSF.

\bibliographystyle{plainnat}  
\bibliography{references}  
\section{CHECKLIST}

\begin{enumerate}

  \item For all models and algorithms presented, check if you include:
  \begin{enumerate}
    \item A clear description of the mathematical setting, assumptions, algorithm, and/or model. [Yes]
    \item An analysis of the properties and complexity (time, space, sample size) of any algorithm. [Yes]
    \item (Optional) Anonymized source code, with specification of all dependencies, including external libraries. [Yes]
  \end{enumerate}

  \item For any theoretical claim, check if you include:
  \begin{enumerate}
    \item Statements of the full set of assumptions of all theoretical results. [Yes]
    \item Complete proofs of all theoretical results. [Yes]
    \item Clear explanations of any assumptions. [Yes]     
  \end{enumerate}

  \item For all figures and tables that present empirical results, check if you include:
  \begin{enumerate}
    \item The code, data, and instructions needed to reproduce the main experimental results (either in the supplemental material or as a URL). [Yes]
    \item All the training details (e.g., data splits, hyperparameters, how they were chosen). [Yes]
    \item A clear definition of the specific measure or statistics and error bars (e.g., with respect to the random seed after running experiments multiple times). [Yes]
    \item A description of the computing infrastructure used. (e.g., type of GPUs, internal cluster, or cloud provider). [Yes]
  \end{enumerate}

  \item If you are using existing assets (e.g., code, data, models) or curating/releasing new assets, check if you include:
  \begin{enumerate}
    \item Citations of the creator If your work uses existing assets. [Yes]
    \item The license information of the assets, if applicable. [Not Applicable]
    \item New assets either in the supplemental material or as a URL, if applicable. [Not Applicable]
    \item Information about consent from data providers/curators. [Not Applicable]
    \item Discussion of sensible content if applicable, e.g., personally identifiable information or offensive content. [Not Applicable]
  \end{enumerate}

  \item If you used crowdsourcing or conducted research with human subjects, check if you include:
  \begin{enumerate}
    \item The full text of instructions given to participants and screenshots. [Not Applicable]
    \item Descriptions of potential participant risks, with links to Institutional Review Board (IRB) approvals if applicable. [Not Applicable]
    \item The estimated hourly wage paid to participants and the total amount spent on participant compensation. [Not Applicable]
  \end{enumerate}

\end{enumerate}

\clearpage
\appendix
\thispagestyle{empty}

\onecolumn
\aistatstitle{APPENDIX}
\section*{APPENDIX OUTLINE}
This appendix is organized as follows:
\begin{itemize}[leftmargin=*]
    \item In \cref{sec:transform}, we derive the transformation from the rectangular matrix $\F$ to an invertible square matrix, corresponding to the transition from \eqref{eq:prob} to \eqref{eq:transform}.
    \item In \cref{sec:KL-proof}, we begin by establishing KL-divergence bounds for \Approx and \Hybrid (\cref{thm:kl-general}) under a more general assumption (\cref{assume:kl-general}) than the diagonal case considered in \cref{sec:KL}. We then provide detailed proofs of \cref{prop:kl-closedform}, \cref{prop:kl-simple}, and \cref{thm:kl-general}.
    \item In \cref{sec:mcmc-proof}, we present proofs for the mixing time bounds of \ApproxMH and \HybridMH, as stated in \cref{thm:mix-general} and \cref{thm:mix-normal}.
    \item In \cref{sec:add-experiments}, we include experimental details of the experiments in \cref{sec:KL} and \cref{sec:mcmc-experiment}.
\end{itemize}

\section{REPARAMETERIZATION TRICK FOR RECTANGULAR $\F$}\label{sec:transform}

Let $\b O \in \mathbb{R}^{d_y \times d_x}$ have reduced SVD $\b O = \U \b S \V_y^\top$. We claim that the construction of $\V_x$ and $\b Z$ as specified in \cref{prop:equi-transform} guarantees that the relationship between $x$ and $y$ given in \cref{eq:prob} is equivalent to the transformed system in \cref{eq:transform}.

\begin{proposition}[Construction of $\V_x$ and $\b Z$ in \cref{eq:transform}]\label{prop:equi-transform}
Let $\V_+ \in \mathbb{R}^{d_u \times (d_x - d_y)}$ be a matrix that has orthonormal columns orthogonal to $\V_y$, i.e. $\V_+^\top \V_y = \b 0$ and $\V_+^\top \V_+ = \b I$. If we construct $\V_x$ and $\b Z$ as described below, then the relationship between $x$ and $y$ in \eqref{eq:prob} is equivalent to that in \eqref{eq:transform}:
\begin{align}\label{eq:transform-vx-z}
 \V_x = [\V_y, \V_+], \qquad \b Z = \b O \V_x .
\end{align}
\end{proposition}
\begin{proof}
It suffices to show that $\b Z \V_x^\top = \b O$. By the definitions of $\V_x$ and $\b Z$ in \eqref{eq:transform-vx-z}, we have
\begin{align}
   \b Z \V_x^\top& = \b O \V_x \V_x^\top\nonumber\\
   & = \U\b S \V_y^\top [\b V_y, \V_1]\begin{bmatrix}
       \V_y^\top \\
       \V_1^\top
   \end{bmatrix} = \U\b S  [\b I,\b 0]\begin{bmatrix}
       \V_y^\top \\
       \V_1^\top
       \end{bmatrix} = \U \b S \V_y^\top = \b O. \nonumber
\end{align}  
\end{proof}

To improve the conditioning of $\V_x^\top \b L^{-1} \b B$ and $\V_x^\top \widetilde{\b L}^{-1} \b B$, we construct $\V_+$ from the first $d_x - d_y$ left singular vectors of 
$(\b I - \V_y \V_y^\top)\widetilde{\b L}^{-1} \b B$, corresponding to its $d_x - d_y$ largest singular values. 
This ensures that $\V_+$ is orthogonal to $\V_y$ while capturing the dominant energy of $\widetilde{\b L}^{-1} \b B$.

\section{\MakeUppercase{Proofs of Section 3}}\label{sec:KL-proof}
Recall that we denote $\KLapprox$ and $\KLnew$ as expected KL-divergence between \Approx and \Hybrid against \Exact:
\begin{align}
    \KLapprox:= 2 \E_y[\KL(\piapprox(x|y)|| \pi(x|y))], \quad  \KLnew:= 2 \E_y[\KL(\pinew(x|y)|| \pi(x|y))].
\end{align}

In this section, we first present the general results for $\KLapprox$ and $\KLnew$ in \cref{thm:kl-general}, located in \cref{sec:kl-general}. We then provide the proofs for \cref{prop:kl-closedform} in \cref{sec:kl-proof-closed}, \cref{prop:kl-simple} in \cref{sec:kl-proof-simple}, and \cref{thm:kl-general} in \cref{sec:kl-proof-general}.

\subsection{\MakeUppercase{Expected KL-divergence bounds for the general setting}}\label{sec:kl-general}

We define matrix $\K \in \mathbb{R}^{d \times d}$, reduced and full SVD of $\A$ and $\Aapprox$ as follows:
\begin{align}
    &\K := \F^{-1} \Fapprox,\\
    \A = \U \b S \Vpar^\top = \U[\b S \vdots \b 0]  \begin{bmatrix}
        \Vpar^\top \\
        \Vperp^\top
    \end{bmatrix},&\quad
    \Aapprox = \Uapprox \Sapprox \Vapproxpar^\top = \Uapprox [\Sapprox \vdots \b 0]  \begin{bmatrix}
        \Vapproxpar^\top \\
        \Vapproxperp^\top
    \end{bmatrix},
\end{align}
where $\b S$ and $\Sapprox$ are diagonal matrices with $(\b S)_{ii} = s_i$ and $\Sapprox_{ii} = \widetilde{s}_i$, for all $i\in[d_y]$.

Next, we introduce the constants $\kappa_+$, $\kappa_ -$, $\epsilon$ and $\tau$  in  \cref{assume:kl-general} to quantify the proximity between $\F$ and $\Fapprox$, as well as $\A$ and $\Aapprox$.

\begin{assumption}[($\kappa_+$, $\kappa_ -$, $\epsilon$, $\tau$)-condition]\label{assume:kl-general} 
We assume there exit constants $\kappa_+,\kappa_-, \tau >0$ and $\epsilon\in(0,1)$ satisfying the following conditions:
\begin{itemize}[leftmargin=*]
    \item $\kappa_+$ and  $\kappa_-$ are the largest and smallest singular values of $\K$ respectively. 
    \item $\forall i \in[d]$, $\b v_i^\top \widetilde{\b v}_i \in[1-\epsilon, 1+\epsilon]$ and $\b u_i^\top \widetilde{\b u}_i \in[1-\epsilon, 1+\epsilon]$.
    \item $\| \Vpar \b S_\sigma \Vpar^\top - \Vapproxpar \Sapprox_\sigma \Vapproxpar^\top\|_F^2 \leq \tau$ and $\| \U \b S_\sigma \U^\top - \Uapprox \Sapprox_\sigma \Uapprox^\top\|_F^2 \leq \tau$, where
    \begin{align}\label{eq:S_sigma}
        \b S_\sigma:= \mathrm{Diag}\left[\frac{1}{1+\sigma^2/s_i^2}\right]_{i\in[d_y]}, \qquad \Sapprox_\sigma : = \mathrm{Diag}\left[\frac{1}{1+\sigma^2/\widetilde{s}_i^2}\right]_{i\in[d_y]},
    \end{align}
\end{itemize}
where $\{ s_i\}_{i\in[d_y]}$ and $\{\widetilde{s}_i\}_{i\in[d_y]}$ are singular values of $\A$ and $\Aapprox$, respectively.
    
\end{assumption}

Specifically, $\kappa_+ $ and $\kappa_-$  capture the similarity between $\F$ and $\Fapprox$, while $\epsilon$ and $\tau$ measure the closeness between $\A$ and $\Aapprox$. Note that when $s_i \gg \sigma$, $(\b S_\sigma)_{ii}$ and $(\Sapprox_\sigma)_{ii}$ are close to 1, whereas when $\sigma \ll s_i$, they are close to 0. Therefore, $\tau$ serves as a measure of how close the singular vectors of $\A$ and $\Aapprox$ that correspond to relatively large singular values are. Clearly, when $\F = \Fapprox$, we have $\kappa_+ = \kappa_- = 1$ and $\epsilon = \tau = 0$. 

Using \cref{assume:kl-general} along with the Gaussian assumptions for the prior and noise from \cref{assume:p-and-q}, we can establish upper bounds for $\KLapprox$ and $\KLnew$ as stated in \cref{thm:kl-general}.

\begin{theorem}\label{thm:kl-general}
Given \cref{assume:p-and-q} and \cref{assume:kl-general} we obtain the following upper bounds for $\KLapprox$ and $\KLnew$ respectively:
\begin{align}
    \KLapprox& \leq 2\epsilon d + \sum_{i\in[d_y]} \left((1+2 \epsilon) \rho_i +\log\gamma_i \right) \nonumber\\
    &+\left( 1+\frac{\|\A\|^2}{\sigma^2}\right) \left\{ \tau + \kappa_a d_y + \sigma^2 \sum_{i\in[d_y]}\left( \left(\frac{1}{\zeta_i} - \frac{1}{\widetilde{\zeta}_i}\right)^2 + \frac{4\epsilon}{\zeta_i \widetilde {\zeta}_i}\right)\right\}, \label{eq:kl-general-approx}\\
    \KLnew & \leq 2 \epsilon_l d + \sum_{i\in[d_y]}\left( (1+ 2\epsilon_l) \rho_i+ \log\frac{\gamma_i}{\kappa_-^2} \right)\nonumber\\
    & + \left(2+\frac{\|\A\|^2}{\sigma^2} \right)\tau + \kappa_l d_y + \sigma^2  \sum_{i \in[d_y]}\left(\frac{\kappa_+^2}{\widetilde{\zeta}_i^2} + \frac{1}{\zeta_i^2}- \frac{2(\kappa_- + \epsilon(\kappa_--1))}{\zeta_i \widetilde{\zeta}_i}\right), \label{eq:kl-general-new}
\end{align}
where $\epsilon_l$, $\kappa_a$ and $\kappa_l$ are constants defined by
\begin{align}
    \epsilon_l = \kappa_+ (1+\epsilon) - 1,\quad \kappa_a =\max\left\{ \left|\frac{1}{\kappa_-} -1\right|,\, \left| \frac{1}{\kappa_+}-1 \right|\right\}^2, \quad \kappa_l = \frac{\kappa_+^2}{\kappa_-^2} -\frac{2\kappa_-}{\kappa_+} +1,
\end{align}
and for all $i \in[d_y]$, $\gamma_i$, $\rho_i$, $\zeta_i$  and $\widetilde{\zeta}_i$ are constants defined by
\begin{align}
\gamma_i = (\widetilde{s}_i^2 + \sigma^2)/(s_i^2+\sigma^2), & \qquad \rho_i = (s_i^2 - \widetilde{s}_i^2)/(\widetilde{s}_i^2 + \sigma^2),\\
\zeta_i  = s_i + \sigma^2/s_i, &\qquad \widetilde{\zeta}_i = \widetilde{s}_i + \sigma^2/\widetilde{s}_i.
\end{align}

\end{theorem}

\paragraph{Remark.} We highlight several observations regarding the bounds established in \cref{thm:kl-general}. First, note that both bounds satisfy $\KLapprox=0$ and $\KLnew=0$ when $\Fapprox = \F$. In many practical applications, the signal-to-noise ratio satisfies $\frac{\|\A\|}{\sigma} \gg 1 $. Assuming $\tau = \mathcal{O}(1)$ and both $d$ and $d_y$ are large,  the bounds can be approximated as
\begin{align}
    \KLapprox \sim \epsilon d + \frac{\|\A\|^2}{\sigma^2}  d_y,\qquad \KLnew \sim \epsilon d + d_y + \frac{\|\A\|^2}{\sigma^2}.
\end{align}
This shows that $\KLapprox$  is highly sensitive to noise. Furthermore, both $\KLapprox$ and $\KLnew$ grow with increasing spectral approximation error and problem dimensions. This is consistent to our analysis in the simpler diagonal case (see \cref{prop:kl-simple}).

\subsection{Proof of \cref{prop:kl-closedform}}\label{sec:kl-proof-closed}

\begin{Lemma}[KL-divergence between two multivariate Gaussians]\label{lm:kl}
\begin{align}
      2  \KL(\mathcal{N}(\mu_1, \bSigma_1) || \mathcal{N}(\mu_2, \bSigma_2))  = \log \frac{|\bSigma_2|}{|\bSigma_1|} + \Tr(\bSigma_2^{-1} \bSigma_1) -d +  (\mu_2 - \mu_1) \bSigma_2^{-1}(\mu_2 - \mu_1).
\end{align}
\end{Lemma}

With the above lemma, we can prove the closed-form expression for $\KLapprox$ and $\KLnew$ in \cref{prop:kl-closedform}:
\begin{proof}[Proof of \cref{prop:kl-closedform}]
When $x\sim \mathcal{N}(0, \b I)$, we have $y \sim \mathcal{N}(\b 0, \A \A^\top + \sigma^2 \b I)$, which is equivalent to the following linear transformation expression for $y$:
\[y = \b M z, \qquad \text{where}\quad   \b M  = (\A\A^\top + \sigma^2 \b I)^{1/2} ,\qquad z \sim \mathcal{N}(\b 0, \b I). \] 

By  \cref{lm:kl}, we have
\begin{align}\label{eq:klnew-1}
    \KLapprox = \left( \log\frac{|\bSigma|}{|\bSigma_a|} + \Tr(\bSigma^{-1} \bSigma_a) - d \right) + \E_y[(\mu_a - \mu)^\top \bSigma^{-1}(\mu_a - \mu)].
\end{align}

By \cref{tab:posterior-gaussian}, we can express the quadratic formula in \Cref{eq:klnew-1} as follows:
\begin{align}
    \E_y[(\mu_a - \mu)^\top \bSigma^{-1}(\mu_a - \mu)] &= \E_z[z^\top  \b M ^\top (\Aapprox^\dagger - \A^\dagger)^\top\bSigma^{-1} (\Aapprox^\dagger - \A^\dagger)  \b M  z ]\\
    & = \E_z[\langle  \b M ^\top (\Aapprox^\dagger - \A^\dagger)^\top\bSigma^{-1} (\Aapprox^\dagger - \A^\dagger)  \b M , z z^\top\rangle] \\
    &= \Tr\left(  \b M ^\top  (\Aapprox^\dagger - \A)^\top\bSigma^{-1} (\Aapprox^\dagger - \A^\dagger)  \b M \right)\\
    &= \frac{1}{\sigma^2}\|\A (\Aapprox^\dagger - \A^\dagger)\A \|_F^2 + \sigma^2\|\Aapprox^\dagger - \A^\dagger\|_F^2\\
    & + \| (\Aapprox^\dagger - \A^\dagger) \A\|_F^2 +  \| \A (\Aapprox^\dagger - \A^\dagger)\|_F^2.
\end{align}

\end{proof}

\subsection{Proof of \cref{prop:kl-simple}}\label{sec:kl-proof-simple}
\begin{proof}
By the diagonal assumptions of $\F$ and $\Fapprox$, $\K=\F^{-1} \Fapprox$ is a diagonal matrix with $\K_{ii} = \alpha_i$ for $\forall i\in[d]$. $\A, \Aapprox\in\mathbb{R}^{d_y \times d}$ are matrices with zero elements except for:
\begin{align}
    \A_{ii} = s_i, \qquad \Aapprox_{ii} =\alpha_i s_i, \quad \forall i\in[d_y].
\end{align}
By definitions in \cref{tab:posterior-gaussian}, $\A^\dagger, \Aapprox^\dagger, \A_l^\dagger \in \mathbb{R}^{d\times d_y}$ with zero elements except for:
\begin{align}
    \A^\dagger_{ii} = \frac{s_i}{s_i^2 +\sigma^2}, \qquad \Aapprox^\dagger_{ii} = \frac{\alpha_i s_i}{\alpha_i^2 s_i^2 + \sigma^2} ,\qquad(\A_l^\dagger)_{ii} = \frac{\alpha^2 s_i}{\alpha_i^2 s_i^2 + \sigma^2}, \quad \forall i \in[d_y].
\end{align}
$\bSigma, \bSigma_a, \bSigma_l \in\mathbb{R}^{d\times d}$ are diagonal matrices with diagonals as  
\begin{align}
   \begin{cases}\label{eq:kl-covariance}
      \bSigma_{ii}= \frac{\sigma^2}{s_i^2 +\sigma^2},\quad (\bSigma_a)_{ii} = \frac{\sigma^2}{\alpha_i^2 s_i^2 +\sigma^2}, \quad (\bSigma_l)_{ii} = \frac{\alpha^2 \sigma^2}{\alpha_i^2 s_i^2 + \sigma^2}, \quad i\leq d_y,\\
        \bSigma_{ii}= 1,\quad (\bSigma_a)_{ii} = 1 ,\quad (\bSigma_l)_{ii} = \alpha_i^2,\quad i >d_y.
    \end{cases}
\end{align}
Thus $\daapprox, \danew\in\mathbb{R}^{d\times d_y}$ are matrices with zero elements except for:
\begin{align}
    &(\daapprox)_{ii} = \Aapprox^\dagger_{ii} - \A^\dagger_{ii} = \frac{s_i (\alpha_i s_i^2-\sigma^2)(1-\alpha_i)}{(\alpha_i^2s_i^2 + \sigma^2)(s_i^2 + \sigma^2)}, \forall i\in[d_y] \label{eq:kl-da-approx}\\
    &(\danew)_{ii} = (\A_l^\dagger)_{ii} - \A^\dagger_{ii} = \frac{\sigma^2 s_i (\alpha_i^2 - 1)}{(\alpha_i^2s_i^2 + \sigma^2)(s_i^2 + \sigma^2)}, \forall i\in[d_y].\label{eq:kl-da-new}
\end{align}
Next we derive the expressions of $\KLapprox$ and $\KLnew$ derived in \cref{prop:kl-closedform}, respectively.
\begin{enumerate}[leftmargin=*]
    \item \Approx: by \cref{eq:kl-covariance} we have
\begin{align}\label{eq:kl-approx-1}
    \log\frac{|\bSigma|}{|\bSigma_a|} = \sum_{i\in[d_y]} \frac{\alpha_i^2 s_i^2 +\alpha^2}{s_i ^2 + \sigma^2} = \sum_{i\in[d_y]} \rho_i,\quad \Tr(\bSigma^{-1} \bSigma_a) = \sum_{i\in[d_y]} \frac{1}{\rho_i} + d- d_y.
\end{align}
By \cref{eq:kl-da-approx}, we have $ \A \daapprox\in\mathbb{R}^{d_y\times d_y}$ is a diagonal matrix with diagonals:
\begin{align}
    (\A \daapprox)_{ii} = \frac{s_i^2 (\alpha_i s_i^2 -\sigma^2)(1-\alpha_i)}{(\alpha_i^2 s_i^2 + \sigma^2)(s_i^2 + \sigma^2)} .
\end{align}
Denote $f_i = (\alpha_i^2 s_i^2 + \sigma^2)^2(s_i^2 + \sigma^2)^2$, then
\begin{align}
   &\frac{1}{\sigma^2} \| \A \daapprox\|_F^2 = \sum_{i\in[d_y]} \frac{s_i^6 (\alpha_i s_i^2 - \sigma^2)^2 (1-\alpha_i)^2/\sigma^2}{f_i},\\
  &\sigma^2 \|\daapprox\|_F^2 =  \sum_{i\in[d_y]}\frac{\sigma^2 s_i^2  (\alpha_i s_i^2 - \sigma^2)^2 (1-\alpha_i)^2}{f_i},\\
 & \| \daapprox \A\|_F^2 = \|\A \daapprox \|_F^2 = \sum_{i\in[d_y]} \frac{s_i^4 (\alpha_i s_i^2 - \sigma^2)^2 (1-\alpha_i)^2}{f_i}.
\end{align}
Thus 
\begin{align}\label{eq:kl-approx-2}
    \frac{1}{\sigma^2} \| \A \daapprox\|_F^2 + \sigma^2 \|\daapprox\|_F^2  + \| \daapprox \A\|_F^2 + \|\A \daapprox \|_F^2 = \sum_{i\in[d_y]} \zeta_i^2 (\alpha_i s_i^2 - \sigma^2)^2 (1-\alpha_i)^2 \frac{s_i^2}{\sigma^2}.
\end{align}
Substitute \cref{eq:kl-approx-1,eq:kl-approx-2} into \cref{eq:kl-approx}, we have the expressions for $\KLapprox$ in \cref{eq:kl-simple-approx}.
\item \Hybrid: by \cref{eq:kl-covariance}, we have
\begin{align}\label{eq:kl-new-1}
 \log\frac{|\bSigma|}{|\bSigma_l|} = \sum_{i\in[d_y]} \log \frac{\rho_i }{\alpha_i^2} +\sum_{i=d_y+1}^d \log\frac{1}{\alpha_i^2},\quad \Tr(\bSigma^{-1} \bSigma_a) = \sum_{i\in[d_y]} \frac{\alpha_i^2}{\rho_i} + \sum_{i=d_y+1}^d \alpha_i^2.
\end{align}
By \cref{eq:kl-da-new}, we have $ \A \danew\in\mathbb{R}^{d_y\times d_y}$ is a diagonal matrix with diagonals:
\begin{align}
    (\A \danew)_{ii} = \frac{\alpha_i^2s_i^2(\alpha_i^2-1)}{(\alpha_i^2 s_i^2 + \sigma^2)(s_i^2 + \sigma^2)} .
\end{align}
Denote $f_i = (\alpha_i^2 s_i^2 + \sigma^2)^2(s_i^2 + \sigma^2)^2$, then
\begin{align}
   &\frac{1}{\sigma^2} \| \A \danew\|_F^2 = \sum_{i\in[d_y]} \frac{\sigma^2 s_i^6(\alpha_i^2-1)^2)}{f_i},\\
  &\sigma^2 \|\danew\|_F^2 =  \sum_{i\in[d_y]}\frac{\sigma^6 s_i^2 (\alpha_i^2-1)^2}{f_i},\\
 & \| \danew\A\|_F^2 = \|\A \danew \|_F^2 = \sum_{i\in[d_y]} \frac{\sigma^4 s_i^4 (\alpha_i^2-1)^2}{f_i}.
\end{align}
Thus 
\begin{align}\label{eq:kl-new-2}
    \frac{1}{\sigma^2} \| \A \danew\|_F^2 + \sigma^2 \|\danew\|_F^2  + \| \danew \A\|_F^2 + \|\A \danew \|_F^2 = \sum_{i\in[d_y]} \zeta_i^2 (\alpha_i^2 - 1)^2 s_i^2 \sigma^2.
\end{align}
Combining \cref{eq:kl-new-1,eq:kl-new-2} with \cref{prop:kl-closedform}, we have the expressions for $\KLapprox$ in \cref{eq:kl-simple-new}.
\end{enumerate}
    
\end{proof}

\subsection{Proof of \cref{thm:kl-general}}\label{sec:kl-proof-general}

\begin{Lemma}\label{lm:pseudo-inverse-diff}
We have the following properties for $\A^\dagger$, $\Aapprox^\dagger$ and $\Anew^\dagger$
\begin{itemize}[leftmargin=*]
    \item $(\Aapprox^\dagger  - \A ^\dagger )\A  = \Aapprox^\dagger  \Aapprox \K^{-1} - \A ^\dagger  \A $.
    \item $(\Anew^\dagger  - \A ^\dagger )\A  = \K  \Aapprox^\dagger  \Aapprox \K^{-1} - \A ^\dagger  \A  $.
    \item $\A  (\Anew^\dagger  - \A ^\dagger )  = \Aapprox \Aapprox^\dagger  - \A  \A ^\dagger $.
\end{itemize}
\end{Lemma}
\begin{proof}
$ $
\begin{itemize}[leftmargin=*]
        \item Note that
\begin{align}\label{eq:step-A0}
\A  = \b O \F = \b O \widetilde{\F}\widetilde{\F}^{-1} \F = \b O \widetilde{\F} (\F^{-1} \widetilde{\F})^{-1} = \Aapprox \K^{-1}.
\end{align}
\item By \cref{tab:posterior-gaussian}, $\Anew^\dagger = \K \Aapprox^\dagger$. The property can be obtained by combing this relation with the previous property.
\item Using \Cref{eq:step-A0}, we have
        \begin{align}
            \A  \Anew^\dagger = \A  \K \Aapprox^\dagger  = \Aapprox \K^{-1} \K \Aapprox^\dagger  = \Aapprox \Aapprox^\dagger .
        \end{align}
    \end{itemize}
\end{proof}

\begin{Lemma}\label{lm:pseudo-inverse}
If \cref{assume:kl-general} holds, we have
    \begin{itemize}[leftmargin=*]
        \item $\|\Aapprox^\dagger \Aapprox - \A^\dagger \A\|\leq \tau$.
        \item $\| \Aapprox \Aapprox^\dagger -  \A\A^\dagger\|\leq \tau$.
    \end{itemize}
\end{Lemma}
\begin{proof}
    By the SVD notation of $\A $, we have
    \begin{align}
        \A ^\dagger = \A ^\top (\A \A ^\top + \sigma^2 \b I)^{-1} = \Vpar \mathrm{Diag}\bigg[ \frac{s_i}{s_i^2 + \sigma^2}\bigg]_{i\in[d_y]} \U^\top.\nonumber
    \end{align}
and then by the definition of $\b S_\sigma$ and $\Sapprox_\sigma$, we have
\begin{align}
        \A ^\dagger \A = \Vpar \b S_\sigma\Vpar^\top , \qquad \A  \A ^\dagger =  \U\b S_\sigma \U.\nonumber
\end{align}
 Similar argument holds for the expressions of $\Aapprox^\dagger \Aapprox$ and $\Aapprox \Aapprox^\dagger$. By \cref{assume:kl-general}, we have $\|\Aapprox^\dagger \Aapprox - \A^\dagger \A\|\leq \tau$ and  $\| \Aapprox \Aapprox^\dagger -  \A\A^\dagger\|\leq \tau$.
\end{proof}

\begin{Lemma}\label{lm:trace-kl-lemma}
We have the following bounds for the covariance related terms in the KL-divergence for \Approx and \Hybrid:
\begin{align}
   \log\frac{|\bSigma|}{|\bSigma_a|} + \Tr (\bSigma^{-1}\bSigma_a) -d &\leq 2\epsilon d + \sum_{i\in[d_y]} (1+2 \epsilon) \rho_i +\log\gamma_i ,\label{eq:covariance-approx}\\
    \log\frac{|\bSigma|}{|\bSigma_l|}+  \Tr (\bSigma^{-1}\bSigma_l) -d &\leq 2 \epsilon_l d + \sum_{i\in[d_y]} (1+ 2\epsilon_l) \rho_i+ \log\frac{\gamma_i}{\kappa_-^2} ,\label{eq:covariance-new}
\end{align}
where 
\begin{align}
   \epsilon_l = \kappa_+ (1+\epsilon) - 1,\qquad  \gamma_i = (\widetilde{s}_i^2 + \sigma^2)/(s_i^2+\sigma^2), \qquad \rho_i = (s_i^2 - \widetilde{s}_i^2)/(\widetilde{s}_i^2 + \sigma^2),\quad \forall i\in[d_y].
\end{align}
\end{Lemma}
\begin{proof}
 For \Approx, by the SVD notation of $\A $ and $\Aapprox$, we have
    \begin{align}
        &\bSigma^{-1} = (\b I - \A^\dagger \A )^{-1} = \V \mathrm{Diag}
        \bigg[\underbrace{\cdots, \frac{s_i^2}{\sigma^2}+1,\cdots}_\text{$i\in[d_y]$}, \underbrace{1, 1, \cdots}_\text{$d-d_y$}\bigg] \V^\top =:\V \b S_1 \V^\top, \nonumber\\
        & \bSigma_a= \widetilde{\V} \mathrm{Diag}\bigg[\underbrace{\cdots, \frac{\sigma^2}{\widetilde{s_i}^2+\sigma^2},\cdots}_\text{$i\in[d_y]$}, \underbrace{1, 1, \cdots}_\text{$d-d_y$}\bigg]\widetilde{\V}^\top =:\widetilde{\V} \b S_2 \widetilde{\V}^\top. \nonumber
    \end{align}

Denote $\V^\top \widetilde{\V}=\b I + \mathbf{\Delta}$, by the \cref{assume:kl-general}, we have diagonals $\mathbf{\Delta}_{ii}\leq \epsilon$, for all $i\in[d]$. Thus
\begin{align}
   \Tr (\bSigma^{-1}\bSigma_a) -d &= \Tr(\V\b S_1 \V^\top \widetilde{\V} \b S_2 \widetilde{\V}^\top)-d\nonumber\\
   &=\Tr(\b S_1(\b I + \mathbf{\Delta}) \b S_2 (\b I + \mathbf{\Delta}^\top))-d \nonumber\\
   &\stackrel{(a)}{=} \Tr(\b S_1 \b S_2) + 2\Tr(\b S_1 \b S_2 \mathbf{\Delta}) - d\nonumber\\
   &\leq  (1+2\epsilon) \Tr(\b S_1 \b S_2) - d\nonumber\\
   &= (1+2 \epsilon) \sum_{i\in[d_y]}\frac{s_i^2 + \sigma^2}{\widetilde{s}_i^2 + \sigma^2} +(1+2\epsilon) (d-d_y) -d\nonumber\\
   &= (1+2\epsilon) \sum_{i\in[d_y]}\frac{s_i^2 - \widetilde{s}_i^2}{\widetilde{s_i}^2 + \sigma^2} + 2\epsilon d,
\end{align}
where step (a) neglected second order term on $\mathbf{\Delta}$.

 For \Hybrid, we can use a similar argument. Since $\bSigma_l = \K \bSigma_a \K^\top$, we have
\begin{align}\label{eq:step-trace-1}
    \Tr(\bSigma^{-1} \bSigma_l) = \Tr(\V \b S_1 \V^\top \K \widetilde{\V} \b S_2 \widetilde{\V}^\top \K^\top) = \Tr(\b S_1 (\V^\top \K \widetilde{\V}) \b S_2 (\V \K \widetilde{\V})^\top).
\end{align}
Denote $\V^\top \K \widetilde{\V} = \b I + \mathbf{\Delta^\prime}$, then the $i$-th diagonal entry of $\mathbf{\Delta^\prime}$ has the following bound:
\begin{align}
    \mathbf{\Delta^\prime}_{ii} & = \b v_i^\top \K \widetilde{\b v}_i -1 = \frac{1}{2}\Tr(\K(\b v_i \widetilde{\b v}_i^\top +  \widetilde{\b v}_i \b v_i^\top)) - 1\nonumber\\
    &\leq \kappa_+ (1+\epsilon) - 1,
\end{align}
where the last inequality follows by \cref{assume:kl-general}.
Now by \Cref{eq:step-trace-1} we have
\begin{align}
     \Tr(\bSigma^{-1} \bSigma_l) - d &= \Tr(\b S_1 (\b I + \mathbf{\Delta^\prime}) \b S_2 (\b I + \mathbf{\Delta^\prime}^\top )) -d \nonumber\\
     & = \Tr(\b S_1 \b S_2) + 2 \Tr(\b S_1 \b S_2  \mathbf{\Delta^\prime}) - d\nonumber\\
     &\leq [2 \kappa_+(1+\epsilon) - 1]\Tr(\b S_1 \b S_2) - d\nonumber\\
     & = [2 \kappa_+(1+\epsilon) - 1]\bigg[\sum_{i\in[d_y]}\frac{s_i^2 +\sigma^2}{\widetilde{s}_i^2 +\sigma^2} + d - d_y\bigg] -d \nonumber\\
     &=  [2 \kappa_+(1+\epsilon) - 1]\sum_{i\in[d_y]}\frac{s_i^2 - \widetilde{s}_i^2}{\widetilde{s}_i^2 + \sigma^2} + 2 [\kappa_+(1+\epsilon) - 1] d.
\end{align}

\end{proof}

\begin{proof}[Proof of \cref{thm:kl-general}]
For the ease of notation, we denote the quadratic terms in the KL-divergence expressions for \Approx and \Hybrid in \cref{prop:kl-closedform} by
\begin{align}
    f &= \underbrace{\frac{1}{\sigma^2}\|\A  (\Aapprox^\dagger - \A ^\dagger)\A  \|_F^2}_\text{$f_1$} + \underbrace{\sigma^2\|(\Aapprox^\dagger - \A^\dagger  ) \|_F^2}_\text{$f_2$}+\underbrace{  \| (\Aapprox^\dagger - \A^\dagger) \A \|_F^2 }_\text{$f_3$}+ \underbrace{ \|\A  (\Aapprox^\dagger - \A^\dagger )\|_F^2}_\text{$f_4$}.\\
  g &= \underbrace{\frac{1}{\sigma^2}\|\A  (\Anew^\dagger - \A^\dagger )\A  \|_F^2}_\text{$g_1$} + \underbrace{\sigma^2 \|(\Anew^\dagger - \A^\dagger  ) \|_F^2}_\text{$g_2$}+\underbrace{ \| (\Anew^\dagger - \A^\dagger ) \A \|_F^2 }_\text{$g_3$}+ \underbrace{\|\A  (\Anew^\dagger - \A^\dagger )\|_F^2}_\text{$g_4$}
\end{align}
\begin{enumerate}[leftmargin=*]
\item \Approx: By \cref{lm:pseudo-inverse-diff}, we have
    \begin{align}
       f_1 &= \frac{1}{\sigma^2}\|\A  (\Aapprox^\dagger - \A^\dagger) \A  \|_F^2 = \frac{1}{\sigma^2}\| \A  (\Aapprox^\dagger \Aapprox \K^{-1} - \A^\dagger \A )\|_F^2 \leq \frac{\|\A \|^2}{\sigma^2} \|\Aapprox^\dagger \Aapprox \K^{-1} -\A^\dagger \A  \|_F^2, \label{eq:step-approx-f1}\\
        f_3 &= \|(\Aapprox^\dagger - \A ) \A \|_F^2 = \|\Aapprox^\dagger \Aapprox \K^{-1} -\A^\dagger \A  \|_F^2.\label{eq:step-approx-f3}
    \end{align}
By \cref{assume:kl-general},
\begin{align}
    \|\Aapprox^\dagger \Aapprox \K^{-1} -\A^\dagger \A  \|_F^2 &= \| \Vapproxpar \Sapprox_\sigma \Vapproxpar^\top \K^{-1} - \Vpar \b S_\sigma \Vpar ^\top\|_F^2\nonumber\\
    & \leq \|\Vapproxpar \Sapprox_\sigma \Vapproxpar^\top \K^{-1} - \Vapproxpar \Sapprox_\sigma \Vapproxpar^\top \|_F^2 + \|\Vapproxpar \Sapprox_\sigma \Vapproxpar^\top -  \Vpar \b S_\sigma \Vpar ^\top\|_F^2\nonumber\\
    & \leq \|\b I - \K^{-1} \|^2 \| \Vapproxpar \Sapprox_\sigma\Vapproxpar^\top \|_F^2 + \|\Vapproxpar \Sapprox_\sigma \Vapproxpar^\top -  \Vpar \b S_\sigma \Vpar ^\top\|_F^2\nonumber\\
    & \leq \max\left\{ \left(\frac{1}{\kappa_-} -1\right)^2,\,\left( \frac{1}{\kappa_+}-1\right)^2\right\}d_y + \tau,\label{eq:step-approx-mid-1}
\end{align}
where the last inequality is derived by \cref{assume:kl-general} and the fact that $\Sapprox_\sigma$ is a diagonal matrix with elements less than 1 (\cref{eq:S_sigma}).

Substitute \Cref{eq:step-approx-mid-1} into \Cref{eq:step-approx-f1,eq:step-approx-f3}, we have
\begin{align}
   f_1 + f_3 \leq \left(1 + \frac{\|\A\|^2}{\sigma^2} \right)\left[ \max\left\{ \left(\frac{1}{\kappa_-} -1\right)^2,\,\left( \frac{1}{\kappa_+}-1\right)^2\right\}d_y + \tau\right].\label{eq:step-approx-f13}
\end{align}
Now we consider $f_2$ and $f_4$:
\begin{align}
   f_2 + f_4 & = \sigma^2 \|\Aapprox^\dagger - \A^\dagger \|_F^2 + \|\A  (\Aapprox^\dagger - \A^\dagger) \|_F^2\nonumber\\
    &\leq (\sigma^2+ \|\A\|^2)\|\Aapprox^\dagger - \A^\dagger\|_F^2.\label{eq:step-approx-mid-2}
\end{align}
Define diagonal matrices $\b S_1 = \mathrm{Diag}\left[s_i/(s_i^2+\sigma^2)\right]_{i\in [d_y]}$ and $\b S_2 = \mathrm{Diag}\left[\widetilde{s}_i/(\widetilde{s}_i^2+\sigma^2)\right]_{i\in [d_y]}$. By \cref{lm:pseudo-inverse}, we have
\begin{align}
    \|\Aapprox^\dagger - \A^\dagger\|_F^2 &\leq \|\Vpar  \b S_1 \U^\top - \Vapproxpar \b S_2 \Uapprox^\top\|_F^2\nonumber\\
    &= \Tr(\Vpar \b S_1^2 \Vpar^\top) + \Tr(\Vapproxpar \b S_2^2 \Vapproxpar^\top)  - 2 \Tr(\Vpar \b S_1 \U^\top \Uapprox \b S_2 \Vapproxpar^\top)\nonumber\\
    & =\sum_{i\in[d_y]}\left(\frac{s_i^2}{(s_i^2 + \sigma^2)^2} + \frac{\widetilde{s}_i^2}{(\widetilde{s}_i^2 + \sigma^2)^2} \right)- 2 \Tr(\b S_1 \U^\top \Uapprox \b S_2 \Vapproxpar^\top \Vpar). \label{eq:step-approx-mid-3}
\end{align}
Denote $\U^\top \Uapprox = \b I + \mathbf{\Delta}_{\U}$ and $\Vpar^\top \Vapproxpar = \b I  + \mathbf{\Delta}_{\V}$, then by \cref{assume:kl-general}, we have
\begin{align}
    \big( \mathbf{\Delta}_{\U}\big)_{ii} = \b u_i^\top \widetilde{\b u}_i - 1 \geq -\epsilon, \qquad  \big( \mathbf{\Delta}_{\V}\big)_{ii} = \b v_i^\top \widetilde{\b v}_i - 1 \geq -\epsilon.
\end{align}
Thus
\begin{align}
    \Tr(\b S_1 \U^\top \Uapprox \b S_2 \Vapproxpar^\top \Vpar)& = \Tr(\b S_1 (\b I +\mathbf{\Delta}_{\U} ) \b S_2 (\b I + \mathbf{\Delta}_{\V})^\top)\nonumber\\
    &= \Tr(\b S_1 \b S_2) + \Tr(\mathbf{\Delta}_{\U} \b S_2 \b S_1) + \Tr(\mathbf{\Delta}_{\V}^\top \b S_1 \b S_2)\nonumber\\
    &\geq (1-2\epsilon) \Tr(\b S_1 \b S_2) = (1-2\epsilon) \sum_{i \in[d_y]}\frac{s_i \cdot \widetilde{s}_i}{(s_i^2+\sigma^2 )( \widetilde{s}_i^2 + \sigma^2)},\label{eq:step-approx-mid-4}
\end{align}
where we neglect the second order term in the derivation of second equality.
Substitute \Cref{eq:step-approx-mid-4} into \Cref{eq:step-approx-mid-3}, we have
\begin{align}
     \|\Aapprox^\dagger - \A^\dagger\|_F^2 & \leq   \sum_{i\in[d_y]}\left(\frac{s_i}{(s_i^2+\sigma^2) } - \frac{\widetilde{s}_i}{(\widetilde{s}_i^2 + \sigma^2)}\right)^2  +\frac{4\epsilon s_i \widetilde{s}_i}{(s_i^2 +\sigma^2)(\widetilde{s}_i^2 +\sigma^2)}.\label{eq:step-approx-mid-5}
\end{align}
Substitute \Cref{eq:step-approx-mid-5} into \Cref{eq:step-approx-mid-2} and combine with \Cref{eq:step-approx-f13}, we have
\begin{align}\label{eq:kl-general-approx-quad} 
    f & = \sum_{i\in[4]}f_i \leq \left( 1+\frac{\|\A\|^2}{\sigma^2}\right) \left\{ \tau + \kappa_a d_y + \sigma^2 \sum_{i\in[d_y]}\left( \left(\frac{1}{\zeta_i} - \frac{1}{\widetilde{\zeta}_i}\right)^2 + \frac{4\epsilon}{\zeta_i \widetilde {\zeta}_i}\right)\right\},
\end{align}
where 
\begin{align}
    \kappa_a & = \max\left\{ \left(\frac{1}{\kappa_-} -1\right)^2,\,\left( \frac{1}{\kappa_+}-1\right)^2\right\},\nonumber\\
    \zeta_i & = s_i + \sigma^2/s_i, \qquad \widetilde{\zeta}_i = \widetilde{s}_i + \sigma^2/\widetilde{s}_i, \qquad \forall i\in[d_y].
\end{align}
By combining \cref{eq:kl-general-approx-quad} and \cref{eq:covariance-approx}, and substituting them into the closed-form expression for $\KLapprox$ given in \cref{prop:kl-closedform}, we obtain the bound for $\KLapprox$ as presented in \cref{eq:kl-general-approx}.
\item \Hybrid: first by \cref{lm:pseudo-inverse} and \cref{lm:pseudo-inverse-diff} we have
\begin{align}
    g_1 + g_4 &=\frac{1}{\sigma^2} \|\A  (\Anew^\dagger - \A^\dagger  )\A  \|_F^2 +  \|\A (\Anew^\dagger -  \A^\dagger  ) \|_F^2\nonumber\\
    &\leq  \left(1 +\frac{ \|\A\|^2}{\sigma^2}\right) \|\Aapprox \Aapprox^\dagger - \A  \A^\dagger\|_F^2 \leq \left(1 +\frac{ \|\A\|^2}{\sigma^2}\right)\tau. \label{eq:step-new-f14}
\end{align}
Define diagonal matrices $\b S_1 = \mathrm{Diag}\left[s_i/(s_i^2+\sigma^2)\right]_{i\in [d_y]}$ and $\b S_2 = \mathrm{Diag}\left[\widetilde{s}_i/(\widetilde{s}_i^2+\sigma^2)\right]_{i\in [d_y]}$. By Lemma~\ref{lm:pseudo-inverse}, we have
\begin{align}
   g_2 &= \sigma^2 \|\Anew^\dagger - \A^\dagger \|_F^2 =\sigma^2  \| \K \Aapprox^\dagger - \A^\dagger\|_F^2 = \sigma^2 \big(\|\K \Aapprox^\dagger \|_F^2 + \|\A^\dagger\|_F^2 - 2\Tr(\K \Aapprox^\dagger \A ^{+\top})\big)\nonumber\\
    &\leq  \sigma^2 \big[\kappa_+^2 \Tr(\b S_2^2) + \Tr(\b S_1^2) - 2 \Tr(\K \Vapproxpar \b S_2 \Uapprox^\top \U \b S_1 \Vpar^\top)\big]. \label{eq:step-new-2}
\end{align}
Let $\Uapprox^\top \U = \b I + \mathbf{\Delta}_{\U}$ and $\Vpar^\top \K  \Vapproxpar = \b I + \mathbf{\Delta}_{\V}$, then by \cref{assume:kl-general},
\begin{align}
    (\mathbf{\Delta}_{\U})_{ii} = \b u_i^\top \widetilde{\b u}_i - 1 \geq -\epsilon, \qquad (\mathbf{\Delta}_{\V})_{ii}  = \b v_i^\top \K \widetilde{\b v}_i  - 1 \geq \kappa_-(1-\epsilon) - 1.
 \end{align}
 Thus
 \begin{align}
      \Tr(\K \Vapproxpar \b S_2 \Uapprox^\top \U \b S_1 \Vpar^\top) &=  \Tr(\b S_2 (\b I + \mathbf{\Delta}_{\U}) \b S_1 (\b I+ \mathbf{\Delta}_{\V}^\top))\nonumber\\
     & = \Tr(\b S_2 \b S_1) +  \Tr(\b S_2 \b S_1 \mathbf{\Delta}_{\V}^\top) + \Tr(\b S_1 \b S_2 \mathbf{\Delta}_{\U})\nonumber\\
     &\geq (\kappa_- + \epsilon(\kappa_--1)) \Tr(\b S_2 \b S_1)\nonumber\\
     & =(\kappa_- + \epsilon(\kappa_--1)) \sum_{i\in[d_y]}\frac{1}{\zeta_i \widetilde{\zeta}_i}.\label{eq:step-new-3}
 \end{align}
 
 Substitute \Cref{eq:step-new-3} into \Cref{eq:step-new-2}, we have
 \begin{align}
    g_2 \leq \sigma^2  \left(\sum_{i \in[d_y]}\frac{\kappa_+^2}{\widetilde{\zeta}_i^2} + \frac{1}{\zeta_i^2}- \frac{2(\kappa_- + \epsilon(\kappa_--1))}{\zeta_i \widetilde{\zeta}_i}\right). \label{eq:step-new-f2}
 \end{align}
 Now we consider $g_3$:
 \begin{align}
     g_3& = \| (\Anew^\dagger -  \A^\dagger)\A \|_F^2 = \| \K \Aapprox^\dagger \Aapprox \K^{-1} - \A^\dagger \A \|_F^2\nonumber\\
     & \leq \|\K \Vapproxpar \Sapprox_\sigma\Vapproxpar^\top \K^{-1} - \Vpar \b S_\sigma\Vpar^\top \|_F^2\nonumber\\
     & \leq \| \Vpar \b S_\sigma\Vpar^\top- \Vapproxpar  \Sapprox_\sigma\Vapproxpar^\top\|_F^2 + \|\K \Vapproxpar  \Sapprox_\sigma\Vapproxpar^\top \K^{-1} - \Vapproxpar \Sapprox_\sigma\Vapproxpar^\top \|_F^2,\label{eq:step-new-4}
 \end{align}
where
\begin{align}
    \|\K \Vapproxpar \Vapproxpar^\top \K^{-1} - \Vapproxpar \Vapproxpar^\top \|_F^2 &= \|\K \Vapproxpar \Vapproxpar^\top \K^{-1}] \|_F^2 + \| \Vapproxpar \Vapproxpar^\top\|_F^2 - 2 \Tr(\K \Vapproxpar \Vapproxpar^\top \K^{-1} \Vapproxpar \Vapproxpar^\top) \nonumber\\
    & \leq \left(\frac{\kappa_+^2}{\kappa_-^2}+ 1 -\frac{2\kappa_-}{\kappa_+}\right) \| \Vapproxpar \Vapproxpar^\top\|_F^2 = \left(\frac{\kappa_+^2}{\kappa_-^2}+ 1 -\frac{2\kappa_-}{\kappa_+}\right)d_y.\label{eq:step-new-5}
\end{align}
Combining \cref{eq:step-new-5}, \cref{assume:kl-general} and \cref{eq:step-new-4}, we cane get
\begin{align}\label{eq:step-new-f3}
    g_3 \leq \tau + \left(\frac{\kappa_+^2}{\kappa_-^2}+ 1 -\frac{2\kappa_-}{\kappa_+}\right)d_y.
\end{align}

Now combine \Cref{eq:step-new-f14,eq:step-new-f2,eq:step-new-f3}, we have
\begin{align}\label{eq:kl-general-new-quad}
    g = \sum_{i\in[4]}g_i \leq \left(2+\frac{\|\A\|^2}{\sigma^2} \right)\tau + \kappa_l d_y + \sigma^2  \sum_{i \in[d_y]}\left(\frac{\kappa_+^2}{\widetilde{\zeta}_i^2} + \frac{1}{\zeta_i^2}- \frac{2(\kappa_- + \epsilon(\kappa_--1))}{\zeta_i \widetilde{\zeta}_i}\right),
\end{align}
where
\begin{align}
    \kappa_l = \frac{\kappa_+^2}{\kappa_-^2} -\frac{2\kappa_-}{\kappa_+} +1.
\end{align}
By combining \cref{eq:kl-general-new-quad} and \cref{eq:covariance-new}, and substituting them into the closed-form expression for $\KLnew$ given in \cref{prop:kl-closedform}, we obtain the bound for $\KLnew$ as presented in \cref{eq:kl-general-new}.
\end{enumerate}
\end{proof}

\section{PROOFS OF SECTION 4}\label{sec:mcmc-proof}
We begin this section by presenting the supporting lemma in \cref{sec:mcmc-lemma}, followed by the proof of \cref{thm:mix-general} in \cref{sec:mcmc-mix-general}. Finally, we provide the proof of \cref{thm:mix-normal} in \cref{sec:mcmc-mix-simple}.
For the ease of notation, given $s \in (0,1/2)$, we define constant $r(s)$ by
\begin{align}
    r(s)  = 2 + 2 \max\left\{\frac{-\log^{1/4}(s)}{d^{1/4}}, \frac{-\log^{1/2}(s)}{d^{1/2}} \right\}.
\end{align}
\subsection{Supporting lemma}\label{sec:mcmc-lemma}
Our proof is based on the theory developed in \cite{log-concave1,log-concave2}. We show the general mixing time bound of the strongly log-concave distribution for IMH in \cref{prop:mix-bound-logconcave}. We first denote $\pi$ as the ground truth posterior, and $\widehat{\pi}$ denote the proposal distribution in the MH step. Define the log-weight function by
\[ \widehat{w}(x) = \log\frac{\pi(x)}{\widehat{\pi}(x)}.    \]

\begin{assumption}\label{assume:mcmc-pi}
$ $
    \begin{itemize}[leftmargin=*]
    \item $x^* = \argmax \log \pi(x)$,  $\widehat{x}^* = \argmax \log \widehat{\pi}(x)$.
        \item $\pi$ satisfies isoperimetric inequality with isoperimetric constant  $\psi(\pi)$ i.e., for any partition $\{\mathsf{S}_1, \mathsf{S}_2, \mathsf{S}_3\}$ of $\mathbb{R}^d$, we have
    \begin{align}
        \pi(\mathsf{S}_3) \geq \psi(\pi) \mathrm{dist}(\mathsf{S}_1, \mathsf{S}_2) \pi (\mathsf{S}_1, \mathsf{S}_2),
    \end{align}
    where $\mathrm{dist}(\mathsf{S}_1, \mathsf{S}_2) = \inf\{\|x-y \|: x\in\mathsf{S}_1, y\in\mathsf{S}_2\}$.
    \item For all $R>0$, there exists $C_a(R) \geq 0$ such that for all $(x,x') \in \mathsf{Ball}(x^*,R)$,  $\|\widehat{w}(x) - \widehat{w}(x')\|\leq C_R\|x-x'\|$.
    \end{itemize}
\end{assumption}

\begin{assumption}[$\alpha$, $s$]\label{assume:mcmc-proposal}
    There exists $\Delta\in(0,1)$, $\delta_\Delta >0$ and $R>0$ satisfying
    \begin{align}\label{eq:assume-cond-1}
        \int_{B_\Delta} \pi(dx) \widehat{\pi}(dy) \geq 1- \delta_\Delta,
    \end{align}
    where 
    \begin{align}
        B_\Delta = \{(x,y) \in \mathsf{Ball}(x^*,R) \times \mathsf{Ball}(x^*,R): \|x-y \|\leq R_\Delta\}, \quad \text{where} \quad  R_\Delta = \frac{-\log(\Delta)}{C_R},
    \end{align}
    and 
    \begin{align}\label{eq:assume-cond-2}
        \frac{1}{1 - \alpha } \delta_\Delta \leq \min \left( \frac{s}{4}, \frac{(2\alpha - 1) \psi(\pi) s}{64 C_R}\right).
    \end{align}
\end{assumption}

\begin{proposition}[Corollary E.5 in \cite{log-concave2}]\label{prop:mix-bound-logconcave}
Let $\alpha(0,1/2)$, $\epsilon\in(0,1)$ and $g$ a $\beta$-warm initial distribution with respect to $\pi$. Assume that \cref{assume:mcmc-pi} and \cref{assume:mcmc-proposal} hold, then we have the following upper bound on the mixing time
\begin{align}
    \tau_{mix}(g,\epsilon) \leq \frac{8}{(1-\alpha^2)} \log\left(\frac{2\beta}{\epsilon}\right) \max\left(1, \frac{64^2 C_R^2}{\psi(\pi)^2 (2 \alpha -1)^2}\right).
\end{align}
    
\end{proposition}

\subsection{Proof of \cref{thm:mix-general}}\label{sec:mcmc-mix-general}
\begin{proof}
Denote $s = \epsilon/2\beta$, let $c_1>0$ be a constant such that $c_1s \in(0,1/2)$. By Lemma 1 in \cite{log-concave1}, we have $\pi(\mathsf{Ball}(x^*, R_1))\geq 1-c_1s$, where
\begin{align}
    R_1 = \sqrt{\frac{d}{m}} r(c_1 s).
\end{align}

\begin{enumerate}[leftmargin=*]
\item \ApproxMH: since $-\log p(x)$ is $m$-strongly convex, $-\log \piapprox(x)$ (\cref{eq:approx-post}) is also $m$-strongly convex. Then let $c_2>0$ be a constant s.t. $c_2 s \in(0,1/2)$, we have $\piapprox(\mathsf{Ball}(\xapprox^*, R_2))\geq 1-c_2 s$, where
\begin{align}
    R_2 = \sqrt{\frac{d}{m}} r(c_2 s).
\end{align}
Define $R_{2,a} = R_2 + \|x^* -\xapprox^* \|$, $R_\Delta = R_1 + R_{2,a}$ and $R = \max(R_1, R_{2,a})$. Then for any $(x,y) \in \mathbf{Ball}(x^*, R_1) \times \mathbf{Ball}(\xapprox^*, R_{2,a})$, we have
\begin{align}
    &\|x-y\|\leq \|x^* - \xapprox^*\| + R_1 + R_2 = R_\Delta,\nonumber\\
    &\|x-x^*\| \leq R_1 \leq R, \qquad \|y-x^*\| \leq R_{2,a} + \|x^* - \xapprox^*\| \leq R\nonumber.
\end{align}
Thus for the set $B_\Delta$ defined in \cref{assume:mcmc-proposal}, we have
\begin{align}
    \mathbf{Ball}(x^*, R_1) \times \mathbf{Ball}(\xapprox^*, R_{2,a}) \subset B_\Delta.
\end{align}
Thus the condition of \cref{eq:assume-cond-1} in \cref{assume:mcmc-proposal} is satisfied since
\begin{align}
    \int_{B_\Delta} \pi(dx) \piapprox(dy) &\geq \int_{\mathbf{Ball}(x^*, R) \times \mathbf{Ball}(\xapprox^*, R_{2,a})} \pi(dx) \piapprox (dy)\nonumber\\
    &\geq (1 - c_1 s)(1 - c_2 s) \geq 1- (c1+c2) s =: 1-\delta_\Delta. \label{eq:mcmc-1}
\end{align}
Now we need to determine $c_1$ and $c_2$ such that \cref{eq:assume-cond-2} is satisfied:
\begin{align}\label{eq:mcmc-2}
    \frac{\delta_\Delta}{1-\alpha} \leq \min \left(\frac{s}{4},\frac{(2\alpha - 1)s}{64 C_R}\psi(\pi)\right) \Longleftrightarrow \begin{cases}
        \delta_\Delta \leq \frac{1-\alpha}{4}s\\
        \delta_\Delta \leq \frac{(1-\alpha)(2\alpha-1)}{64 C_R}\psi(\pi) s.
    \end{cases}
\end{align}
Let $\alpha = 3/4$, by \cref{eq:mcmc-2} and \cref{eq:mcmc-1}, it is enough if the following conditions are satisfied: 
\begin{align}\label{eq:mcmc-3}
    c_1+ c_2 \leq \frac{1}{16},\qquad C_R \leq \frac{\psi(\pi)}{512 (c_1+c_2)}.
\end{align}
By Theorem 4.4 in \cite{Cousins}, a $m$-strongly log concave distribution $\pi$ is isoperimetric with the isoperimetric constant $\psi(\pi) =\log (2\sqrt{m})$. Let $c_1 = 1/17$, $c_2 = 1/16 - 1/17$. Thus  if 
\begin{align}
    C_a(R_a) \leq \frac{\log 2 \sqrt{m}}{32},
\end{align}
where 
\begin{align}
     R_a = \max\left(\sqrt{\frac{d}{m}}r\left(\frac{\epsilon}{17 \beta}\right),\sqrt{\frac{d}{m}}r\left(\frac{\epsilon}{272\beta}\right) + \|x^* - \xapprox ^* \|\right),
\end{align}
conditions of \eqref{eq:mcmc-3}  are satisfied and thus \cref{assume:mcmc-pi} and \cref{assume:mcmc-proposal} are satisfied. Then by \cref{prop:mix-bound-logconcave}, we have
\begin{align}
 \mixapprox(\epsilon) \leq 128 \log\left(\frac{2\beta}{\epsilon}\right) \max\left(1, \frac{128^2C_a(R_a)^2}{(\log 2)^2 m}\right).
\end{align}
\item \HybridMH: the result can be proved by an argument similar to the proof for \ApproxMH. By \cref{eq:hybrid-post}, we have $\pinew(x) \propto q(y-\A x) p(\Fapprox^{-1} \F x)$. It is easy to see that $\pinew(x)$ is $m\sigma_{\min}^2(\Fapprox^{-1} \F)$. Thus we have $\pinew(\mathsf{Ball}(\xnew^*, R_2)) \geq 1 - c_2 s$, where
\begin{align}
    R_2 = \sqrt{\frac{d}{m}}\frac{1}{\sigma_{\min}(\Fapprox^{-1} \F)} r(c_2 s).
\end{align}
Then if $C_l(R_l) \leq \frac{\log 2 \sqrt{m}}{32}$, where
\begin{align}
    R_l =  \max\left(\sqrt{\frac{d}{m}}r\left(\frac{\epsilon}{17 \beta}\right),\sqrt{\frac{d}{m}} \frac{1}{\sigma_{\min}(\Fapprox^{-1} \F)} r \left(\frac{\epsilon}{272\beta}\right)  + \|x^* - \xnew ^* \|\right),
\end{align}
conditions of \cref{eq:mcmc-3} are satisfied and the  bound for mixing time of \HybridMH is proved.
\end{enumerate}

\end{proof}

\subsection{Proof of \cref{thm:mix-normal}}\label{sec:mcmc-mix-simple}
\begin{proof}
Without loss of generality, we assume that the posterior mode $x^*$ is located at the origin. The key step is to establish the local Lipschitz continuity of the log-weight function on the ball $\mathsf{Ball}(x^*,\Rapprox)$.

\begin{enumerate}[leftmargin=*]
\item \ApproxMH: the log-weight function has the following form:
\begin{align}
    \wapprox(x) = \log \frac{\pi(x)}{\piapprox(x)} &= -\frac{1}{2\sigma^2} \big[\| y-\A x\|^2 - \|y - \Aapprox x \|^2\big]\nonumber\\
    &= -\frac{1}{2\sigma^2}\big [x^\top (\A - \Aapprox)^\top \big(2 y - (\A +\Aapprox )x\big)\big]\nonumber\\
    &=  -\frac{1}{2\sigma^2}\big[x^\top \Cmatapprox x + 2 x^\top \Delta \A^\top y\big],
\end{align}
where we define 
\begin{align}
    \Cmatapprox := \A^\top \A - \Aapprox^\top \Aapprox = \mathrm{Diag}\big(\underbrace{\cdots, (1-\alpha_i^2) s_i^2, \cdots}_{\text{first } d_y \text{ coordinates}}, 0, \cdots\big), \quad \Delta \A := \A - \Aapprox.
\end{align}
Thus for any $x_1, x_2\in \mathsf{Ball}(x^*, \Rapprox)$, we have
\begin{align}\label{eq:pf-approx-1}
    |\wapprox(x_1) - \wapprox(x_2)| &= \frac{1}{2\sigma^2}\big|x_1^\top \Cmatapprox x_1 + 2 x_1^\top \Delta \A^\top y - x_2^\top \Cmatapprox x_2 - 2 x_2^\top \Delta \A^\top y\big|\nonumber\\
    & \leq\frac{1}{2\sigma^2}\big|x_1^\top \Cmatapprox x_1 - x_2^\top \Cmatapprox x_2 \big| + \frac{1}{\sigma^2}\big| (x_1-x_2)^\top \Delta \A^\top y \big|.
\end{align}
For the first term of the last inequality of \Cref{eq:pf-approx-1}, we have
\begin{align}\label{eq:pf-approx-2}
    \left|x_1^\top \Cmatapprox x_1 - x_2^\top \Cmatapprox x_2 \right| &= |(x_1 - x_2)^\top\Cmatapprox (x_1+x_2)|\nonumber\\
    &\leq \|\Cmatapprox\| \|x_1+x_2\| \|x_1-x_2\| \leq 2R_a \left(\max_{i\in[d_y]} |1-\alpha_i^2| s_i^2 \right)\|x_1-x_2\|.
\end{align}

For the second term of last inequality of \Cref{eq:pf-approx-1}, we have
\begin{align}\label{eq:pf-approx-3}
    \left| (x_1-x_2)^\top \Delta \A^\top y \right| \leq \left(\max_{i\in[d_y]} |1-\alpha_i| s_i \right)\|y\| \|x_1-x_2\|.
\end{align}
Substitute \Cref{eq:pf-approx-2,eq:pf-approx-3} into \Cref{eq:pf-approx-1}, we have
\begin{align}
 |\wapprox(x_1) - \wapprox(x_2)| \leq \frac{1}{2\sigma^2}\left\{2R_a \left(\max_{i\in[d_y]} |1-\alpha_i^2| s_i^2 \right) +  \left(\max_{i\in[d_y]} |1-\alpha_i| s_i \right)\|y\|\right\} \|x_1- x_2\|.
\end{align}
By \Cref{thm:mix-general}, $\Rapprox\sim 2\sqrt{d/m}$ when $d$ is large. Thus, the Lipschitz constant scale as the following when $d$ is large,
\begin{align}
    \Capprox({\Rapprox}) \sim \frac{d}{m}\max_{i\in[d_y]} |1-\alpha_i^2|\frac{s_i^2}{\sigma^2}
\end{align}
\item \HybridMH: the log-weight function has the following form
\begin{align}
    \wnew(x) &= \log\frac{\pi(x)}{\bpinew(x)}= \log \frac{p(x)}{p(\Fapprox^{-1} \F x)}\nonumber\\
    &= -\frac{1}{2}\big[\|x\|^2 - \|\Fapprox^{-1} \F x\|^2\big].
\end{align}
Then
\begin{align}
    |\wnew(x_1) - \wnew(x_2)|&= \frac{1}{2}\left| \|x_1\|^2 - \|x_2\|^2 - \|\Fapprox^{-1} \F x_1\|^2 + \| \Fapprox^{-1} \F x_2\|^2\right|\nonumber\\
    &= \frac{1}{2}| x_1^\top \Cmatnew x_1 - x_2 ^\top \Cmatnew x_2|,
\end{align}
where we define
\begin{align}
    \Cmatnew := \b I - (\Fapprox^{-1} \F)^\top (\Fapprox^{-1} \F) = \mathrm{Diag}\left(1 - \frac{1}{\alpha^2} \right)_{i\in[d]}.
\end{align}
Thus
\begin{align}
     |\wnew(x_1) - \wnew(x_2)| &= \frac{1}{2}\left|(x_1-x_2)^\top \Cmatnew (x_1+x_2) \right|\nonumber\\
     &\leq \frac{1}{2}\|\Cmatnew\| \|x_1+x_2\|  \|x_1-x_2\|\leq R_l \left(\max_{i\in[d]}\left|1-\frac{1}{\alpha_i^2}\right|\right) \|x_1-x_2\|.
\end{align}
By \Cref{thm:mix-general}, $\Rnew\sim \sqrt{d/m}$ when $d$ is large. Thus, the Lipschitz constant scale as the following when $d$ is large,
\begin{align}
    \Cnew({\Rnew}) \sim  \frac{d}{m} \max_{i\in[d]}\left|1-\frac{1}{\alpha_i^2}\right|.
\end{align}
Substitute results for $\Capprox({\Rapprox})$ and $\Cnew({\Cnew})$ into \Cref{thm:mix-general}, we can get the results for mixing time.
\end{enumerate}    
\end{proof}

\section{ADDITIONAL EXPERIMENTAL DETAILS}\label{sec:add-experiments}

All experiments were conducted on the Vista 600 Grace Hopper (GH) nodes at the Texas Advanced Computing Center (TACC).

\subsection{Sensitivity test in \cref{sec:KL}}
\begin{figure*}[t]
\small
    \centering
\begin{tikzpicture}
\node[inner sep=0pt] (a1) at (0,0) {\includegraphics[width=3.3cm]{figures/plots-kl/snr_abskl.pdf}};
\node[inner sep=0pt] (a2) at (3.3,0) {\includegraphics[width=3.3cm]{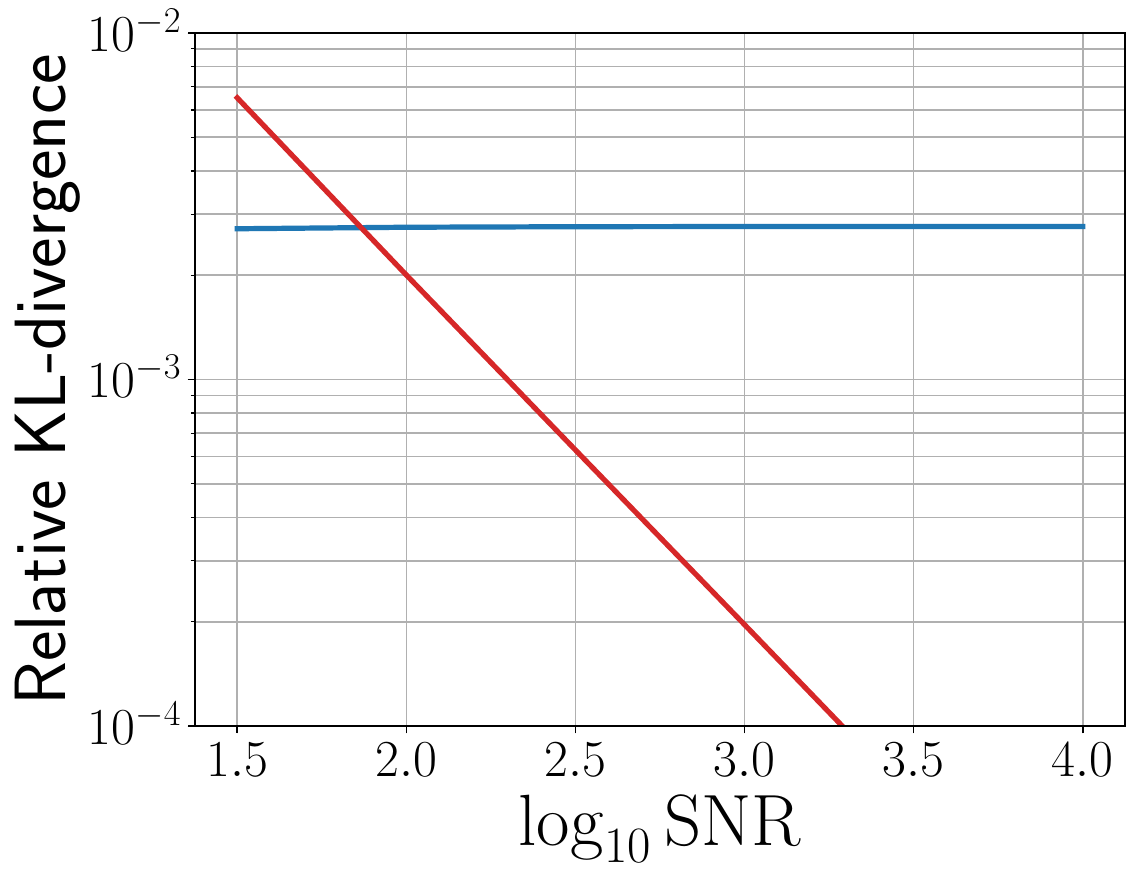}};
\node[inner sep=0pt] (b1) at (6.9,0) {\includegraphics[width=3.3cm]{figures/plots-kl/df_abskl.pdf}};
\node[inner sep=0pt] (b2) at (10.2,0) {\includegraphics[width=3.3cm]{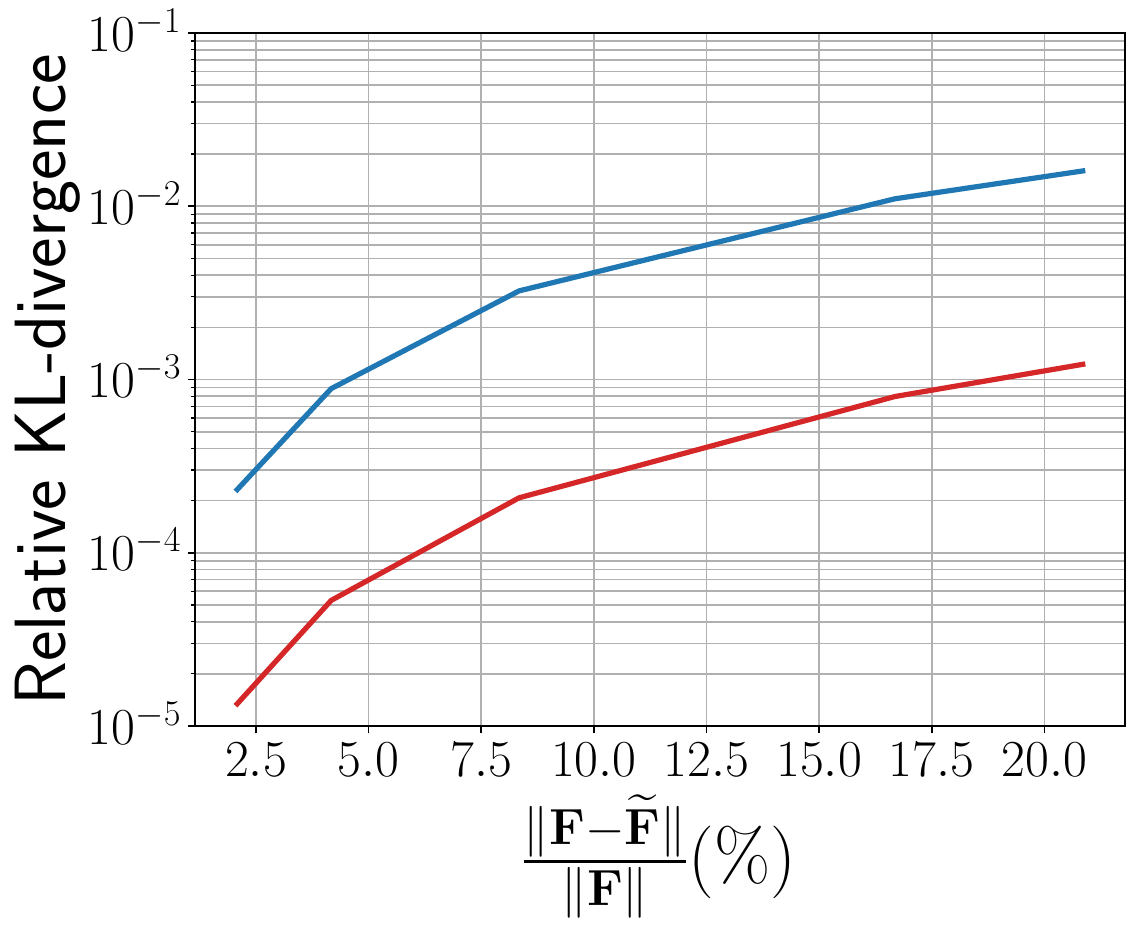}};
\node[inner sep=0pt] (c1) at (0,-3.3) {\includegraphics[width=3.3cm]{figures/plots-kl/ratio_abskl.pdf}};
\node[inner sep=0pt] (c2) at (3.3,-3.3) {\includegraphics[width=3.3cm]{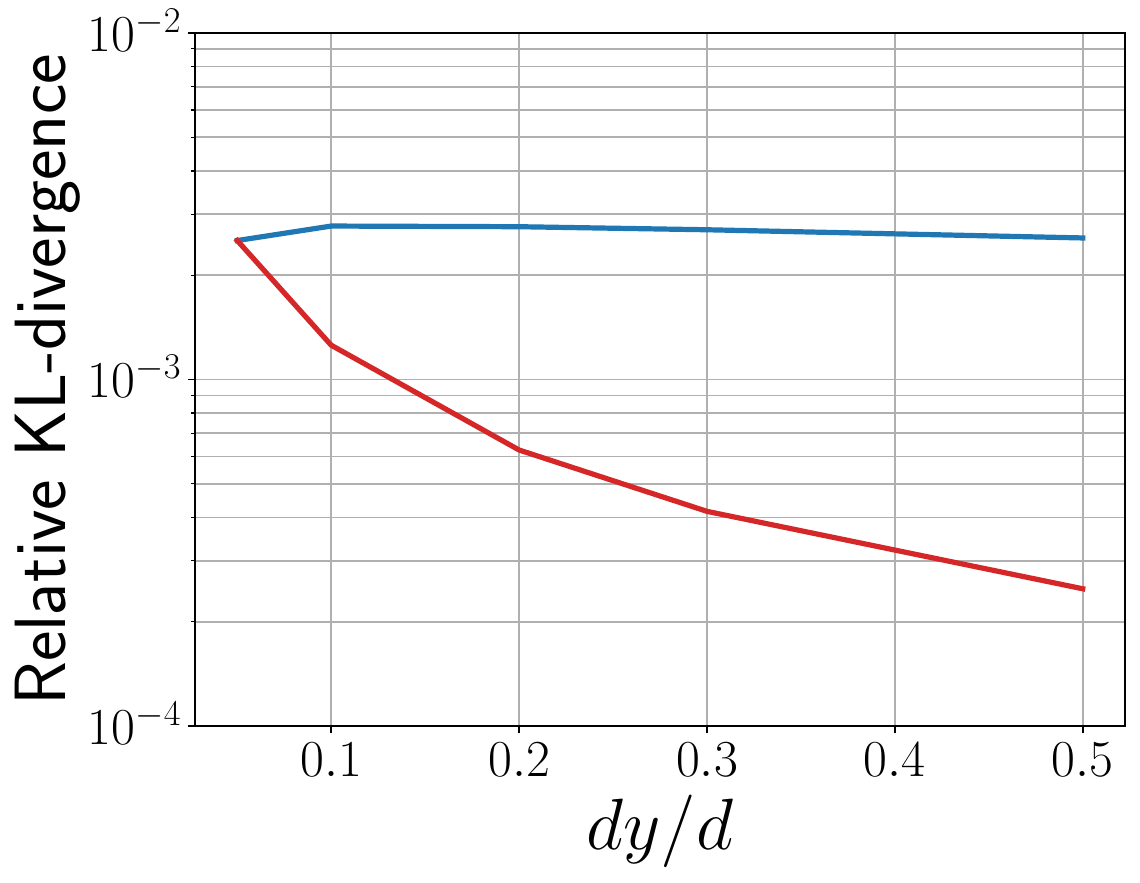}};
\node[inner sep=0pt] (d1) at (6.9,-3.3) {\includegraphics[width=3.3cm]{figures/plots-kl/d_abskl.pdf}};
\node[inner sep=0pt] (d2) at (10.2,-3.3) {\includegraphics[width=3.3cm]{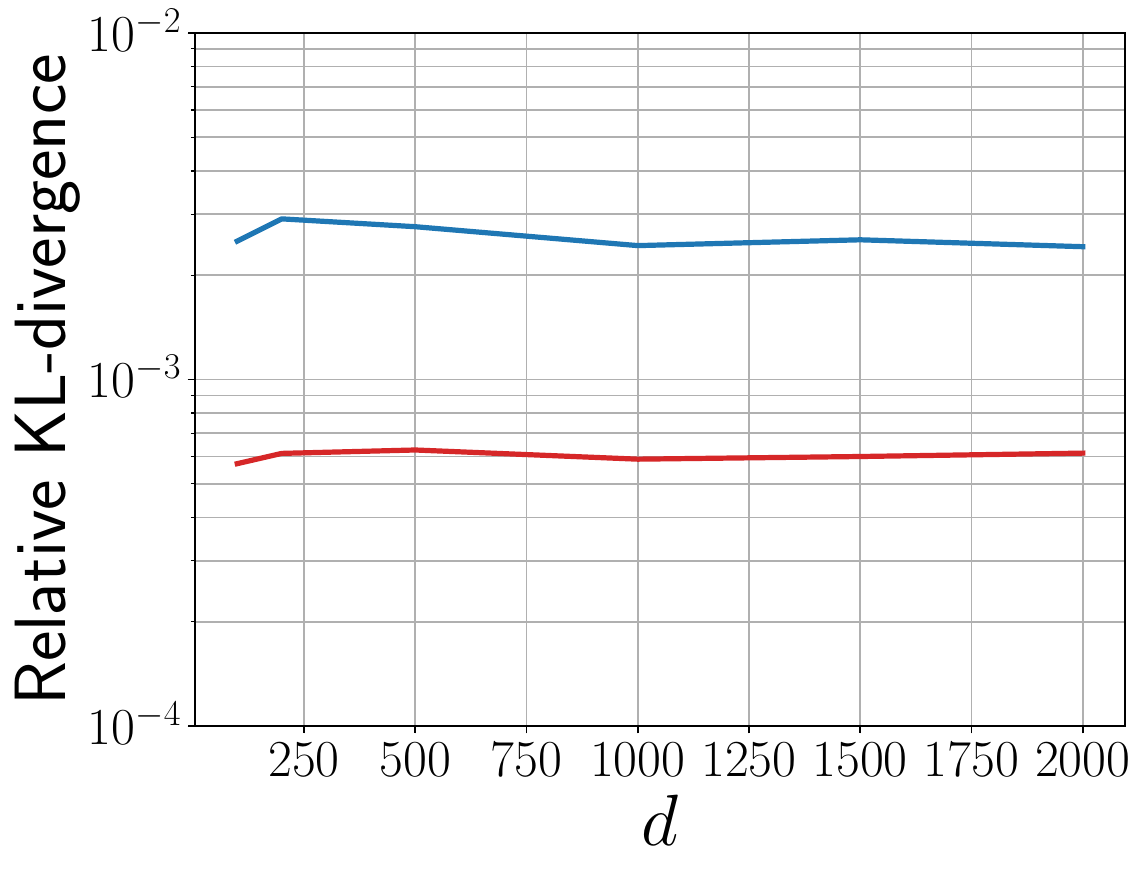}};
\draw[line width=0.5mm,dotted] (-1.7, -1.4) -- (12, -1.4);
\draw[line width=0.5mm,dotted] (5.15, 2) -- (5.15, -4.7);
\node[draw,rounded corners,fill=black!10] at (1.7, 1.7) {Noise level};
\node[draw,rounded corners,fill=black!10] at (8.7, 1.7) {Spectral error};
\node[draw,rounded corners,fill=black!10] at (1.7, -1.7) {Observation ratio};
\node[draw,rounded corners,fill=black!10] at (8.7, -1.7) {Dimension};
\draw (-0.5, 2.1) rectangle (10.65, 2.7);
\draw[tabblue, thick] (0.,2.4) -- (.8,2.4);
\node[] at (3, 2.4) {\Approx ($\KLapprox$)} ;
\draw[tabred, thick] (5.5,2.4) -- (6.3,2.4);
\node[] at (8.5, 2.4) {\Hybrid ($\KLnew$)} ;
\draw[line width=0.5mm,dotted] (-1.7,-4.7) rectangle (12, 2);
\end{tikzpicture}
    \caption{Sensitivity test results for the expected KL-divergence of \Approx and \Hybrid relative to \Exact.}
    \label{fig:kl-experiments-relative}
\end{figure*}

The design of the sensitivity tests for comparing the KL-divergence between \Approx, \Hybrid with \Exact is presented in \cref{tab:kl-experiments}. In \cref{fig:kl-experiments-relative} shows both the absolute KL-divergence and relative KL-divergence results for the sensitivity tests.

\begin{table*}[t]
\small
    \centering
    \caption{Experimental setup for comparing \Approx and \Hybrid with \Exact.}
    \label{tab:kl-experiments}
    \begin{tabular}{ccccc}
      \toprule  
  Effects  & $\log_{10}\mathrm{SNR}$ & $\frac{\| \F - \Fapprox\|}{\| \F\|}$  & $d_y/d$ & $d$\\
      \midrule
  Noise level     & 0.5 \textemdash 4.0 & 6\% & 0.2  & 500\\
 Spectral error & 2.5& 2\% \textemdash 21\%& 0.2&500 \\
  Observation ratio &2.5  &6\% &0.05 \textemdash 0.5 & 500\\
  Dimension &2.5 &6\% &0.2 & 100 \textemdash 2000\\
         \bottomrule
    \end{tabular}
\end{table*}


\begin{table}[t]
\centering
\caption{Relative mean error comparisons between different MCMC sampling methods.
Each entry shows: (\# exact forward/adjoint solves, relative mean error). 
The approximate operator used in Latent-2stage, Approx-2stage, and Latent-IMH has approximation error 
$\|\bf I - \widetilde{F}^{-1} F\| = 2.4\%$ (top) and $\|\bf I - \widetilde{F}^{-1} F\|  =9.0\%$ (bottom).}
\label{tab:compare_delay}
\renewcommand{\arraystretch}{1.25}


\begin{tabular}{lcccc}
\toprule
\textbf{Latent-2stage} & \textbf{Approx-2stage} & \textbf{MALA} & \textbf{NUTS} & \textbf{Latent-IMH} \\
\midrule
(1.2$\times 10^{5}$, 0.37) & (1.2$\times 10^{5}$, 0.38) & (5.0$\times 10^{5}$, 0.36) & (1.8$\times 10^{5}$, 0.20) & (1.0$\times 10^{3}$, 0.32) \\
(2.4$\times 10^{6}$, 0.08) & (2.4$\times 10^{6}$, 0.088) & (1.0$\times 10^{7}$, 0.08) & (7.5$\times 10^{5}$, 0.066) & (2.0$\times 10^{4}$, 0.07) \\
(4.8$\times 10^{6}$, 0.057) & (4.8$\times 10^{6}$, 0.062) & (2.0$\times 10^{7}$, 0.055) & (1.3$\times 10^{6}$, 0.044) & (2.0$\times 10^{5}$, 0.023) \\
\bottomrule
\end{tabular}

\vspace{1.2em}


\begin{tabular}{lccc}
\toprule
\textbf{Latent-2stage} & \textbf{Approx-2stage} & \textbf{Latent-IMH} \\
\midrule
(1.2$\times 10^{5}$, 0.36) & (1.3$\times 10^{5}$, 0.49) & (5.0$\times 10^{3}$, 0.59) \\
(2.4$\times 10^{6}$, 0.084) & (2.6$\times 10^{6}$, 0.11) & (1.0$\times 10^{5}$, 0.12) \\
(4.9$\times 10^{6}$, 0.059) & (5.2$\times 10^{6}$, 0.088) & (5.0$\times 10^{5}$, 0.056) \\
\bottomrule
\end{tabular}
\end{table}

\begin{figure*}[t]
\scriptsize
    \centering
\begin{tikzpicture}
\node[inner sep=0pt] (a0) at (0,-3.7) {\includegraphics[width=9cm]{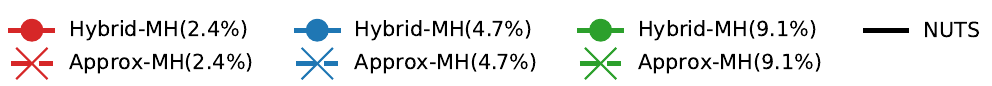}};
\node[inner sep=0pt] (a1) at (0,-5.3) {\includegraphics[width=10cm]{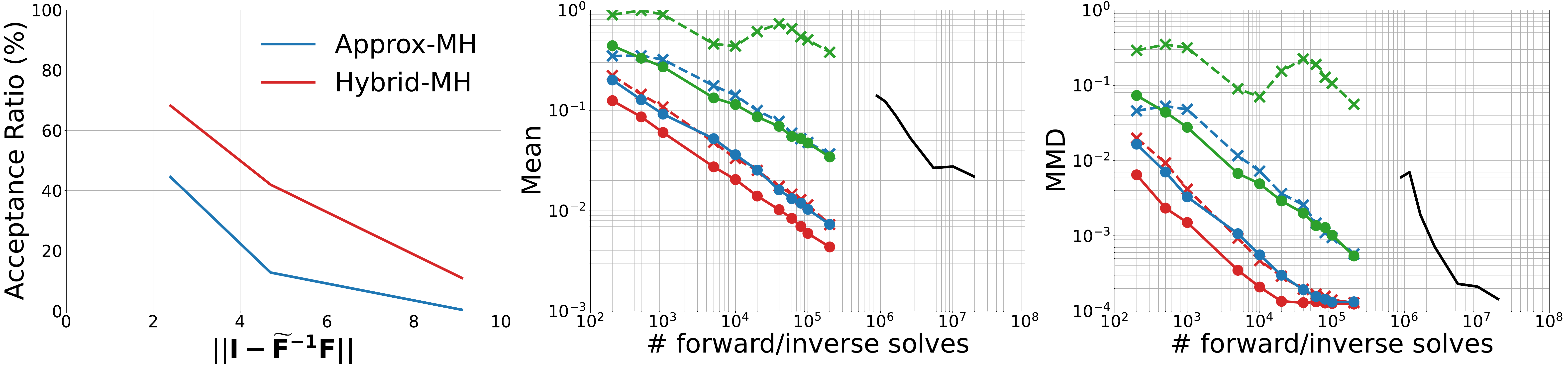}};
\end{tikzpicture}
\caption{Sample efficiency comparison for ill-conditioned Gaussian prior $p(x) \sim \mathcal{N}(\b 0,\b \bSigma)$,  where $\kappa(\bSigma) \approx 1000$. The results in the plots reflect the average of 5 independent runs for each test. The numbers in parentheses in the top legend for \ApproxMH and \HybridMH indicate the spectral error $\|\b I - \Fapprox^{-1} \F\|_2$.
  The numbers in parentheses above the histogram plots denote the total number of forward and inverse solves of $\F$ required by each sampler.}
    \label{fig:mcmc-ill-condition}
\end{figure*}

{
\begin{figure}[!htbp]
    \centering
\begin{tikzpicture}
    \node[inner sep=0pt] (a) at (0,0) {\includegraphics[height=2cm]{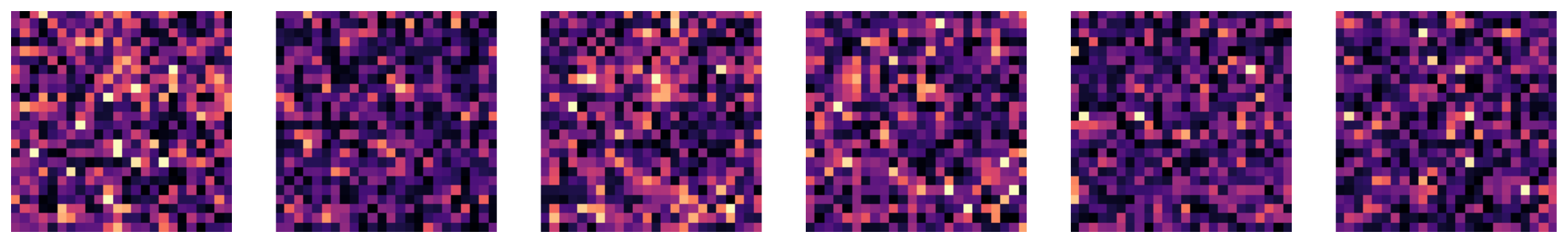}};
    \node[inner sep=0pt] (b) at (0,-2.2) {\includegraphics[height=2cm]{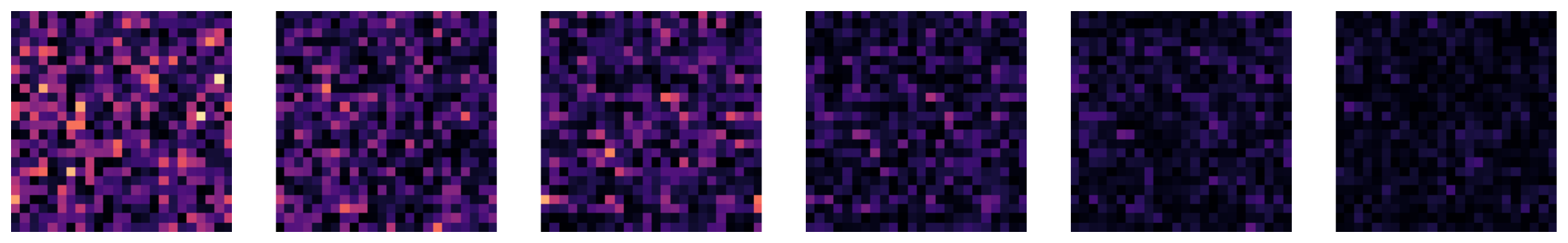}};
    \node[inner sep=0pt] (c) at (7.5, -1.1) {\includegraphics[height=4.2cm]{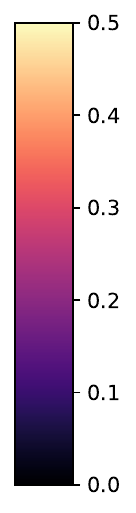}};
\node[rotate=90,align=center] at (-6.8,0) {\color{tabblue}\small \textbf{\ApproxMH}};
\node[rotate=90,align=center] at (-6.8,-2.2) {\color{tabred}\small \textbf{\HybridMH}};
\node[] (A1) at (-5.5,1.3) {\small 100 steps};
\node[] (A2) at (-3.3,1.3) {\small 200 steps};
\node[] (A3) at (-1.1,1.3) {\small 500 steps};
\node[] (A4) at (1.1,1.3) {\small 1,000 steps};
\node[] (A5) at (3.3,1.3) {\small 2,000 steps};
\node[] (A6) at (5.5,1.3) {\small 10,000 steps};
\end{tikzpicture}
    \caption{Convergence of the mean error map for the scattering problem test. }
    \label{fig:helmholtz-mean-err}
\end{figure}
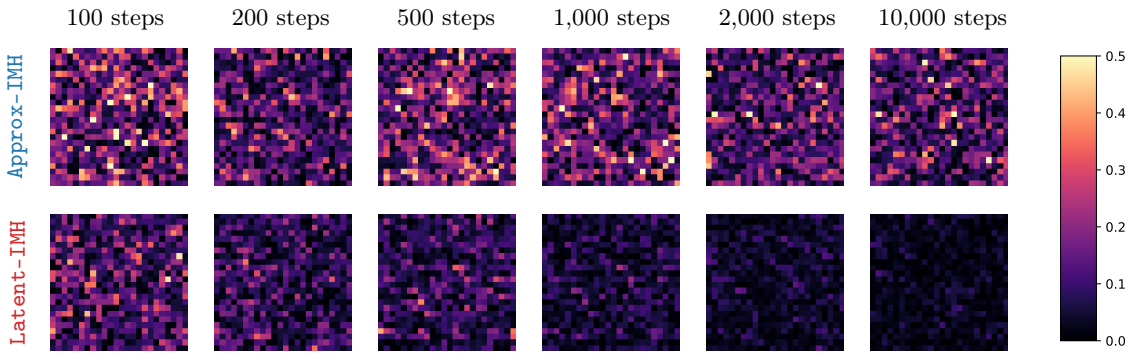

\begin{figure}[!htbp]
    \centering
\begin{tikzpicture}
    \node[inner sep=0pt] (a) at (0,0) {\includegraphics[height=2cm]{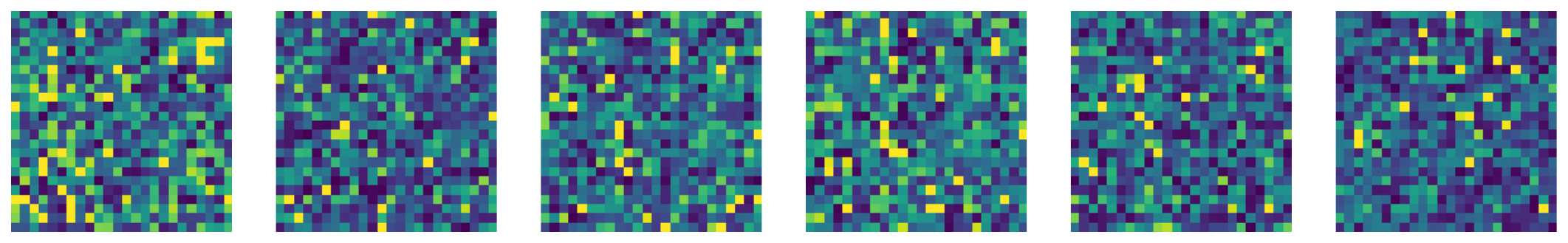}};
    \node[inner sep=0pt] (b) at (0,-2.2) {\includegraphics[height=2cm]{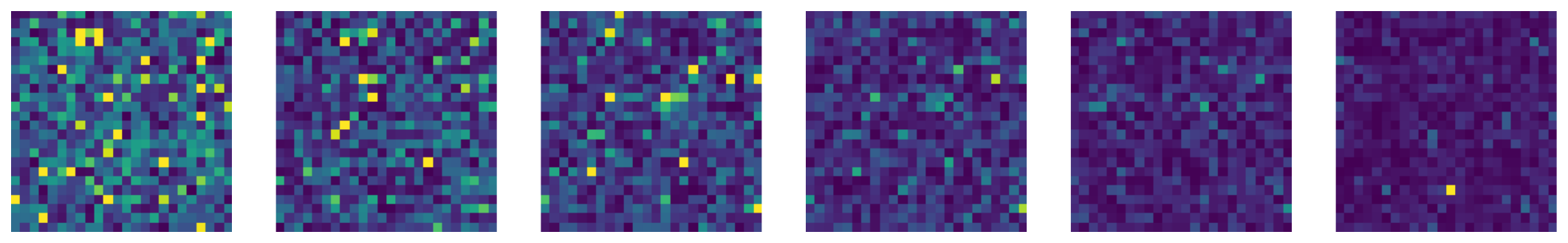}};
    \node[inner sep=0pt] (c) at (7.5, -1.1) {\includegraphics[height=4.2cm]{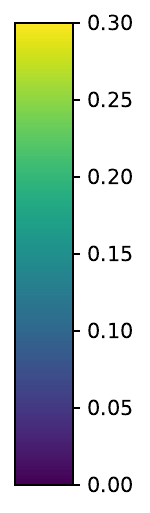}};
\node[rotate=90,align=center] at (-6.8,0) {\color{tabblue}\small \textbf{\ApproxMH}};
\node[rotate=90,align=center] at (-6.8,-2.2) {\color{tabred}\small \textbf{\HybridMH}};
\node[] (A1) at (-5.5,1.3) {\small 100 steps};
\node[] (A2) at (-3.3,1.3) {\small 200 steps};
\node[] (A3) at (-1.1,1.3) {\small 500 steps};
\node[] (A4) at (1.1,1.3) {\small 1,000 steps};
\node[] (A5) at (3.3,1.3) {\small 2,000 steps};
\node[] (A6) at (5.5,1.3) {\small 10,000 steps};
\end{tikzpicture}
    \caption{Convergence of the variance error map for the scattering problem test. }
    \label{fig:helmholtz-var-err}
\end{figure}
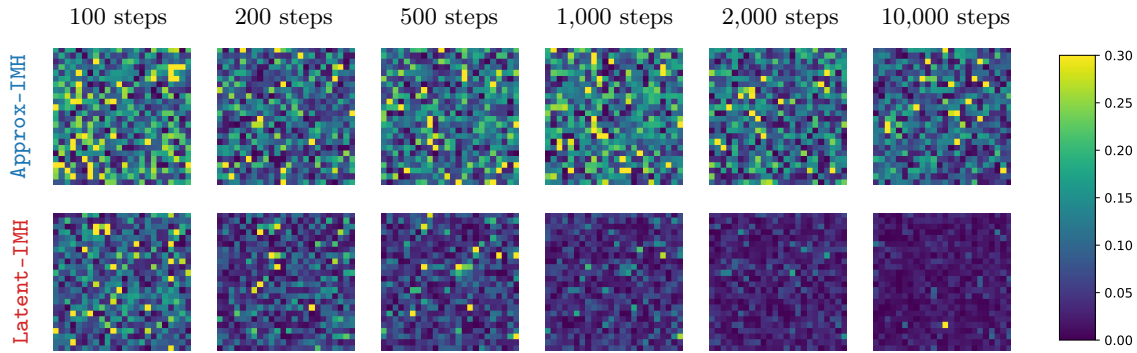
}

\subsection{Additional details for \cref{sec:mcmc-experiment}}\label{sec:add-results}

\paragraph{Comparison with delayed-acceptance MCMC.}
We compare the sampling efficiency of different MCMC methods under two approximate operators (with varying levels of accuracy) in \cref{tab:compare_delay}. Both \texttt{Latent-2stage} and \texttt{Approx-2stage} improve efficiency relative to standard MALA. For example, to achieve comparable accuracy, \texttt{Latent-2stage} requires only about $25\%$ of the exact forward/inverse solves used by MALA. However, despite these improvements, both two-stage methods still underperform compared to Latent-IMH.

\paragraph{Evaluation Metric -- Maximum Mean Discrepancy (MMD).} MMD is a kernel-based distance between two distributions $P$ and $Q$, defined as
\[
\mathrm{MMD}^2(P,Q) = \mathbb{E}_{x,x' \sim P}[k(x,x')] 
+ \mathbb{E}_{y,y' \sim Q}[k(y,y')] 
- 2 \, \mathbb{E}_{x \sim P, y \sim Q}[k(x,y)],
\]
where $k(\cdot, \cdot)$ is a positive-definite kernel. In our experiments, we use the \emph{RBF kernel}
\[
k(x,y) = \exp\Big(-\gamma \|x-y\|^2 \Big),
\]
and set the kernel parameter $\gamma$ using the \emph{median of pairwise distances} between all ground truth samples (the median heuristic). Intuitively, a smaller MMD indicates that the generated samples are closer to the ground-truth distribution in the reproducing kernel Hilbert space defined by $k$.

\paragraph{Ill-conditioned Gaussian prior.} The sample efficiency comparison for the ill-conditioned Gaussian prior test is presented in \cref{fig:mcmc-ill-condition}.

\paragraph{Scattering problem.}
For the TV prior $p(x)\exp(-\lambda \text{TV}(x))$, we use a smoothed TV distance using finite difference:
\begin{align}
    TV_\epsilon(x) = \sum_{i,j} \sqrt{(x_{i+1,j} - x_{i,j})^2 + (x_{i,j+1} - x_{i,j})^2 + \epsilon^2}.
\end{align}
Here $x_{i,j}$ is the pixel at row $i$ and column $j$. We use $\epsilon>0$ ensures differentiability for gradient-based samplers like NUTS.

To assess the convergence of the samples from \ApproxMH and \HybridMH, we compare the sample
mean and variance of the generated samples with those computed from ground-truth
posterior samples.
\begin{align}
\mu_{\text{true}} = \frac{1}{N_{\text{true}}} \sum_{i=1}^{N_{\text{true}}} x_i^{\text{true}},\qquad
\sigma_{\text{true}}^2 = \frac{1}{N_{\text{true}} - 1} 
\sum_{i=1}^{N_{\text{true}}} \left(x_i^{\text{true}} - \mu_{\text{true}}\right)^2.
\end{align}
and for the MCMC chain up to iteration $t$:
\begin{align}
\hat{\mu}_t = \frac{1}{t} \sum_{i=1}^{t} x_i^{\text{mcmc}}, \qquad
\hat{\sigma}_t^2 = \frac{1}{t - 1} 
\sum_{i=1}^{t} \left(x_i^{\text{mcmc}} - \hat{\mu}_t\right)^2.
\end{align}

The ground-truth samples are generated using NUTS for the true posterior, collecting $2\times 10^5$ samples from 10 independent runs. Both $\mu$ and $\sigma^2$ are images, so we visualize convergence via the mean error map $|\hat{\mu}t - \mu{\text{true}}|$ and variance error map $|\hat{\sigma}t^2 - \sigma{\text{true}}^2|$ shown in \cref{fig:helmholtz-mean-err} and \cref{fig:helmholtz-var-err}. The plots indicate that \HybridMH achieves faster convergence than \ApproxMH.

\end{document}